\documentclass[10pt]{article} 
\usepackage[preprint]{tmlr}

\usepackage{amsmath,amsfonts,bm}

\def\eqref#1{equation~\ref{#1}}

\def\1{\bm{1}}

\DeclareMathAlphabet{\mathsfit}{\encodingdefault}{\sfdefault}{m}{sl}
\SetMathAlphabet{\mathsfit}{bold}{\encodingdefault}{\sfdefault}{bx}{n}

\PassOptionsToPackage{table}{xcolor}
\PassOptionsToPackage{table}{dvipsnames}
\usepackage[pagebackref,breaklinks,colorlinks]{hyperref}
\usepackage{url}
\usepackage{graphicx}
\usepackage{amsmath}
\usepackage{amssymb}
\usepackage{booktabs}
\usepackage{wrapfig}
\usepackage{mathtools}
\usepackage{colortbl}
\usepackage{soul}
\usepackage{multirow}

\usepackage[utf8]{inputenc} 
\usepackage[T1]{fontenc}    
\usepackage{amsfonts}       
\usepackage{nicefrac}       
\usepackage{microtype}      
\usepackage{bbding}
\usepackage[ruled]{algorithm2e}
\usepackage{algpseudocode}

\usepackage[toc,page,header]{appendix}
\usepackage{minitoc}

\usepackage[capitalize]{cleveref}
\crefname{section}{Sec.}{Secs.}
\Crefname{section}{Section}{Sections}
\Crefname{table}{Table}{Tables}
\crefname{table}{Tab.}{Tabs.}

\title{Training Vision-Language Transformers from Captions}

\author{
\begin{center}
    Liangke Gui$^{*1}$ \hspace{2em} 
    Yingshan Chang$^{*1}$ \hspace{2em} 
    Qiuyuan Huang$^{2}$ \hspace{2em} 
   Subhojit Som$^{2}$\\
   Alex Hauptmann$^{1}$ \hspace{3em} 
   Jianfeng Gao$^{2}$ \hspace{3em} 
   Yonatan Bisk$^{1}$\\
   \textnormal{$^1$Carnegie Mellon University} \hspace{3em} 
   \textnormal{$^2$Microsoft Research}\\
   \href{https://github.com/guilk/VLC}{github.com/guilk/VLC}
   \end{center}
}

\newcommand{\acronym}{\textbf{VLC}}
\newcommand{\acronymbase}{\textbf{VLC}$\mathrm{_{Base}}$}
\newcommand{\acronymlarge}{\textbf{VLC}$\mathrm{_{Large}}$}
\newcommand{\longname}{\textbf{V}ision-\textbf{L}anguage from \textbf{C}aptions}

\def\month{MM}  
\def\year{YYYY} 
\def\openreview{\url{https://openreview.net/forum?id=XXXX}} 

\begin{document}
\maketitle

\doparttoc 
\faketableofcontents 

\begin{abstract}
Vision-Language Transformers can be learned without low-level human labels (e.g. class labels, bounding boxes, etc). Existing work, whether explicitly utilizing bounding boxes ~\citep{chen2020uniter,tan2019lxmert,lu2019vilbert} or patches ~\citep{kim2021vilt}, assumes that the visual backbone must first be trained on ImageNet~\citep{russakovsky2015imagenet} class prediction before being integrated into a multimodal linguistic pipeline. We show that this is not necessary and introduce a new model \longname{} (\acronym) built on top of Masked Auto-Encoders~\citep{he2021masked} that does not require this supervision. In fact, in a head-to-head comparison between ViLT, a strong patch-based vision-language transformer which is pretrained with supervised object classification, and our model, \acronym, we find that our approach 1. outperforms ViLT on standard benchmarks, 2. provides more interpretable and intuitive patch visualizations, and 3. is competitive with many larger models that utilize ROIs trained on annotated bounding-boxes.
\end{abstract}

\section{Introduction}
\label{sec:intro}

\begin{wrapfigure}[24]{r}{0.5\textwidth}
\vspace{-30pt}
\centering
\includegraphics[width=0.95\linewidth]{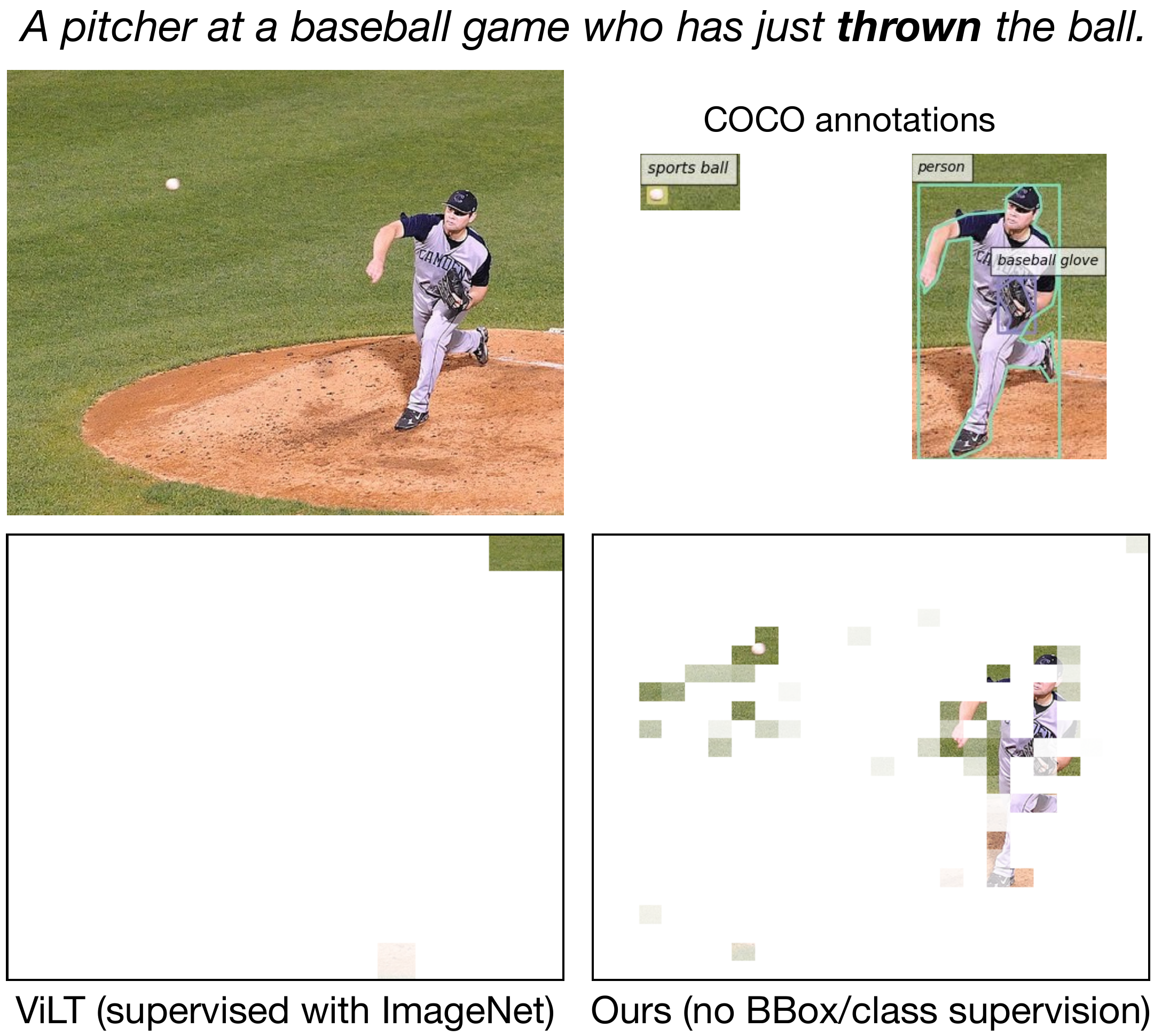}
\label{fig:fig1}

\caption{Visualized are the model's top aligned patches with the word \textbf{thrown}. Compared with VilT, our proposed model (\acronym{}) produces a more meaningful (if diffuse) distribution over the relevant patches. This work argues for the following factors towards learning such faithfully-aligned and flexible vision-language representations: (1) Masked Image Modeling, (2) Avoiding a vision-only processing stack, (3) Avoiding priors from ImageNet pretraining.}

\vspace{5pt}
\end{wrapfigure}

Should vision guide language understanding or does language structure visual representations? Vision-language (VL) transformers have put language first. Most popular vision-language transformers~\citep{chen2020uniter,tan2019lxmert,li2019visualbert,lu2019vilbert} only integrate vision from selected bounding boxes extracted by pretrained ImageNet~\citep{russakovsky2015imagenet} classifiers.  In this paradigm, the bag of visual tokens are embedded into an existing linguistic space (\emph{i.e.}, the lexical embeddings of BERT~\citep{Devlin2019}).

The introduction of ViT~\citep{dosovitskiy2020image} empowered the community to flip the paradigm.  Notably, ViLT~\citep{kim2021vilt} initializes with ViT, so it places visual representations as the initial conceptual space to which language must adhere. Additionally, there are engineering benefits to this paradigm as it removes the computationally expensive need for Region of Interest (ROI) extraction. However, because ViT is trained with supervised ImageNet labels, its representation may be constrained by the finite label space. The ImageNet concept space is still somewhat linguistic in nature when initialized, and requires expensive annotation, a hindrance to scaling to arbitrarily many visual categories.
\vspace{10pt}

We take the important next step and remove the need for supervised pretraining. An unsupervised visual semantics is learned via Masked Auto-Encoders~\citep{he2021masked} before language is integrated. This leads to both a better performing and more general model. In addition, every component can be improved and scaled with unsupervised and weakly aligned data -- removing the need for future annotation efforts while still scaling to open-vocabulary domains in the wild.

Our \longname{} (\acronym{}) model matches or outperforms nearly all vision-language transformers despite being 1. Smaller and faster at inference, 2. Avoiding use of ROIs, and 3. Not leveraging object-level supervised labels for pretraining. For downstream usage, token representations can be naturally adapted to producing class-labels \textbf{and} bounding-boxes, with moderate finetuning. We demonstrate strong performance of \acronym{} across a comprehensive set of benchmarks covering V+L understanding, retrieval, reasoning and grounding. Performance also continues to improve with data and model size scaling, and as it relies only on weak alignment of image-text pairs, future work with access to large compute may be able to continue driving up performance. Ablation study shows that key to consistence improvement on downstream tasks is the masked image modeling objective, which is in sharp contrast to existing findings. Finally, we provide several analyses on the underlying representations to understand what our models are learning and guide future V+L transformer research.

\section{Related Work}
\paragraph{Vision-Language Modeling.} Based on how they encode images, most existing works on vision-language modeling fall into three categories. The first category~\citep{lu2019vilbert,tan2019lxmert,li2019visualbert,su2019vl,chen2020uniter,li2020oscar,zhang2021vinvl,li2020unimo,qi2020imagebert} focuses on using pre-trained object detectors to extract region-level visual features (e.g., by Faster R-CNN~\citep{ren2015faster}). 
In particular, OSCAR~\citep{li2020oscar} and VinVL~\citep{zhang2021vinvl} further boost the performance by feeding additional image tags into the transformer model. However, extracting region-level features requires pretrained object detectors with high-resolution inputs that can be time-consuming. To tackle these two issues, the second category~\citep{huang2020pixel,huang2021seeing,jia2021scaling} proposes to encode images by using grid features from convolutional neural networks. SOHO~\citep{huang2021seeing} discretizes the grid features by a learnable vision dictionary, and feeds the discretized features to their cross-modal module. The third category~\citep{dou2021meter,xue2021probing,li2021align,radford2021learning} uses a Vision Transformer (ViT)~\citep{dosovitskiy2020image} as the image encoder and designs different objective functions for vision-language pretraining. The most similar to our work is ViLT~\citep{kim2021vilt}. ViLT does not use pretrained object detectors or extra visual embedders for visual embedding, but still needs weights pretrained on ImageNet-21K for initialization. Different from the previous work, we show such supervised initialization, pretrained object detectors or visual embedders are not necessary. While momentum distillation~\citep{li2021align} and image-text contrastive loss~\citep{li2021align,li2022blip} are shown effective in previous work, such techniques are orthogonal to our work and not included in our discussion.

\vspace{-0.1cm}
\paragraph{Masked Language Modeling.} Masked language modeling (MLM) and its auto-regressive counterparts are widely used in natural language processing for learning text representations. MLM~\citep{Devlin2019} trains a model to predict a random sample of input tokens that have been masked in a multi-class setting. In vision-language pretraining, MLM has shown useful to enforce the consistency across modalities~\citep{dou2021meter,zhang2021vinvl,li2020oscar,kim2021vilt}. In vision-language modeling, we randomly mask some of the input tokens, and the model is trained to reconstruct the original tokens given the masked tokens and their corresponding visual inputs. To be consistent with previous work, we follow the default settings of training BERT~\citep{Devlin2019} for masked language modeling.

\vspace{-0.1cm}
\paragraph{Masked Image Modeling.} Masked image modeling (MIM) is a pretext task to learn representations from images corrupted by masking. Inspired by the success of masked language modeling (MLM) in NLP, different masked prediction objectives have been proposed for image tasks. iGPT~\citep{chen2020generative} predicts unknown pixels of a sequence. ViT~\citep{dosovitskiy2020image} predicts mean colors of masked patches. BEiT~\citep{bao2021beit} proposes to use a pre-trained discrete variational autoencoder (dVAE)~\citep{ramesh2021zero} to encode masked patches. MaskFeat~\citep{wei2021masked} predicts HoG~\citep{dalal2005histograms} features of the masked image regions. SimMIM~\citep{xie2021simmim} and MAE~\citep{he2021masked} predict RGB values of raw pixels by direct regression. MIM has also been explored in the field of vision-language representation learning by either regressing the masked feature values~\citep{tan2019lxmert,chen2020uniter,dou2021meter,kim2021vilt} or predicting a distribution over semantic classes for the corresponding image region~\citep{chen2020uniter,lu2019vilbert,su2019vl}. In contrast to previous approaches~\citep{chen2020uniter, kim2021vilt, dou2021meter} that show MIM does not contribute to or hurt the performance on downstream tasks, we show that using MIM can consistently improve the performance as the training steps increase. 

\section{Method}
\begin{figure*}[!ht]
    \centering
    \includegraphics[trim={0cm 5pt 0 5pt},clip, 
    width=0.8\textwidth]{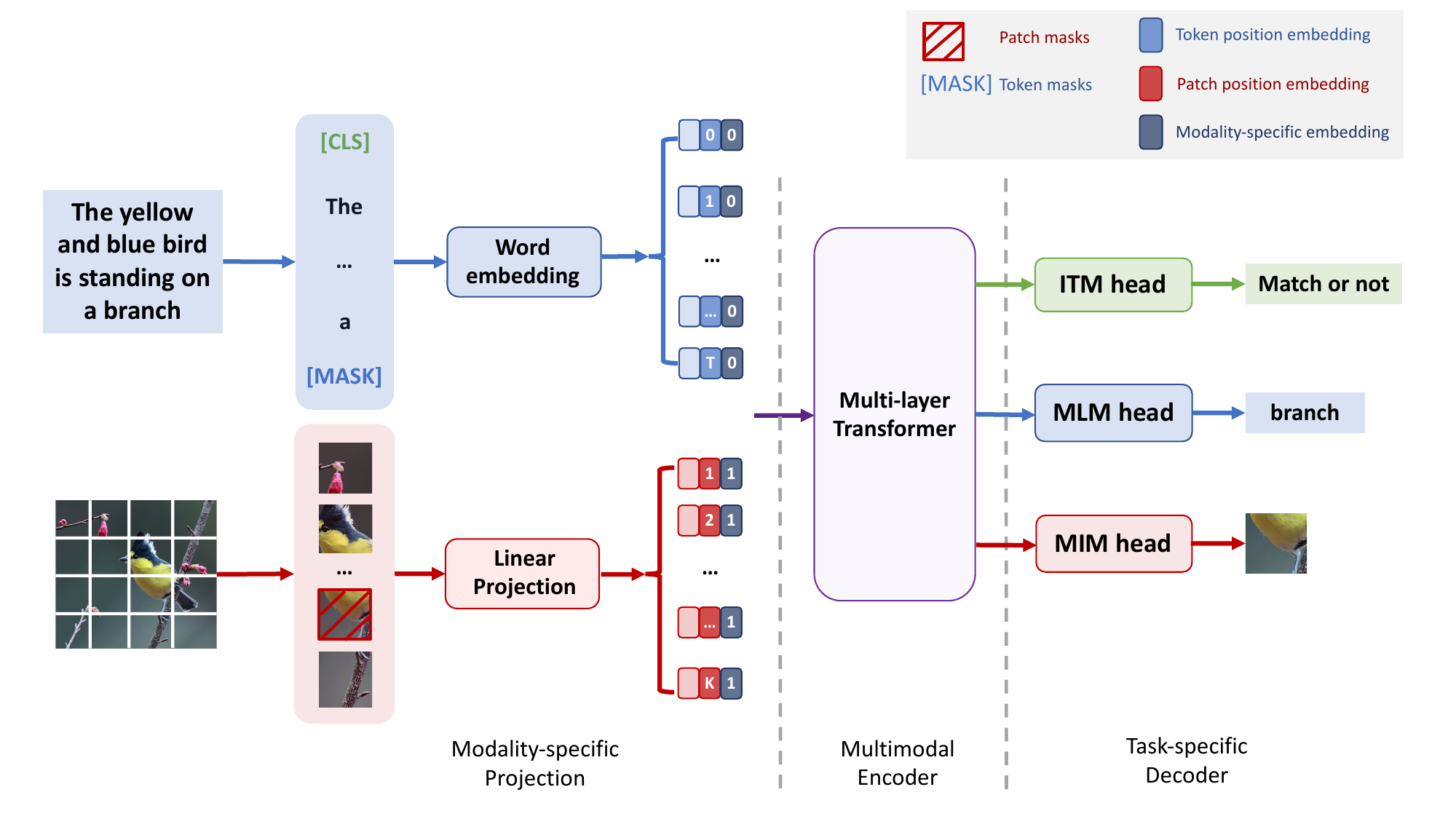}
    \vspace{-2mm}
\caption{The overall architecture of \acronym{}. Our model consists of three modules: (1) Modality-specific projection. A simple linear layer projects image patches and token embeddings into the same representation space; (2) Multi-modal encoder. We use a 12-layer ViT~\citep{dosovitskiy2020image} initialized from MAE~\citep{he2021masked} (ImageNet-1K without labels) as our Transformer backbone; (3) Task-specific decoder. We pretrain multi-modal representations with masked image/language modeling (MIM/MLM) and image-text matching objectives. This work justifies the importance of the MIM objective throughout V+L pretraining, in addition to providing good initializations for the Transformer encoder.}
\label{fig:framework}
\vspace{-4mm}
\end{figure*}

\subsection{Model Architecture}
Our aim is a vision-language transformer that can be trained without the need for expensive object-level supervised labels (\emph{e.g.}, class labels or object bounding boxes). More concretely, our results empirically show that such object based supervised signals are \emph{not} necessary for vision-language pretraining. To this end, we use a ViT-based framework to learn multi-modal representations by 1) intra-modal reconstruction through masked image/language modeling; 2) inter-modal alignment through image-text matching. The architecture of our proposed \acronym{} framework is illustrated in Figure~\ref{fig:framework}. \acronym{} consists of a modality-specific projection module~\ref{sec:projection}, a multi-modal encoder~\ref{sec:encoder} and three task-specific decoders~\ref{sec:decoder}. We aim for minimal visual and textual embedding designs during pretraining. 

\subsection{Modality-specific Projection Module}
\label{sec:projection}
While most of existing methods rely on complex ResNeXt~\citep{huang2020pixel} or object detection components~\citep{chen2020uniter, lu2019vilbert,zhang2021vinvl,li2020oscar}, we use a trainable \emph{linear projection} layer to map flattened visual patches to the visual embedding space. The patch embeddings are represented as $\mathbf{v}=\{v_{1},...,v_{K}\}\in \mathbb{R}^{K\times d}$, where $K$ is the number of image patches and $d$ is the hidden dimension of our model. For text embedder, we follow BERT~\citep{Devlin2019} to tokenize the input sentence into WordPieces~\citep{wu2016google}. We then adopt a word embedding lookup layer to project tokenized words to the textual embedding space. Here we use $\mathbf{w}=\{w_{CLS}, w_{1},...,w_{T}\}\in \mathbb{R}^{T\times d}$ to represent the token embeddings, where $T$ is the number of tokens and the special token \texttt{CLS} denotes the start of the token sequence. We encode patch and token positions separately by $v^{pos}\in \mathbb{R}^{1\times d}$ and $w^{pos}\in \mathbb{R}^{1\times d}$. We use $v^{type}\in \mathbb{R}^{1\times d}$ and $w^{type}\in \mathbb{R}^{1\times d}$ as modality-type embeddings to distinguish the modality difference between patch and token embeddings. The final representations of each patch $v_i$ and token $w_j$ are calculated as 
\begin{align}
    \hat{v}_i =& \text{LayerNorm}(v_i+v^{pos}_{i}+v^{type}), \hspace{1em} \mathrm{and} \\
    \hat{w}_j =& \text{LayerNorm}(w_j+w^{pos}_{j}+w^{type}).
\end{align}

\subsection{Multi-modal Encoder}
\label{sec:encoder}
To learn the contextual representations from both visual and textual modality, we follow single-stream approaches~\citep{kim2021vilt, chen2020uniter} and use the ViT-B/16 architecture as our multi-modal encoder. ViT-B/16 consists 12 alternating layers of multiheaded self-attention (MSA) and MLP blocks. LayerNorm comes before every block and residual connections after every block~\citep{dosovitskiy2020image}. 

We use a merged-attention~\citep{dou2021meter} mechanism to fuse the visual and textual modalities. More specifically, we concatenate the token and patch embeddings together as $\{\hat{w}_{CLS}, \hat{w}_1,..., \hat{w}_T, \hat{v}_1,...,\hat{v}_K\}$, then feed them into the transformer. We use the hidden states $h$ at the output of the last layer of the encoder as the contextual representations $\{h_{CLS}, h_1^{w},...,h_T^w, h_1^v, ..., h_K^v\}$. 

In sharp contrast to existing approaches that use object detectors, visual detectors pretrained with supervised labels or pretrained language models (\emph{e.g.}, BERT, Roberta), we initialize our model with MAE pretrained on ImageNet-1K with no labels.

\subsection{Pretraining Objectives}
\label{sec:decoder}
To learn a universal visual and textual representation for vision-and-language tasks, we apply self-supervised methods to pre-train a model on a large aggregated dataset. Unlike previous approaches that only mask text tokens, we randomly mask both image patches and text tokens simultaneously. We train our model with three objectives: masked image modeling (MIM), masked language modeling (MLM) and image-text matching (ITM). 

\paragraph{Masked Language Modeling.} In language pretraining, MLM randomly masks input tokens, and the model is trained to reconstruct the original tokens based on unmasked context. Following BERT~\citep{Devlin2019}, we randomly mask text tokens with a probability of $0.15$, and replace the masked ones $\mathbf{w_m}$ with a special token \texttt{[MASK]}. The goal is to predict the masked tokens based on both non-masked text tokens $\mathbf{w_{\setminus m}}$ and image patches $\mathbf{v_{\setminus m}}$. The learning target $\mathcal{L}_{MLM}$ can be formulated as 
\begin{equation}
    \mathcal{L}_{MLM} = -\mathbb{E}_{(\mathbf{w},\mathbf{v})\sim D} \log p(\mathbf{w_{m}|\mathbf{w_{\setminus m}}, \mathbf{v_{\setminus m}}}).
\end{equation}
We use a linear layer with default parameters~\citep{Devlin2019} as the MLM head to output logits over the vocabulary, which are used to compute the negative log likelihood loss for the masked text tokens.

\paragraph{Masked Image Modeling.} Existing approaches explore MIM either by regressing the masked features values~\citep{chen2020uniter,kim2021vilt,xue2021probing} or by predicting a distribution over semantic classes for a certain image region~\citep{chen2020uniter,lu2019vilbert,dou2021meter}. In contrast, we follow MAE~\citep{he2021masked} to randomly mask image patches with a probability of $0.6$, and reconstruct the missing pixels based on both non-masked tokens $\mathbf{w_{\setminus m}}$ and patches $\mathbf{v_{\setminus m}}$. The learning target $\mathcal{L}_{MIM}$ can be formulated as
\begin{equation}
    \mathcal{L}_{MIM}=\mathbb{E}_{(\mathbf{w},\mathbf{v})\sim D}f(\mathbf{v_{m}}|\mathbf{w_{\setminus m}}, \mathbf{v_{\setminus m}}),
\end{equation}
where the feature regression objective $f$ is to regress the masked image patch representations to pixel values. We use 8-layer transformer as the MIM head $r$. For a masked image patch $v_i$, the objective $f$ can be formulated as: $f(v_{i}|\mathbf{w_{\setminus m}}, \mathbf{v_{\setminus m}}) = ||r(h_i^v)-v_{i}||^2$. Each output of the MIM head is a vector of pixel values representing a patch. Different from the observations in ViLT~\citep{kim2021vilt} and METER~\citep{dou2021meter}, we show MIM can consistently improve the performance on downstream tasks as training steps increase.

\paragraph{Image-Text Matching.} Given a batch of image and text pairs, the ITM head identifies if the sampled pair is aligned. We randomly replace the aligned image with a different one with a probability of $0.5$. We use the special token \texttt{[CLS]} as the fused representation of both modalities, and feed $h_{CLS}$ to the ITM head. The learning target $\mathcal{L}_{ITM}$ can be formulated as 
\begin{equation}
    \mathcal{L}_{ITM} = -\mathbb{E}_{(\mathbf{w},\mathbf{v})\sim D} \log p(y|\mathbf{w}, \mathbf{v}),
\end{equation}

Where $y\in\{0,1\}$ indicates whether the image and text are matched ($y=1$) or not ($y=0$). We use a single linear layer as the ITM head and compute negative log likelihood loss as our ITM loss.
We weight the pretraining objectives equally  so the full pre-training objective is:
\begin{equation}
    \mathcal{L} = \mathcal{L}_{MLM} + \mathcal{L}_{ITM} + \mathcal{L}_{MIM}
\end{equation}

For a fair comparison with existing approaches, We do not include image-text contrastive loss~\citep{li2021align, li2022blip}, momentum distillation~\citep{li2021align} and other techniques in our implementation.

\section{Experimental Setup}

\subsection{Pre-training Datasets}
Following previous work~\citep{chen2020uniter,kim2021vilt,li2021align,dou2021meter}, our pre-training corpus comprises four commonly used vision-language datasets including COCO~\citep{lin2014microsoft}, Visual Genome~\citep{krishna2017visual}, Google Conceptual Captions~\citep{sharma2018conceptual} and SBU Captions~\citep{ordonez2011im2text}, totalling $4.0$M unique images and $5.1$M image-text pairs. To show the benefits of data-scaling, we also use the VinVL~\citep{zhang2021vinvl} pretraining data which includes Flickr30k~\citep{young2014image}, GQA~\citep{hudson2019gqa}, VQA~\citep{goyal2017making}, VG-QAs~\citep{krishna2017visual} and a subset of OpenImages~\citep{openimages}. This larger pre-training corpus contains $5.65$M unique images (see detailed statistics in Appendix A.1). 
Future work can trivially grow the size of the corpus by including large-scale web crawls.

\subsubsection{Fine-tuning Datasets}

We fine-tune our model on \textbf{image-text retrieval:} Flickr30K~\citep{plummer2015flickr30k} and MSCOCO~\citep{lin2014microsoft}, \textbf{image-text understanding:} VQAv2~\citep{goyal2017making} and NLVR$^2$~\citep{suhr2018corpus}, and \textbf{image-text grounding:} Refcoco~\citep{Refcoco, Refcocog} tasks. We fine-tune all encoder layers together with task-specific heads. For retrieval tasks, we follow the standard splits and evaluate our models in the finetuning settings.
For VQAv2, we follow the standard practice~\citep{chen2020uniter,li2021align} to train the models with both training, validation and additional question-answer pairs from Visual Genome while reserving $1,000$ validation samples for internal validation. For Refcoco, we propose a new algorithm to obtain bounding box outputs based on affinities between patch and text encodings. This only requires fine-tuning the encoder backbone without introducing extra parameters. Finally, ablation studies validate our key design components. More experimental details can be found in Appendix A.2 and A.3.

\subsection{Implementation Details}
We pretrain two variants of the multi-modal encoder which uses a $86$M parameter ViT-B/16 denoted as \acronymbase{} and $307$M parameter ViT-L/16 denoted as \acronymlarge{}. Both variants are initialized with MAE pre-trained on ImageNet-1K without labels. For text inputs, we tokenize text with the \textit{bert-base-uncased} and \textit{bert-large-uncased} tokenizer, respectively. The text embedding parameters are learned from scratch, in lieu of loading pre-trained BERT weights. We randomly mask image patches with a probability of $0.6$ and text tokens with a probability $0.15$. To accelerate training, we follow MAE~\citep{he2021masked} and skip the mask token \texttt{[MASK]} in the encoder and only apply it in the lightweight decoder. We use AdamW~\citep{loshchilov2017decoupled} with a weight decay of $0.01$. The learning rate is warmed-up to $1e^{-4}$ in the first $10\%$ of total training steps and is decayed to zero for the rest of the training following a linear schedule. During pre-training, we resize the shorter edge of input images to $384$, take random image crops of resolution $384\times384$, and apply RandAugment~\citep{cubuk2020randaugment}. We pre-train for $200k$ steps with a batch size of $4,096$. 
For the parameter estimation, we exclude the textual embedder as it is shared by all vision-language transformers. We also exclude the parameters of all the auxiliary heads as they are only required during pretraining. Unless otherwise specified, we use the \emph{base} version of \acronym{} for downstream evaluation and visualization.

\section{Downstream Experiments \& Results}

We evaluate the effectiveness of our pre-training strategy under both zero-shot and fine-tuning settings. We produce competitive results across a diverse set of standardized V+L benchmarks, broadly categorized into Image-Text Retrieval, Image-Text Understanding and Image-Text Grounding. In addition to evaluating on the benchmarks involved in finetuning, we also perform inference-only experiments on multiple challenging benchmarks testing for non-trivial reasoning \citep{Winoground, kilogram} and open-domain question answering \citep{schwenk2022okvqa}.

\subsection{Zero-shot Experiments}
\begin{wraptable}[15]{r}{0.35\linewidth}
\vspace{-10pt}
\centering
\renewcommand{\arraystretch}{1.2}
\renewcommand{\tabcolsep}{1.2mm}
\small
\begin{tabular}{@{}l@{\hspace{5pt}} c@{\hspace{5pt}} c@{\hspace{10pt}} c@{} }
\toprule
Input Condition & ViLT & Ours & Human \\  
\midrule
WHOLE+BLACK & 12.9 & \textbf{13.8} & 47.7\\
PARTS+BLACK & 12.5 & \textbf{15.2} & 49.1 \\
WHOLE+COLOR & 11.7 & \textbf{13.9} & 49.5 \\
PARTS+COLOR & 10.7 & \textbf{13.5} & 63.0 \\
\bottomrule
\end{tabular}
\caption{Zero-shot inference accuracy on Kilogram (dev), a challenging image-text matching task for recognizing abstract visual entities from linguistic descriptions. \acronym{} consistently outperforms ViLT across all input conditions. 
} 
\label{exp:zeroshot}
\end{wraptable}
First we verify the strength of our approach in a zero-shot setting before investing in larger fine-tuning results.
We adapt our model to perform zero-shot visual reasoning on the Kilogram \citep{kilogram} dataset. Kilogram challenges a model to recognize an abstract Tangram shape from a language description. Tangram shapes drastically differ from natural scenes as they only contain abstract and implicit visual clues. Hence, recognition demands more sophisticated reasoning about the interplay between visual and linguistic cues. The task is formulated into a 10-way classification problem where the model chooses the most relevant image given an abstract shape description. Specifically, we rank image relevance by the logits calculated by the pretrained image-text matching head. Table~\ref{exp:zeroshot} shows that our model consistently outperforms ViLT across all input conditions. This supports our design consideration that freeing up the model from priors of limited ImageNet concepts  helps it generalize to arbitrarily many linguistically-specified visual categories. Further improvement may come from additional finetuning to familarize our model with the Tangram shape domain, which is left for future work.

\begin{table*}[t]
\small
\centering
\renewcommand{\arraystretch}{1.2}
\renewcommand{\tabcolsep}{1.2mm}
\resizebox{1.0\linewidth}{!}{
\begin{tabular}{@{}l@{\hspace{15pt}}r  @{\hspace{15pt}} c@{\hspace{5pt}} c@{\hspace{5pt}} c@{\hspace{1em}} c@{\hspace{5pt}} c@{\hspace{5pt}} c @{\hspace{15pt}} c@{\hspace{5pt}} c @{\hspace{5pt}}c @{\hspace{1em}} c @{\hspace{5pt}}c @{\hspace{5pt}}c@{}}
\toprule
\multirow{3}{*}{Model} & &  \multicolumn{6}{c}{\bf Text Retrieval} & \multicolumn{6}{c}{\bf Image Retrieval} \\
& & \multicolumn{3}{@{}c@{\hspace{1em}}}{Flickr30K (1K)} 
& \multicolumn{3}{@{}c@{\hspace{2em}}}{MSCOCO (5K)} 
& \multicolumn{3}{@{}c@{\hspace{1em}}}{Flickr30K (1K)} 
& \multicolumn{3}{@{}c@{}}{MSCOCO (5K)} \\
& Params & @1 & @5 & @10 & @1 & @5 & @10 & @1 & @5 & @10 & @1 & @5 & @10 \\ 
\midrule 
ALBEF$^{\dag}$~\citep{li2021align} & 163M & \underline{94.3} & \underline{99.4} & \underline{99.8} & 73.1 & 91.4 & 96.0 & \textbf{82.8} & \textbf{96.7} & \textbf{98.4} & 56.8 & 81.5 & 89.2\\
VinVL$\mathrm{_{Large}}$~\citep{zhang2021vinvl}& 452M & - & - & - & 75.4 & 92.9 & 96.2 & - & - & - & \textbf{58.8} & \underline{83.5} & \underline{90.3} \\
UNITER$\mathrm{_{Large}}$~\citep{chen2020uniter} & 371M & 87.3 & 98.0 & 99.2 & 65.7 & 88.6 & 93.8 & 75.6 & 94.1 & 96.8 & 52.9 & 79.9 &  88.0\\
METER-Swin$\mathrm{_{Base}}$~\citep{li2020oscar} & 288M & 92.4 & 99.0 & 99.5 & \underline{76.2} & \underline{93.2} & \underline{96.8} & 79.0 & 95.6 & 98.0 & 54.9 & 81.4 & 89.3\\
PixelBERT~\citep{huang2020pixel} & 144M & 87.0 & 98.9 & 99.5 & 63.6 & 87.5 & 93.6 & 71.5 & 92.1 & 95.8 & 50.1 & 77.6 & 86.2 \\
ViLT~\citep{kim2021vilt} & 86M & 83.5 & 96.7 & 98.6 & 61.5 & 86.3 & 92.7 & 64.4 & 88.7 & 93.8 & 42.7 & 72.9 & 83.1 \\ 
\midrule
\midrule
\acronymbase{} (with 5.6M examples) & 86M & 89.2 & 99.2 & 99.8 & 71.3 & 91.2 & 95.8 & 72.4 & 93.4 & 96.5 & 50.7 & 78.9 & 88.0 \\
\acronymlarge{} (with 5.6M examples) & 307M & \textbf{94.4} & \textbf{99.6} & \textbf{99.9} & \textbf{76.7}  & \textbf{94.5}  & \textbf{97.3}  & \underline{79.1} & \underline{95.8} & \underline{98.2} & \underline{58.4} & \textbf{84.0}  & \textbf{91.1}  \\
\bottomrule
\end{tabular}}
\caption{On Image-text retrieval tasks, we see substantial gains over state-of-the-art methods with bounding-box pretraining. $^{\dag}$ALBEF uses specifically designed coarse-to-fine objectives for the image-text retrieval task.} 
\label{exp:retrieval}
\vspace{-.3cm}
\end{table*}

\subsection{Finetuning Experiments}

Next we investigate the role that fine-tuning has on our downstream performance across traditional retrieval and understanding benchmarks, before analyzing training and performance curves.

\subsubsection{Image-Text Retrieval} 
We begin with a proof of concept experiment, evaluating our model on the Karpathy splits of the Flickr30K~\citep{plummer2015flickr30k} and MSCOCO~\citep{lin2014microsoft} benchmarks. Table \ref{exp:retrieval} compares VLC to strong multimodal transformers which leverage ROIs, more parameters, and are pretrained on ImageNet classification. Note that as most of detection-based models have the advantage of using Faster R-CNN~\citep{ren2015faster} pre-trained on VG~\citep{krishna2017visual} or MSCOCO~\citep{lin2014microsoft}. 

The closest comparison to \acronymbase{} is ViLT as it is the same model size, though still requires more supervised data in the form of ImageNet classification pretraining for ViT \citep{dosovitskiy2020image}\footnote{ViLT uses ViT-B/32 pretrained with ImageNet-21K and finetuned on ImageNet-1K with supervised labels.}. When comparing to dual-encoder models, our \acronymlarge{} achieves competitive results across all settings. ALBEF uses pre-trained ViT and BERT model for initialization. Additionally, it specifically designs the coarse-to-fine objectives while we directly fine-tune the pre-trained ITM head for retrieval tasks. Thus we treat ALBEF as a strongest available baseline.

\subsubsection{Image-Text Understanding}

\begin{wraptable}[30]{r}{0.5\linewidth}
\vspace{-10pt}
\centering
\renewcommand{\arraystretch}{1.2}
\renewcommand{\tabcolsep}{1.2mm}
\resizebox{1.0\linewidth}{!}{
\begin{tabular}{@{}l@{\hspace{0pt}} l@{\hspace{0pt}} r@{\hspace{5pt}} c@{\hspace{5pt}} c@{\hspace{10pt}} c@{\hspace{5pt}} c@{} }
\toprule
& \multirow{2}{*}{Model} & &  \multicolumn{2}{@{}c@{}}{\bf VQAv2} test & \multicolumn{2}{@{}c@{}}{\bf NLVR$^2$} \\  
& & Param & -dev & -std & dev & test \\ 

\midrule
\multicolumn{7}{@{}c@{}}{\emph{Supervised ImageNet Bounded Boxes}}\\
& ViLBERT~\citep{lu2019vilbert} & 274M & 70.55 & 70.92 & - & - \\
& LXMERT~\citep{tan2019lxmert} & 240M & 72.42 & 72.54 & 74.90 & 74.50 \\
& VisualBERT~\citep{li2019visualbert} & 170M &  70.80 & 71.00 & 67.4 & 67.0 \\
  & UNITER$\mathrm{_{Large}}$~\citep{chen2020uniter} & 371M & 73.82 & 74.02 & 79.12 & 79.98 \\
  & OSCAR$\mathrm{_{Large}}$~\citep{li2020oscar} & 371M & 73.61 & 73.82 & 79.12 & 80.37 \\
  & VinVL$\mathrm{_{Large}}$ $^{\dag}$~\citep{zhang2021vinvl} ($5.6$M) & 452M & \underline{76.52} & \underline{76.60} & \textbf{82.67} & \textbf{83.98} \\
\midrule
\multicolumn{7}{@{}c@{}}{\emph{Supervised ImageNet Classes}}\\
& METER-Swin$\mathrm{_{Base}}$ $^\ddag$~\citep{dou2022empirical} & 288M & 76.43 & 76.42 & 82.23 & 82.47 \\ 
& ALBEF~\citep{li2021align} & 163M & 74.54 & 74.70 & 80.24 & 80.50 \\
 & Visual Parsing~\citep{xue2021probing} & 180M & 74.00 & 74.17 & 77.61 & 78.05 \\
& PixelBERT~\citep{huang2020pixel} & 144M & 74.45 & 74.55 & 76.5\phantom{0} & 77.2\phantom{0} \\
 & ViLT~\citep{kim2021vilt} & 86M & 71.26 & - & 75.70 & 76.13 \\ 
\midrule
\multicolumn{7}{@{}c@{}}{\emph{No supervised classes or bounding boxes}}\\
& \acronymbase{} (ours -- 4M) & 86M & 72.98 & 73.03 & 77.04 & 78.51 \\
& \acronymbase{} (ours -- 5.6M) & 86M & 74.02 & 74.0\phantom{0} & 77.70 & 79.04 \\ 
& \acronymlarge{} (ours -- 5.6M) & 307M & \textbf{76.95} & \textbf{77.02}  & \underline{82.27} & \underline{83.52} \\ 
\midrule
\multicolumn{7}{@{}c@{}}{\emph{Pre-trained or initialized with $>10$M data}}\\
& {\color{gray}{X-VLM~\citep{zeng2021multi} (16M)}} & {\color{gray}{216M}} & {\color{gray}{78.22}} & {\color{gray}{78.37}} & {\color{gray}{84.41}} & {\color{gray}{84.76}} \\ 
& {\color{gray}{BLIP~\citep{li2022blip} (129M)}} & {\color{gray}{252M}} & {\color{gray}{78.25}} & {\color{gray}{78.32}} & {\color{gray}{82.15}} & {\color{gray}{82.24}} \\ 
& {\color{gray}{OFA~\citep{wang2022unifying} (54M)}} & {\color{gray}{930M}} & {\color{gray}{82.0\phantom{0}}} & {\color{gray}{82.0\phantom{0}}} & {\color{gray}{-}} & {\color{gray}{-}} \\
& {\color{gray}{CoCa~\citep{yu2022coca} (4.8B)}} & {\color{gray}{2.1B}} &{\color{gray}{82.3\phantom{0}}} & {\color{gray}{82.3\phantom{0}}} &{\color{gray}{86.1\phantom{0}}} & {\color{gray}{87.0\phantom{0}}} \\

\bottomrule
\end{tabular}}
\caption{
Despite that \acronym{} is only pretrained with weakly-aligned image-caption pairs, it matches or outperforms larger and more heavily supervised approaches within a similar scale of training data and model size. $^{\dag}$VinVL uses the object detector trained with $6.8M$ labeled imageNet images and 2.5M images with bounding box annotations. $^\ddag$METER-Swin$\mathrm{_{Base}}$ uses Swin-B trained with $14M$ labeled ImageNet as the image encoder and pretrained Roberta as the text encoder.} 

\label{exp:vqa}
\vspace{-.45cm}
\end{wraptable}

Table~\ref{exp:vqa} presents results on two popular image-text understanding datasets: VQAv2 and NLVR$^2$. 
We use the same training data as ViLT denoted as $4$M and VinVL denoted as $5.6$M.

\paragraph{Comparison to models supervised/initialized with ImageNet bounded boxes.} Most of these models use object detectors pretrained on VG~\citep{krishna2017visual} or MSCOCO~\citep{lin2014microsoft} to extract region features. Object detectors help in VQA tasks as they mainly ask about objects. Within the similar scale of pretraining data, our model achieves competitive performance on both tasks. Note that our model uses $384\!\times\!384$ or $576\!\times\!576$ as input resolution during our fine-tuning stages. This resolution is much lower compared with previous work using $800\!\times\!1333$~\citep{lu2019vilbert,chen2020uniter}. In particular, \emph{VinVL}~\citep{zhang2021vinvl} has a multi-stage pre-training for its object detector that has access to ImageNet-5K~\citep{xie2017aggregated} (6.8M images from 5K classes) and four object detection datasets~\citep{shao2019objects365,krishna2017visual,lin2014microsoft,openimages} (2.5M images with bounding box annotations). 
Our \acronymlarge{}, which has a similar model size, achieves better performance on the VQA task and competitive results on the NLVR task without any supervised initialization.

\paragraph{Comparison to models with supervised ImageNet classes.} Most of these approaches use additional visual embedders together with a pretrained BERT as their backbones. For example, ALBEF~\citep{li2021align}, Visual Parsing~\citep{xue2021probing}, PixelBERT~\citep{huang2020pixel} use pre-trained ViT-B/16, Swin transformer, ResNeXt-152 as their visual embedder, respectively. All these embedders are trained with labeled ImageNet data. In particular, \emph{METER-Swin$\mathrm{_{Base}}$} uses Swin-B/16 pretrained with more than $14$M labeled ImageNet22K images as the image encoder and pretrained Roberta as the text encoder. Experiments show that our model achieves better results than larger and more heavily supervised approaches.

Note that there are some other baselines which have access to a much larger scale of data during pretraining. For example, METER-CLIP-ViT$\mathrm{_{Base}}$ uses CLIP as the image encoder which is trained with $400$M image-text pairs. X-VLM~\citep{zeng2021multi} uses the same image encoder as METER-Swin$\mathrm{_{Base}}$ but with extra bounding box annotations and two times larger training data. While these models achieve higher performance, the comparisons are out of the scope.

\subsubsection{Image-Text Grounding}
\label{sec:grounding}

\begin{figure}[t!]
    \includegraphics[width=0.5\linewidth]{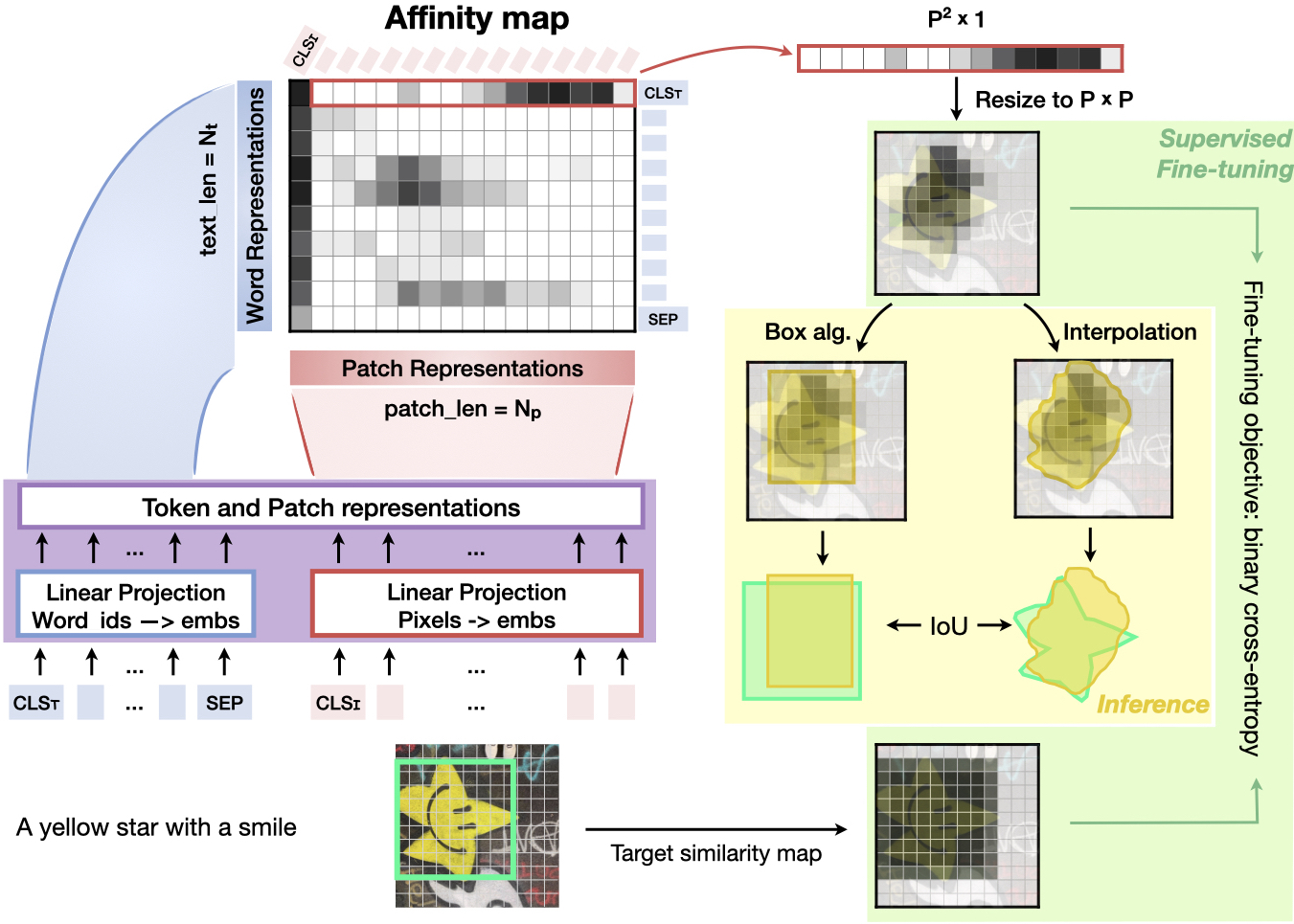}
    \hspace{3pt}
    \includegraphics[width=0.46\linewidth]{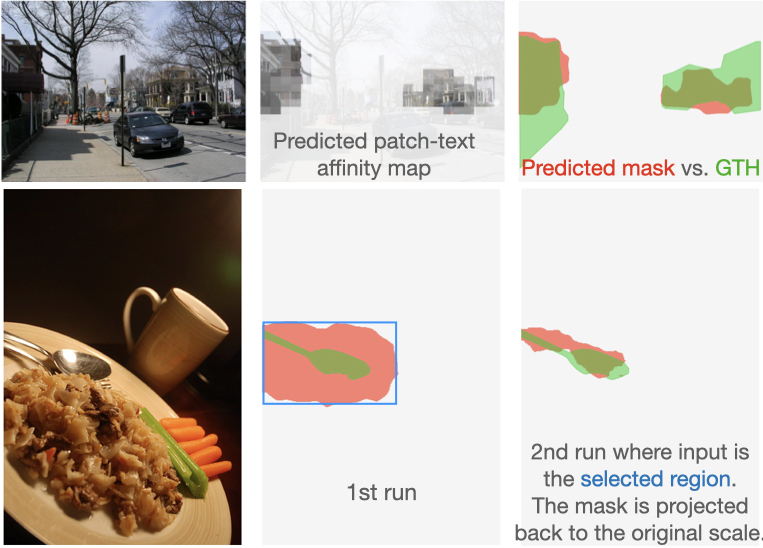}
    \caption{\textbf{Left:} Extending VLC to efficiently performing image-text grounding tasks: We finetune the Transformer backbone with supervision on similarities between patch and text representations. During inference, a bbox/mask can be obtained from the affinity scores with minimal computation overheads. \textbf{Right:} Reasonable segmentation masks resulted from interpolating patch-text affinities. Running the same inference procedure twice with coarse-to-fine resolution improves precision.}
    \label{fig:REC}
    \vspace{-0.4cm}
\end{figure}

Since the model was not pretrained to generate bounding boxes, we propose a novel algorithm which produces a bounding box based on patch-text affinities in a single forward pass. This results in much less inference time and higher parameter efficiency compared with competitive pipelines tailored to bounding box generation. See Figure~\ref{fig:REC} (Left) for an overview of the proposed grounding workflow and Appendix A.4 for details.

We finetune the encoder on the training sets of Refcoco, Refcoco+, and Refcocog \citep{Refcoco, Refcocog} following the umd split. We convert ground-truth bounding boxes to patch-text affinity scores and use them to finetune the preceding representations. At inference time, we algorithmically predict a bounding box for each referential expression from affinity scores between the last layer patch and text representations.

\paragraph{Comparison to modularized models} Previous models either use an RoI (Region-of-Interest) extractor to produce candidate bboxes to choose from \citep{chen2020uniter, VILLA, VL-T5, ReCLIP}, or directly predict the coordinates and dimensions of bboxes in the image coordinate system \citep{TransVG}. The former approach relies on an off-the-shelf detector, while the latter requires gating mechanisms to infuse linguistic information into a visual backbone. Our approach outperforms all previous approaches with modular designs (Table~\ref{tb:REC}, Top), meanwhile achieving the best inference-time efficiency. This justifies our unified approach in which both understanding and localization tasks can benefit from large-scale representation learning.

\paragraph{Comparison to unified models that have far more parameters
or computation} Most of performant models under comparison incorporated bounding box annotations from VG~\citep{krishna2017visual} in their pretraining data (Table~\ref{tb:REC}, Bottom). With a much lighter architecture and much less annotated data, our model already achieves competitive performance. This holds promise that the performance will continue to improve as resolution and data are scaled up during pretraining. 

\begin{wraptable}[9]{r}{0.4\linewidth}
\small
\vspace{-.4cm}
\resizebox{1.0\linewidth}{!}{
\begin{tabular}{@{}lc@{\hspace{10pt}}c@{\hspace{6pt}}c@{\hspace{6pt}}@{}}
\toprule
 Model & \multicolumn{3}{@{}c@{}}{Refcoco/+/g(umd) val} \\
\midrule
 TSEG-CRF~\citep{TSEG} & 25.95 & 22.62 & 23.41 \\
\acronymbase{} (ours) & 50.65 & 45.88 & 46.37  \\
\bottomrule

\end{tabular}}
\vspace{-.2cm}
\caption{We achieve non-trivial pixelwise IoU on Referring Expression Segmentation through interpolating low-resolution affinity maps into segmentation masks without dense supervision.}
\label{tab:RES}
\end{wraptable}

\paragraph{Generalization to dense prediction} Moreover, demonstrates that interpolating well-aligned path-text affinities naturally results in segmentation masks. Table~\ref{tab:RES} reports pixelwise-IoU between our interpolated masks and ground-truth masks on Refcoco/+/g. We have doubled the performance of TSEG~\citep{TSEG}, the previous state-of-the-art method on Referring Expression Segmentation \textit{without} training on ground-truth masks. Note, performing a second forward pass with zoomed-in regions leads to significantly finer mask contours (Figure~\ref{fig:REC} Right). We leave a closer investigation of this inference trick to future work.

\begin{wraptable}[17]{r}{0.3\linewidth}
\vspace{-10pt}
\resizebox{1.0\linewidth}{!}{
\begin{tabular}{@{}l@{\hspace{5pt}} c@{\hspace{5pt}} c@{\hspace{10pt}} c@{} }
\toprule
Model & Text & Image & Group \\  
\midrule
MTurk Human & 89.50 & 88.50 & 85.50 \\
Random Chance & 25.00 & 25.00 & 16.67 \\
\midrule
UNITER$\mathrm{_{Base}}$ & 32.25 & 13.25 & 10.00 \\
VILLA$\mathrm{_{Base}}$ & 30.00 & 12.00 & 8.00 \\
ViLT$\mathrm{_{Base}}$ & 34.75 & 14.00 & 9.25 \\
CLIP$\mathrm{_{Base}}$ & 30.75 & 10.50 & 8.00 \\
FLAVA-ITM$\mathrm{_{Base}}$ & 32.25 & \textbf{20.50} & 14.25 \\
\textbf{VLC$\mathrm{_{Base}}$} & 28.00 & 19.75 & 12.50 \\
\midrule
UNITER$\mathrm{_{Large}}$ & \textbf{38.00} & 14.00 & 10.50 \\
VILLA$\mathrm{_{Large}}$ & 37.00 & 13.25 & 11.00 \\
\textbf{VLC$\mathrm{_{Large}}$} & 32.00 & 20.00 & \textbf{14.75} \\
\bottomrule
\end{tabular}}

\caption{Winoground is a challenging test-only set for visio-linguistic compositional reasoning. \acronymlarge{} is competitive among similar-sized models without a second-stage pretraining.}

\label{tab:Winoground}
\end{wraptable}

\paragraph{Better Grounding translates to compositional reasoning} We find that the finetuned patch-text affinities translate to greater compositional reasoning performance. We report inference-only results on Winoground~\citep{Winoground} with our Refcoco-finetuned model in Table~\ref{tab:Winoground}. Winoground requires pairing up two sets of images and sentences with minimally contrastive semantics. Instead of directly predicting a pairing logit using the image-text matching head, it is more effective to measure image-sentence association via grounding success conditioned on an input sentence. Concretely, by treating the max affinity as the image-sentence-matching score, our Refcoco-finetuned model achieves a state-of-the-art Group Score and outstanding Image Scores on Winoground.\footnote{Please refer to the Winoground~\citep{Winoground} paper for how their evaluation metrics are defined.} This superior reasoning performance again verifies that our pretrained representations are more capable of using text/images to disambiguate each other.

\subsection{Ablations}

\begin{wrapfigure}[14]{r}{0.35\textwidth}
\vspace{-20pt}
\centering
  \includegraphics[width=1.0\linewidth]{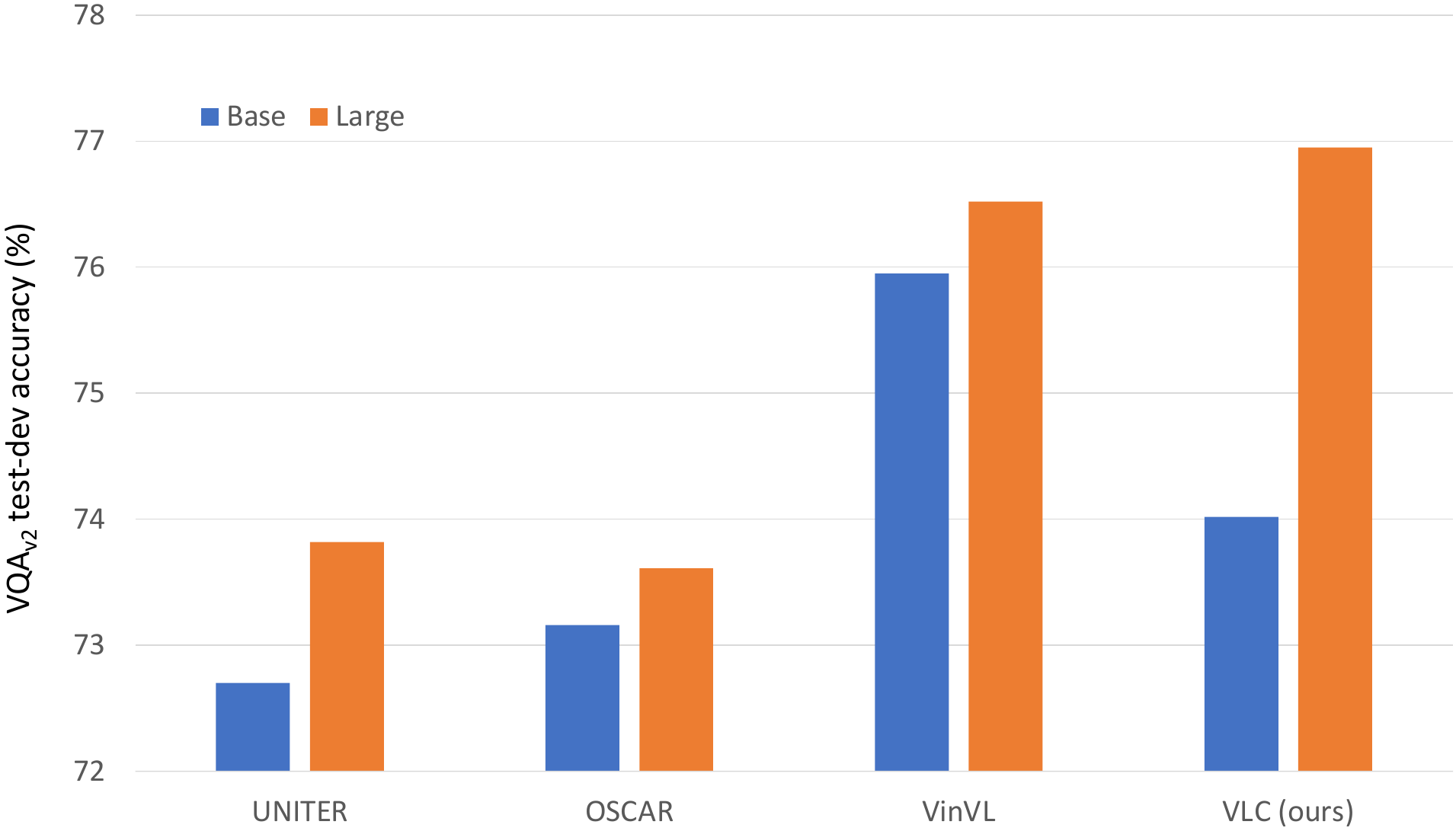}
  \caption{Comparison of the benefit of model scaling. While models initialized with supervised checkpoints inherit strong priors, \acronym{} exhibits the most substantial improvement when scaling the model size.}
\label{fig:modelsize}
\end{wrapfigure}


\paragraph{Initialization+Scaling}In Figure~\ref{fig:modelsize}, we compare VQAv2 test-dev accuracy between \acronym{} with three baselines that use pretrained object detectors and BERT for initialization. We see that while supervised initialization provides strong priors for the VQA task (\emph{e.g.}, VinVL), scaling the model size only has a marginal improvement. As a comparison, our model is initialized with MAE pretrained on ImageNet1K without labels. There is a substantial gain when scaling to a larger model.

\begin{figure}[t]
\begin{tabular}{@{}cc@{}}
\includegraphics[height=0.45\linewidth]{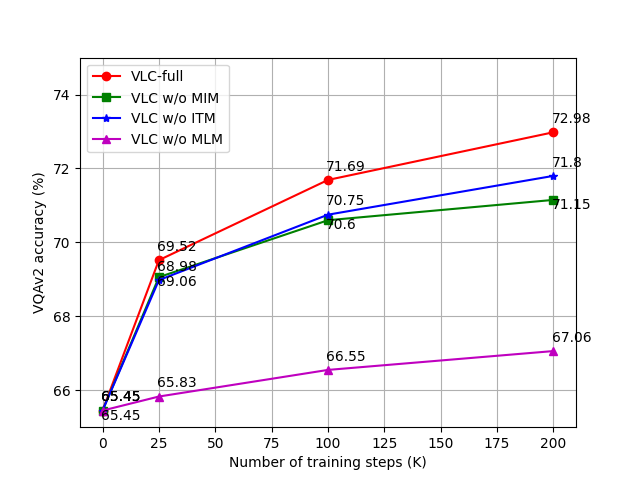} &
\includegraphics[trim={.1cm .65cm 1cm 1.75cm},clip, height=0.4\linewidth]{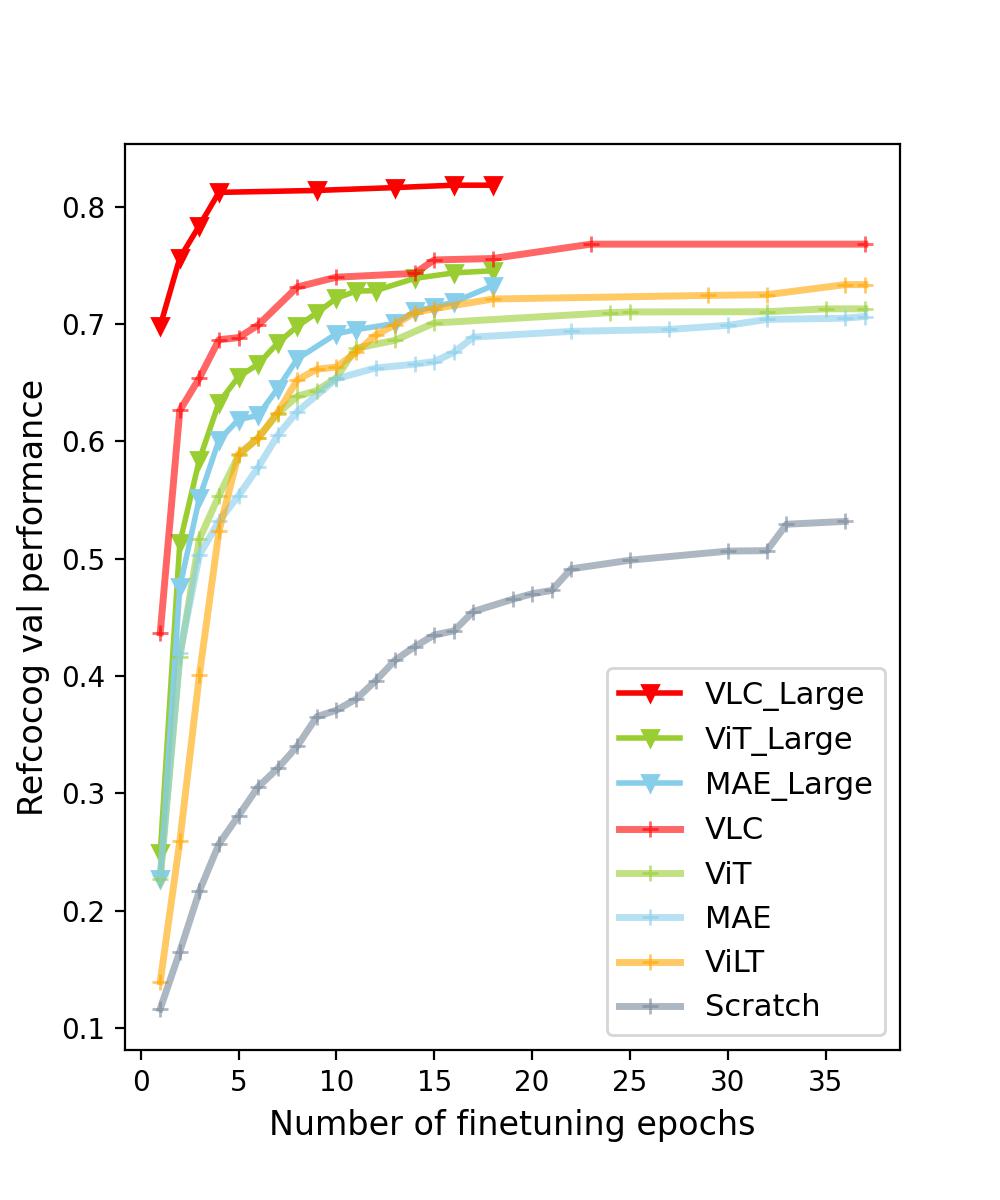}\\
\end{tabular}
\caption{\textbf{Left:} Role of objective functions. Our experiments show that \acronymbase{} sees gains from including MIM and that increased  training steps continues to improve VQAv2 performance. \textbf{Right:} Comparing the effects of different pretrained V+L checkpoints on downstream adaptation. Our experiments show that the pretrained \acronym{}
 representations enable faster finetuning and better downstream performance.}
\label{fig:ablation}
\end{figure}

\paragraph{Pretraining Objective} We ablate different combinations of objectives and train \acronymbase{} with $4$M image-text pairs. Figure~\ref{fig:ablation} (left) shows that, as the training steps increase, there is a consistent improvement for \acronym{} with MIM. This contrasts to findings in previous work~\citep{dou2021meter,kim2021vilt,li2021align}. We compare \acronymbase{} with different initializations and the benefit of unimodal pretraining in \ref{sec:init-pretrain}.

\paragraph{Downstream Adaptation} 

Since we finetune for 30 epochs on the grounding task vs. 10 epochs on the retrieval or understanding tasks, we would like to understand whether the grounding capability is simply a result of intensive finetuning. To this end, we experiment with our \acronym{} checkpoint as well as other three pretrained checkpoints: ViT~\citep{dosovitskiy2020image}, ViLT~\citep{kim2021vilt} and MAE~\citep{he2021masked}. We follow the same finetuning and inference procedures for all checkpoints. Figure~\ref{fig:ablation} (right) plots validation accuracy curves against the number of finetuning epochs. \acronym{} exhibits a considerable advantage right from the beginning, and maintains a clear margin until convergence. The relative advantage of \acronym{} becomes even greater when we compare the \emph{Large} model variants. This result validates that our proposed pretraining scheme provides a better foundation for finetuning, easing the computation burden of downstream adaptation.

\makeatletter
\g@addto@macro{\endtabular}{\rowfont{}}
\makeatother
\newcommand{\rowfonttype}{}
\newcommand{\rowfont}[1]{
\gdef\rowfonttype{#1}#1\ignorespaces%
}
\makeatother

\definecolor{Gray}{gray}{0.9}

\begin{table}[!thb]
\begin{minipage}{.6\linewidth}
\renewcommand{\arraystretch}{1.2}
\renewcommand{\tabcolsep}{1.2mm}
\centering

\footnotesize
\resizebox{1\linewidth}{!}{
\begin{tabular}{@{\hspace{2pt}}>{\rowfonttype}l@{\hspace{2pt}}>{\rowfonttype}l@{\hspace{4pt}}
rrrrr@{}
}

 & Model & Time & Param & \multicolumn{3}{@{}c@{}}{Refcoco/+/g(umd) val} \\
\toprule
\multirow{6}{*}{\rotatebox{90}{Modular}} 
 & VL-T5~\citep{VL-T5} & \phantom{0}5.9x & 290M & --- & --- & 71.2 \\
 & TransVG~\citep{TransVG} & \phantom{0}2.8x & 169M & 80.83 & 68.00 & 68.71 \\
 & TransVG++~\citep{TransVG++} & \phantom{0}--- & --- & 86.28 & 75.39 & 76.18 \\
 & QRNet~\citep{QRNet} & \phantom{0}5.5x & 273M &  84.01 & 72.94 &  73.03 \\
 & RefTrans~\citep{ReferringTrans} & \phantom{0}--- & --- & 85.65 & 77.55 & 79.25 \\
 & SeqTR~\citep{SeqTR} & \phantom{0}4.4x & 108M & 83.72 & 71.45 & 74.86 \\
 & UNITER$\mathrm{_{Large}}$~\citep{chen2020uniter} & \phantom{0}--- & 371M & 81.41  & 75.90 & 74.86 \\
 & VILLA$\mathrm{_{Large}}$~\citep{VILLA} & \phantom{0}--- & 371M & 82.39 & 76.17 & 76.18 \\
 \midrule
 & \acronymbase{} (ours) & \phantom{0}1.0x & 110M & 86.67 & 77.44 & 80.43 \\
 \midrule
 \midrule

\multirow{7}{*}{\rotatebox{90}{Unified}} 
 & OFA$\mathrm{_{Medium}}$~\citep{OFA} & \phantom{0}--- & 93M & 85.34 & 76.09 & 78.76 \\
 & Mdetr-R101~\citep{Mdetr} & \phantom{0}3.0x & 185M & 86.75 & 79.52 & 81.64 \\
 & Mdetr-ENB3~\citep{Mdetr} & \phantom{0}2.6x & 153M & 87.51 & 81.13 & 83.35 \\
 & {\color{gray}UniTAB~\citep{UniTAB}} & {\color{gray}19.4x} & {\color{gray}198M} & {\color{gray}88.59} & {\color{gray}80.97} & {\color{gray}84.58} \\
 & {\color{gray}OFA$\mathrm{_{Base}}$~\citep{OFA}} & {\color{gray}\phantom{0}6.0x} & {\color{gray}180M} & {\color{gray}88.48} & {\color{gray}81.39} & {\color{gray}82.29} \\
 & {\color{gray}OFA$\mathrm{_{Large}}$~\citep{OFA}} & {\color{gray}11.5x} & {\color{gray}470M} & {\color{gray}90.05} & {\color{gray}85.80} & {\color{gray}85.89} \\
 & {\color{gray}{OFA~\citep{OFA}}} & {\color{gray}{\phantom{0}---}} & {\color{gray}{930M}} & {\color{gray}{92.04}} & {\color{gray}{87.86}} & {\color{gray}{88.07}} \\
\bottomrule
\end{tabular}}
\caption{Referring Expression Comprehension organized by:  efficiency-performance and -versatility. \emph{Modularized} baselines 
include bespoke components in their architecture. \emph{Unified} baselines emphasize versatility over efficiency, and are pretrained with 
bounding box annotations. Our model 
is much lighter and faster. Params counts \textit{all} parameters required for inference, including RoI-proposals, the architecture, and output heads (see \S \ref{sup:grounding_analysis}).
}
\label{tb:REC}

\end{minipage}
~
\begin{minipage}{.38\linewidth}
\vspace{-75pt}
\renewcommand{\arraystretch}{1.2}
\renewcommand{\tabcolsep}{1.2mm}
\centering
\resizebox{1.0\linewidth}{!}{
\begin{tabular}
{@{\hspace{4pt}}l@{\hspace{6pt}} lc@{\hspace{4pt}} c@{}}
&       & \multicolumn{2}{c}{Top-1}\\ 
& Model &       $\mathrm{Base}$  &       $\mathrm{Large}$\\ 

\toprule
\multirow{3}{*}{\rotatebox{90}{\footnotesize \emph{Supervised}}}
& ViT-B/16~\citep{dosovitskiy2020image}  & 77.9 & 76.5 \\
& DeiT$\mathrm{_{Base}}$~\citep{touvron2021training}  &  83.1 & -\\
& Swin$\mathrm{_{Base}}$~\citep{liu2021swin}  & \bf 84.5  & -\\

\midrule
\multirow{5}{*}{\rotatebox{90}{\footnotesize \emph{Self-supervised}}} 
& DINO~\citep{caron2021emerging}    & 82.8 & -\\
& MoCo v3~\citep{chen2021empirical}  & 83.2 & 84.1\\
& MaskFeat~\citep{wei2021masked}     & 83.6 & 85.7\\
& SimMIM~\citep{xie2021simmim}       & 83.8 & 85.4\\
& BEiT$^*$~\citep{bao2021beit}       &  \bf 84.6 & 85.2\\
\midrule
& MAE~\citep{he2021masked}           & 83.6 & 85.9\\
& \textbf{\acronym{}} (ours)         & \bf 84.5 & \bf 86.3\\
\bottomrule
\end{tabular} 
}
\caption{Our pretraining does not compromise the quality of image representations, as evidenced by our superior results on ImageNet classification. \emph{Supervised} models are pretrained on ImageNet 1K and \emph{self-supervised} models are evaluated by end-to-end fine-tuning. $^*$BEiT uses a DALLE pretrained tokenizer.}
\label{exp:classification}
\end{minipage}
\end{table}

\vspace{-.1cm}
\section{Understanding the models}
\acronym{} is designed in importantly different ways from traditional bounding-box based approaches. In particular, while the visual stack is traditionally frozen in those models, now the entire ``backbone" is learnable. Also, where previously, the goal was to ``map" vision to language, now the two are learned jointly. We therefore take this opportunity to investigate the models to better understand how their behaviors differ due to the two key design choices. For a fair comparison with ViLT, we use \acronymbase{} which is trained with the same model architecture and image-text pairs.

\begin{figure}[t]
\begin{tabular}{@{}ccc@{}}
\includegraphics[width=0.3\linewidth]{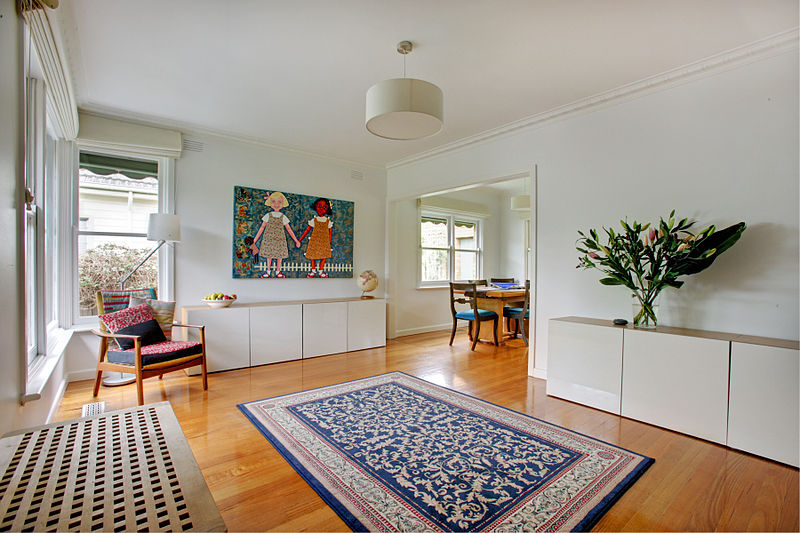} & 
\includegraphics[width=0.3\linewidth]{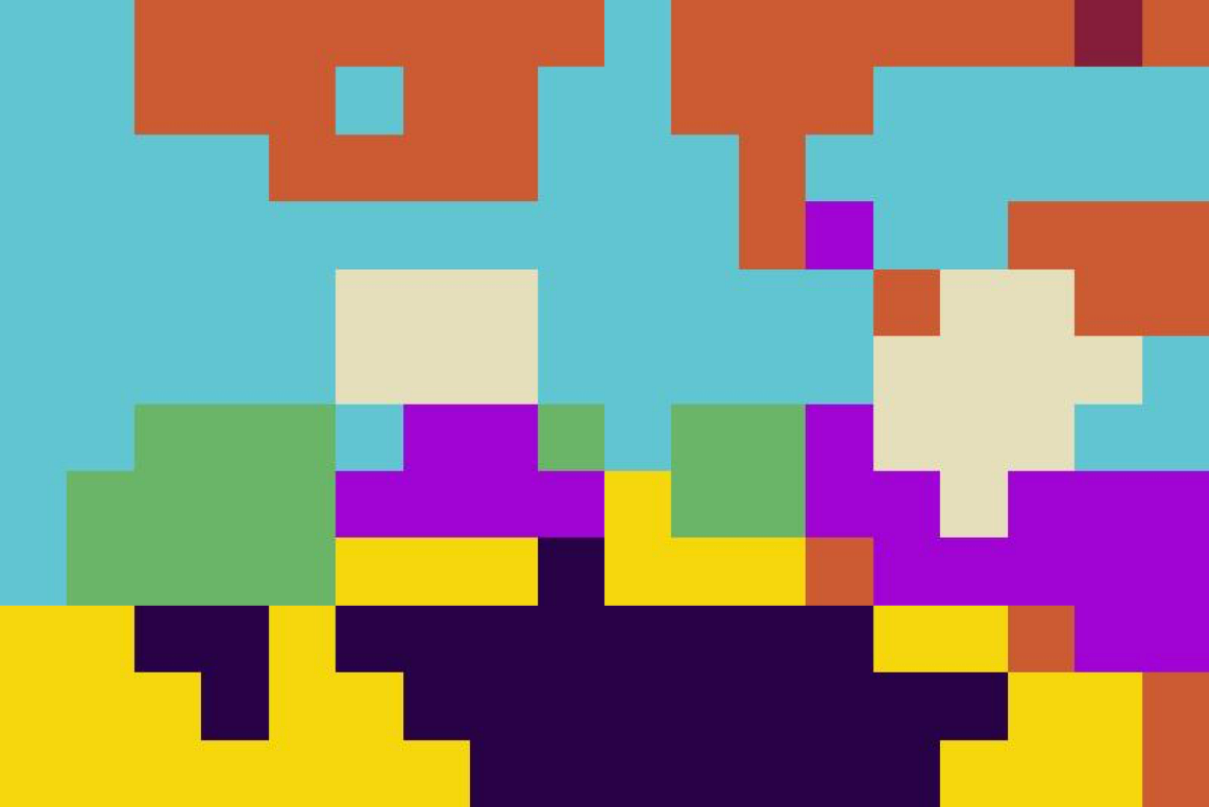} &
\includegraphics[width=0.3\linewidth]{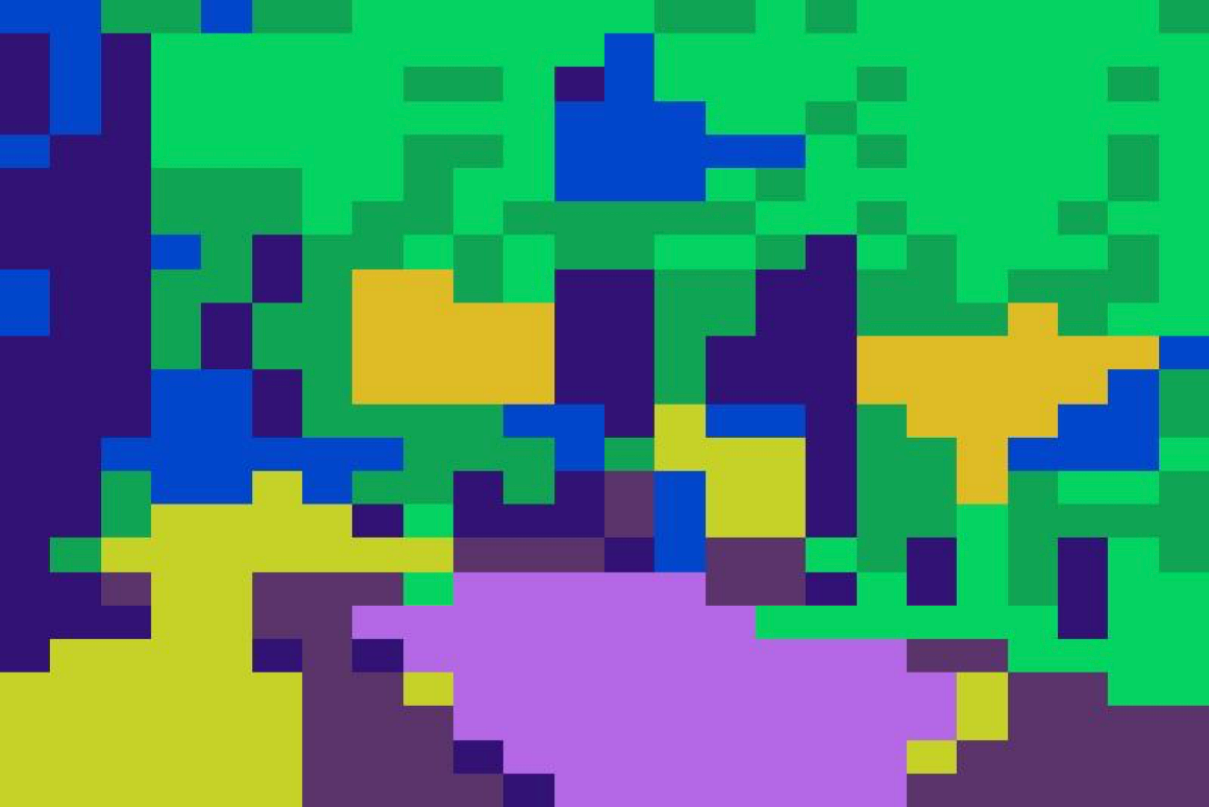} \\
Original Image & ViLT clusters & \acronym{} clusters \\
\end{tabular}
\caption{Visualization of patch clusters for an example image as produced from ViLT (many densely clustered patches) versus \acronym's more fine-grained and diffuse representations. We believe this representational difference makes for easier and faster learning and scaling -- akin to ``fast mapping" in language acquisition.}
\label{fig:clusters}
\vspace{-.4cm}
\end{figure}

\paragraph{Understanding Patches.} We begin with a simple patch clustering visualization (Figure \ref{fig:clusters}). Without the inclusion of any language, we can simply cluster (and color) the visual patch embeddings of ViLT and \acronym. ViLT relies on on larger patches ($32\!\times\!32$) for higher resolution ($384\!\times\!640$). We instead use smaller patches and lower resolution ($16\!\times\!16$ for $384\!\times\!384$). 
It is easy to see how both models are identifying key semantic regions of the image (e.g. the rug, painting and plant). Also note, both models incorrectly place the painting and plant in the same cluster. 

\begin{figure}[h]
    \centering
    \includegraphics[width=\linewidth]{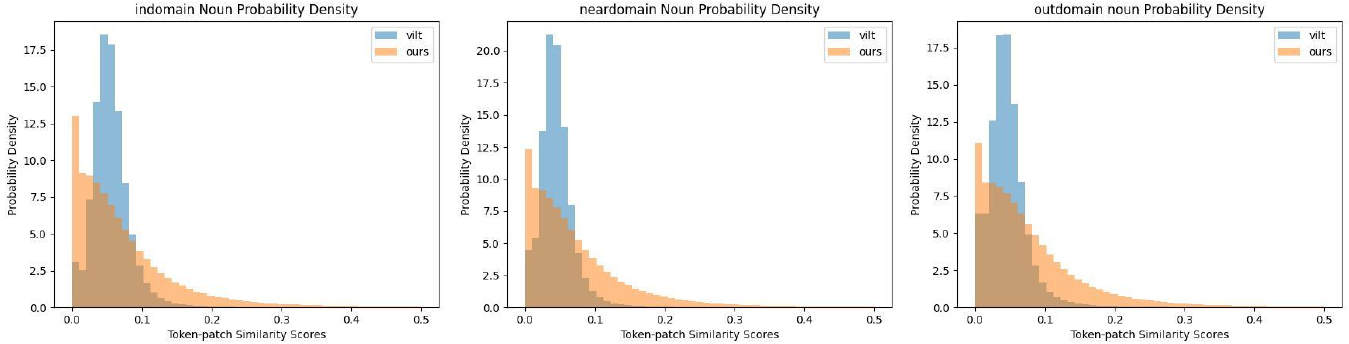}
    \caption{We plot the distribution of patch-noun similarities to compare the representation quality between ViLT and \acronymbase{}. ViLT rarely represents patches in great proximity to nouns, likely due to its discriminative pretraining objective. VilT also tend do produce weaker alignment for images dissimilar to ImageNet images. In contrast, \acronymbase{} produces patch representations strongly-aligned to corresponding nouns and does not show a preference towards ImageNet images.}
    \label{fig:nocaps}
\end{figure}

We leverage the nocaps dataset \citep{agrawal2019nocaps} to investigate representation diversity at scale. Nocaps provides captions for images from three buckets: inside COCO (in-domain), similar to COCO (near-domain), and outside COCO (out-domain). We compute representation similarities between patches and nouns for these three image buckets separately, and seek to answer the following questions: (1) Are ViLT patches more tightly clustered -- perhaps due to the discriminative training objective, and (2) How do both models' representation diversity change for images more (or less) like ImageNet images. In Figure \ref{fig:nocaps}, we see several trends.  First, ViLT's ``most similar" patch to the noun rarely has a similarity greater than 0.1, indicating that information initially coming from two modalities are not well integrated into each other's representations.  Second, we see the mass of patch-noun similarity distributions for ViLT slightly shift towards zero as we move from in-domain to out-domain, indicating that ViLT's difficulty in finding alignments to novel words. \acronym{} has a significantly smoother distribution of patch-noun similarities. \acronym{} often represents a nontrivial set of patches largely close to nouns across all image buckets. Also, the centers of similarity mass do not shift as much as those in ViLT plots as we move to out-domain images. A caveat is that these plots do \textit{not} justify the semantic meaningfulness of the alignment, but they do show starkly different behaviors between ViLT and \acronym{}. Our findings are supported by more visualized examples in \ref{sup:examples}.

\paragraph{Image Classification.}

Given that the underlying visual representations are shifting through the cross-modal training, we run a simple image classification experiment to see the effects language training has on the underlying visual ``backbone". We compare \acronym{} with state-of-the-art models on ImageNet-1K classification and report top-$1$ validation accuracy of a single $384\times384$ crop. 

As shown in Table~\ref{exp:classification}, \acronym{} learns generic representations which are transferable to ImageNet classification. After finetuning on ImageNet-1K, our model matches the performance of Swin$\mathrm{_{Base}}$~\citep{liu2021swin} that is trained with supervised labels. Note that BEiT~\citep{bao2021beit} is a two-stage pretrained model of which the image tokenizer is trained on $250$M examples of DALLE~\citep{ramesh2021zero} data. Compared with MAE~\citep{he2021masked}, our model learns competitive multi-modal representations from vision-language pretraining while retains high-quality image representations.

\begin{wraptable}{r}{0.5\linewidth}
\centering
\vspace{-10pt}
\renewcommand{\arraystretch}{1.2}
\renewcommand{\tabcolsep}{1.2mm}
\resizebox{1.0\linewidth}{!}{
\begin{tabular}{l@{\hspace{15pt}} c@{\hspace{5pt}} c@{\hspace{5pt}} c@{\hspace{0pt}}}
\toprule
Model & Model Size & MC-test & DA-test \\
\midrule
Random & 0 & 25.36 & 0.06 \\ 
\midrule
\multicolumn{4}{@{}l}{Large-scale Pretrained model} \\
   BERT~\citep{Devlin2019} & 110M & 33.54 & \phantom{0}8.41 \\
   GPT-3~\citep{brown2020language}& 175B & 35.21 & 11.49 \\
   ResNet~\citep{he2016deep}& 23M & 28.81 & \phantom{0}2.30 \\
   CLIP~\citep{radford2021learning} & 150M & \textbf{51.01} & \phantom{0}7.10 \\
\midrule
\multicolumn{4}{@{}l}{Specialized model}\\
   ViLBERT~\citep{lu2019vilbert} & 274M & 41.5\phantom{0} & 25.9\phantom{0} \\
   LXMERT~\citep{tan2019lxmert} & 240M & 41.6\phantom{0} & 25.9\phantom{0} \\
   KRISP~\citep{marino2021krisp} & 300M & 42.2\phantom{0} & \underline{27.1}\phantom{0} \\
   \acronymbase{} (ours) & 195M & \underline{44.82} & \textbf{27.49} \\ 
\cline{2-4}
   {\color{gray}{ClipCap$^{\dag}$~\citep{mokady2021clipcap}}} & {\color{gray}{930M}} &{\color{gray}{51.43}} &{\color{gray}{25.90}} \\
   {\color{gray}{GPV-2$^{\ddag}$~\citep{kamath2022webly}}}& {\color{gray}{380M}} & {\color{gray}{53.7\phantom{0}}} & {\color{gray}{40.7\phantom{0}}} \\
\bottomrule
\end{tabular}
}
\caption{Results on A-OKVQA under both Multiple-Choice (MC) and Direct Answer (DA) settings. $^{\dag}$ClipCap uses pretrained CLIP + GPT2$\mathrm{_{Large}}$. $^{\ddag}$GPV2 uses pretrained VinVL + T5$\mathrm{_{Base}}$ and learns a large number of concepts with Bing data. We report accuracy (\%) returned from the evaluation server.}
\label{exp:aokvqa}
\end{wraptable}

\paragraph{Evaluation on Open-domain VQA.} To investigate if the alignments between image patches and text tokens are semantically meaningful, we evaluate our \acronym{} on A-OKVQA dataset~\citep{schwenk2022okvqa}. A-OKVQA is more demanding than VQA~\citep{goyal2017making} for it requires commonsense reasoning about the scene depicted in the image. In \emph{multiple choice} (MC) setting, a model chooses its answer from one of four options. In the \emph{direct answer} (DA) setting, a model can generate any text as its answer, which better applies to real-world scenarios. We use \acronym{} as the mulitmodal encoder and a pretrained BERT to generate answers. In Table~\ref{exp:aokvqa}, we compare \acronym{} with large-scale pretrained \textbf{discriminative} models - BERT~\citep{Devlin2019} and ResNet~\citep{he2016deep}, \textbf{contrastive} model - CLIP~\citep{radford2021learning}, \textbf{generative} model - GPT3~\citep{brown2020language}, and models \textbf{specialized} on open-domain VQA. While CLIP stands out in the MC setting, it performs worse than other baselines in the DA setting. KRISP~\citep{marino2021krisp} ensembles different pretrained image classification and object detection models to exact image features. As a comparison, our \acronym{} outperforms KRISP in both settings. This implies that our model provides richer and language-aware image features which are conducive to open-vocabulary text generation. Note that ClipCap~\citep{mokady2021clipcap} and GPV2~\citep{kamath2022webly} use either much more data for pretraining or finetuning on the open-domain VQA task.

\clearpage
\newcommand\imgwidth{0.2\linewidth}
\begin{wrapfigure}{r}{0.5\linewidth}
  \centering
  \resizebox{1.0\linewidth}{!}{
  \begin{tabular}{@{}ccc@{}}
        \toprule
        \multicolumn{3}{@{}l@{}}{Caption with \textcolor{red}{focus}} \\[5pt]
        Original Image & 
        ViLT &
        \acronym{} 
        \\
        \midrule
        \multicolumn{3}{@{}l@{}}{A person on a beach holding a kite {\color{red}{string}} and a kite is in the air}\\[5pt]
        \raisebox{-.5\height}{\includegraphics[width=\imgwidth{}]{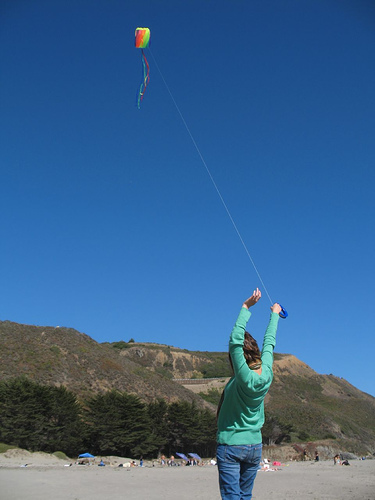}}
        &
        \raisebox{-.5\height}{\fbox{\includegraphics[width=\imgwidth{}]{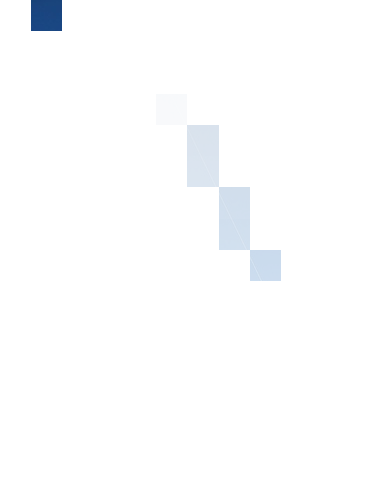}}}
        &
        \raisebox{-.5\height}{\fbox{\includegraphics[width=\imgwidth{}]{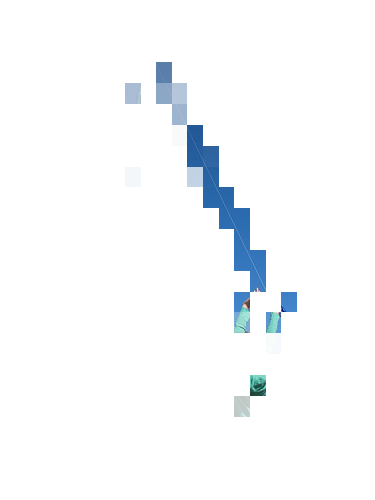}}}
        \\
        \midrule
        \multicolumn{3}{@{}l@{}}{A cat sitting on a chair, that is blue and {\color{red}{yellow}}}\\[5pt]
        
        \raisebox{-.5\height}{\includegraphics[width=\imgwidth{}]{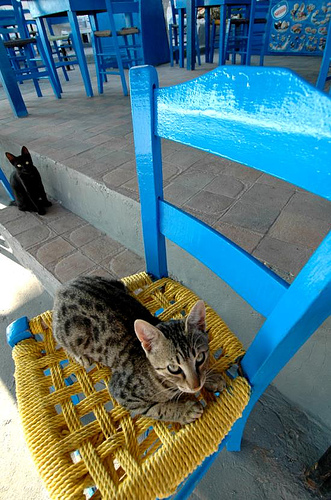}}
        &
        \raisebox{-.5\height}{\fbox{\includegraphics[width=\imgwidth{}]{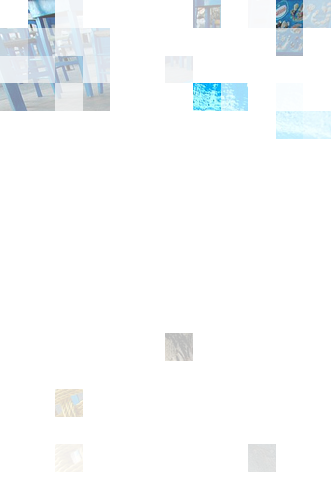}}}
        &
        \raisebox{-.5\height}{\fbox{\includegraphics[width=\imgwidth{}]{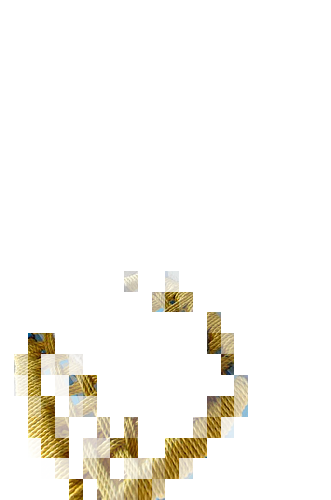}}}
        \\
        \midrule
       \multicolumn{3}{@{}l@{}}{A baseball player {\color{red}{swinging}} a baseball bat at a baseball}\\[5pt]
        \raisebox{-.5\height}{\includegraphics[width=\imgwidth{}]{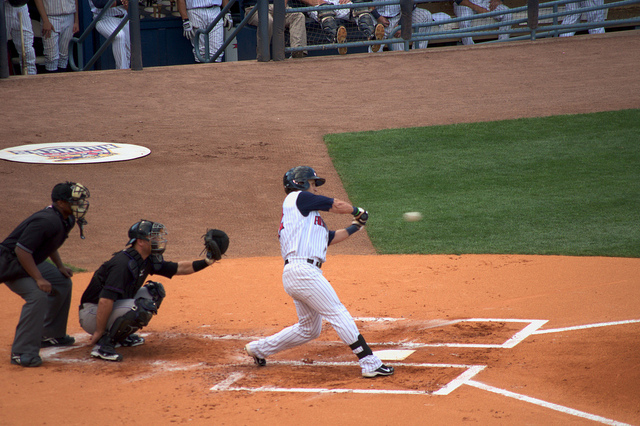}}
        &
        \raisebox{-.5\height}{\fbox{\includegraphics[width=\imgwidth{}]{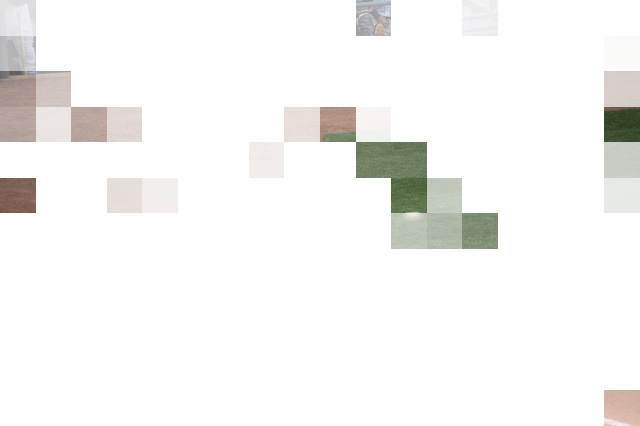}}}
        &
        \raisebox{-.5\height}{\fbox{\includegraphics[width=\imgwidth{}]{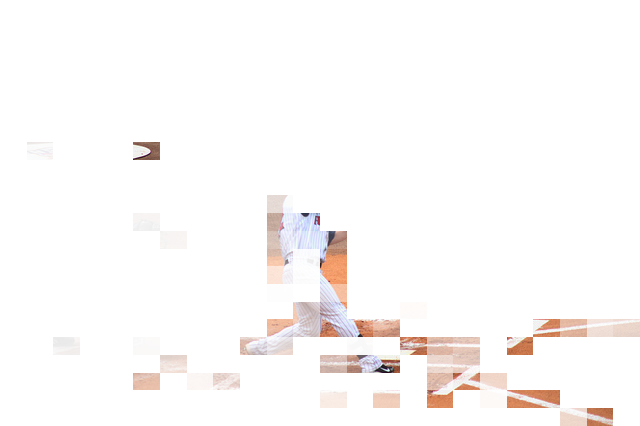}}}
        \\
\bottomrule
\end{tabular}}
\caption{Our model well handles out-of-domain concepts and images (i.e. not in ImageNet-1K). We visualize patch-similarities with a noun (top), an adjective (middle) and a verb (bottom), respectively. The model  delicately avoids nearby but distinct concepts (e.g. the cat on the chair or irrelevant parts of the baseball field). More examples and analysis can be found in \ref{sup:examples}.}
\label{fig:examples}
\vspace{-0.4cm}
\end{wrapfigure}

\section{Visualizations}
\begin{figure}[t]
    \centering
  \resizebox{0.8\linewidth}{!}{
  \begin{tabular}{cccc}
        \multicolumn{4}{c}{The bird is on the branch with leaves alone} \\[5pt]
        \includegraphics[width=0.2\linewidth]{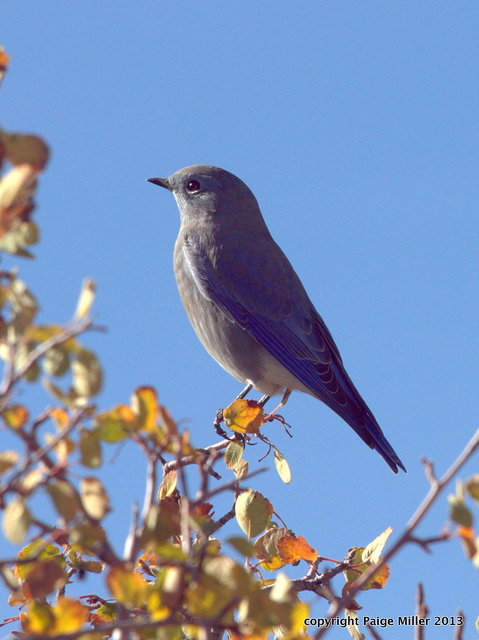}
        &
        \fbox{\includegraphics[width=0.2\linewidth]{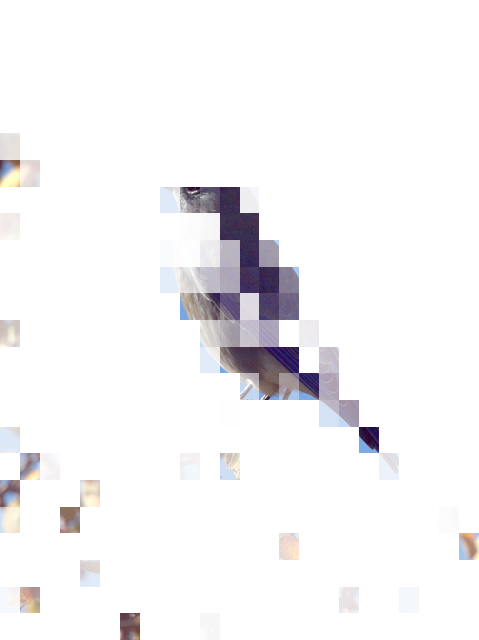}}
        &
        \fbox{\includegraphics[width=0.2\linewidth]{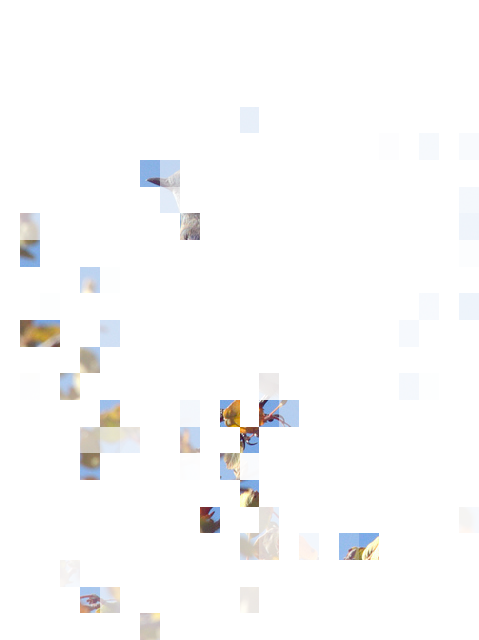}}
        &
        \fbox{\includegraphics[width=0.2\linewidth]{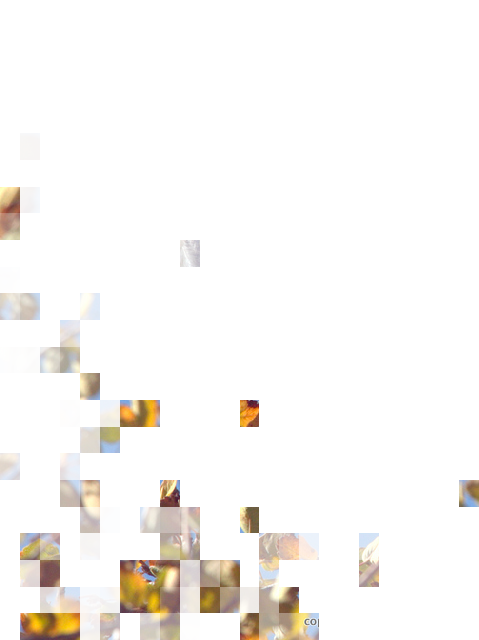}}\\
        & 
        bird & 
        branch & 
        leaves \\
    \end{tabular}}
    \caption{Finer-grained patch-token alignments is implicitly learned from weakly-aligned pretraining data. We visualize patch-similarities with different tokens from the same caption to see how our model uniquely represents them. It is interesting to note that our model attempts to differentiate branches from leaves.}
    \label{fig:main-visualizations}
    \vspace{-0.4cm}
\end{figure}

Patch-based transformer architectures allow for intuitive visualizations of the lexical alignment. We first demonstrate that faithful alignment corresponding to entity names or attributes already exist in the pretrained representations. Note, there is no explicit supervision on fine-grained patch-token alignment, yet the model learns to do so from weakly-aligned data. In Figure \ref{fig:main-visualizations}, we show results from visualizing three different words in the same caption for an image from COCO.  Note that for the word branch, the model is actively attempting to avoid the abundant leaves.  Second, since there is nothing about our model besides the MAE initialization that should be biased (as shown previously) towards ImageNet classes, we present three images in Figure \ref{fig:examples} that highlight words not present in the standard ImageNet1K training split used by other models. Specifically, a noun (\textit{string}), adjective (\textit{yellow}), and verb (\textit{swinging}). These demonstrate the general trend of ViLT often focusing on surprising locations.

Moreover, Section~\ref{sec:grounding} shows that finetuning can further incentivize the affinity patterns to behave like bounding boxes, which is the standard output format for localization. We visualize predicted alignment after finetuning in Table.~\ref{tb:viz_positive}. We highlight a greater reasoning ability beyond recognition in terms of disambiguating visual entities based on how they are referenced linguistically.

\definecolor{Salmon}{HTML}{F69289}
\definecolor{Green}{HTML}{00A64F}

\begin{table*}[t]
    \centering
    \small
    \begin{tabular}{@{}p{0.23\textwidth}p{0.23\textwidth}p{0.23\textwidth}p{0.23\textwidth}@{}}
    \toprule
    Input image & Predicted affinity map & Predicted bbox reduced & \textcolor{Salmon}{Predicted} \& \textcolor{Green}{Ground-truth} \\
    \midrule
    \multicolumn{4}{@{}c@{}}{\includegraphics[width=0.8\linewidth]{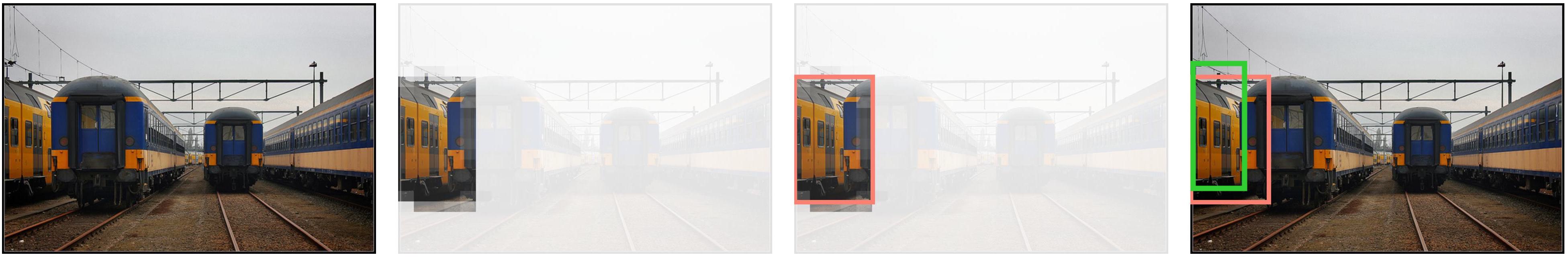}} \\
    \textbf{the leftmost train} & \multicolumn{3}{@{}r@{}}{The model succeeded at distinguishing same-type instances via absolute position.}\\
    \multicolumn{4}{@{}c@{}}{\includegraphics[width=0.8\linewidth]{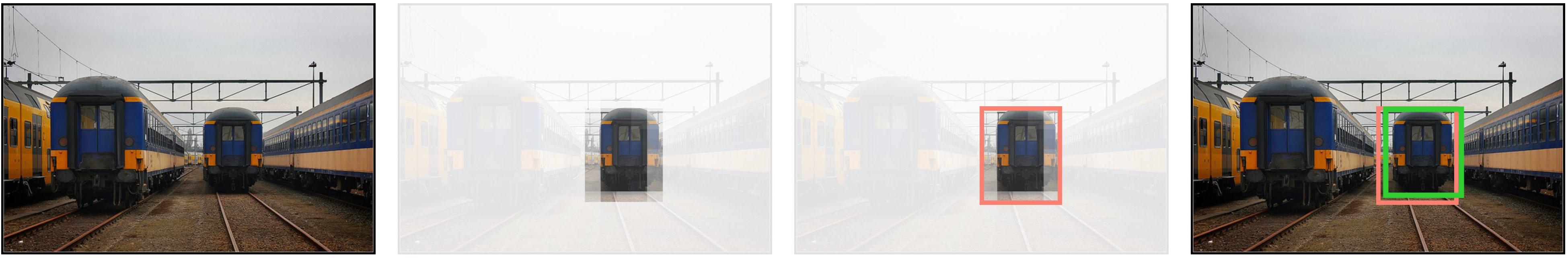}} \\
    \textbf{the smallest train} & \multicolumn{3}{@{}r@{}}{The model succeeded at distinguishing same-type instances via absolute size.} \\
    \bottomrule
    \end{tabular}
    \caption{Patch-token affinities can be finetuned towards localizing objects specified by referring expressions.}
    \label{tb:viz_positive}
    \vspace{-0.5cm}
\end{table*}

\section{Conclusion}
We present \longname{} (\acronym{}), a generic vision-language model pretrained with \emph{only} image-caption pairs. It uses a single linear layer to project raw pixel stimuli and token embeddings into the same representation space, followed by Transformer blocks jointly modeling two modalities. By removing the dependency on image region proposals, our model is both (1) more data efficient, for it does not require pretraining-scale class labels or bounding box annotation, and (2) faster at inference, for it does not require a tedious vision-only branch. 

Despite being lighter and faster, \acronym{} performs competitively on a diverse set of vision-language tasks, as compared to existing approaches relying on detection or ImageNet supervision. The MIM pretraining objective encourages richer and language-aware visual representations, which implicitly results in finer-grained alignment between patch and token representations. With moderate downstream finetuning, \acronym{} can be easily adapted to (1) perform multi-modal retrieval, (2) answer questions, (3) reason about visual information guided by free-form language, or (4) ground linguistically-referenced objects into bounding boxes. \acronym{}'s strong performance across nine downstream benchmarks clearly demonstrate the wide task applicability of a unified vision-language encoder. As performance scales with increased training data, this opens an exciting avenue for large-scale weakly-supervised open-domain vision-language models.

\bibliography{tmlr}
\bibliographystyle{tmlr}

\clearpage
\appendix
\part{} 
\addcontentsline{toc}{section}{Appendix} 
\part{Appendix} 
\parttoc 

\fboxsep=0pt
\setcounter{table}{0}
\renewcommand{\thetable}{A\arabic{table}}
\setcounter{figure}{0}
\renewcommand{\thefigure}{A\arabic{figure}}


\section*{Appendix}
This supplementary material has seven sections. Section~\ref{sup:dataset} describes the details of our pretraining datasets. Section~\ref{sup:Grounding_Inference} measures the inference time of different models and highlights the efficiency of our model. Section~\ref{sup:implementation1} and \ref{sup:implementation2} describe implementation details for downstream tasks. Section~\ref{sup:examples} shows more visualization examples with more comparisons. 
Section~\ref{sup:grounding_analysis} includes additional results and error analysis of \acronym{} finetuned on grounding tasks.
Section~\ref{sup:limitation} discusses the limitations and societal impact of this work.

\section{Pre-training dataset}
\label{sup:dataset}
The statistics of the pretraining dataset is shown in Table~\ref{exp:dataset}. Most of the existing approaches, such as UNITER~\cite{chen2020uniter} and ViLT~\cite{kim2021vilt}, use MSCOCO, VG, GCC and SBU to pre-train their models. We denote this training set as \emph{base}. To verify the scalability of our model, we follow VinVL~\cite{zhang2021vinvl} to further incorporate VQA, VG-QA, GQA, Flickr30K and OpenImages. As there are some overlaps among VG, MSCOCO and VQA, we exclude all those training images that appear in the downstream tasks via URL matching.

\begin{table*}[!htb]
\small
\centering
\begin{tabular}{ccccc|ccccc}
Dataset & MSCOCO & VG & GCC & SBU & VGA & GQA & VG-QA & Flickr30K & OpenImages \\
\midrule
\# Images & 113K & 100K & 2.95M & 860K & 83K & 79K & 87K & 29K & 1.67M  \\
\# Text & 567K & 769K & 2.95M & 860K & 545K & 1026K& 931K & 145K & 1.67M 
\end{tabular}
\caption{Statistics of the pre-training dataset}
\label{exp:dataset}
\end{table*}

\begin{figure}[t]
    \centering
    \includegraphics[width=\linewidth]{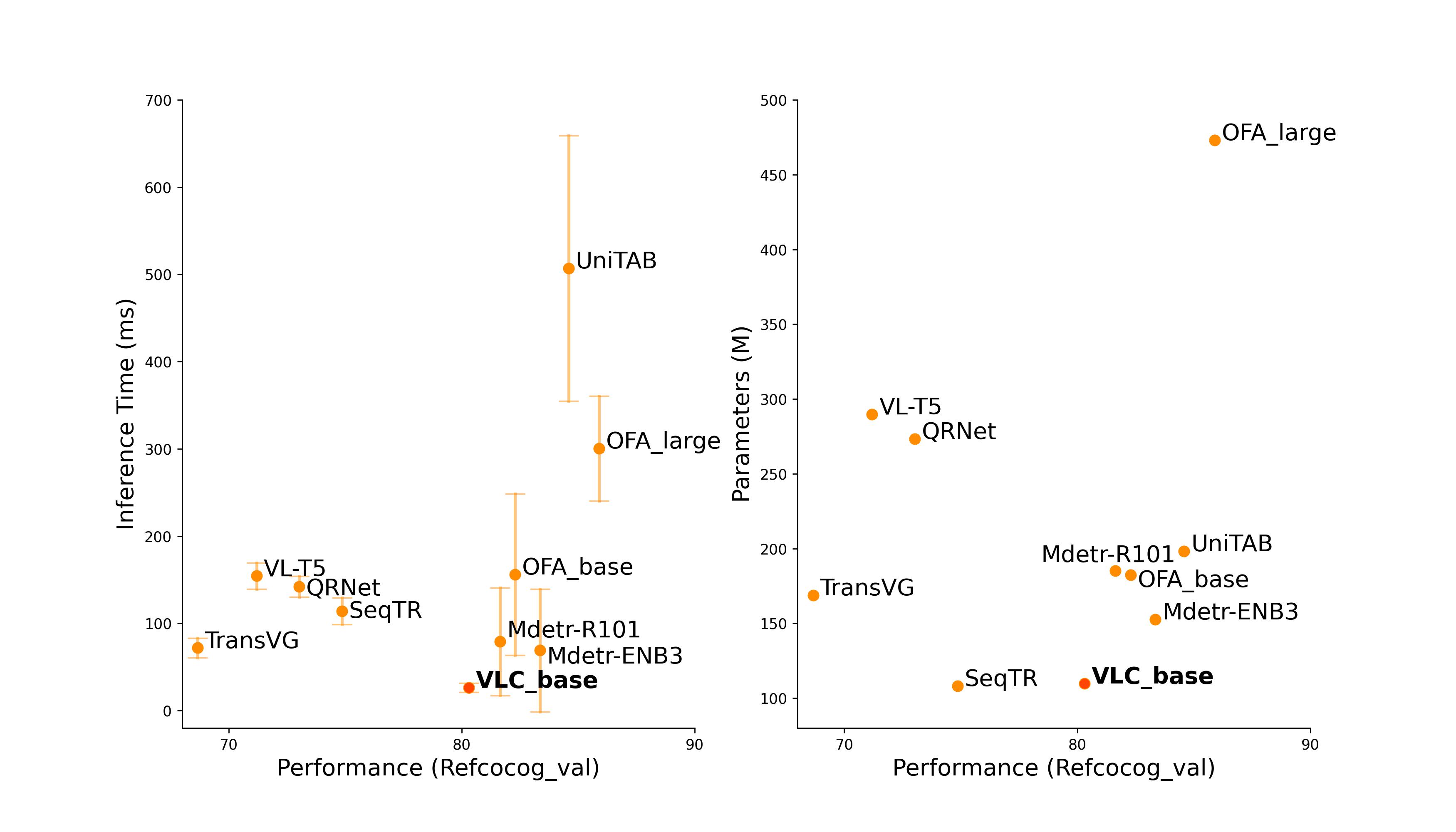}
    \caption{Inference time and parameter counts of different Referring Expression models. Our \acronymbase{} achieves competitive performance with considerably less inference time (left) and parameter count (right).}
    \vspace{-5pt}
    \label{fig:grounding_infrtime}
\end{figure}

Our extended VLC-for-grounding pipeline also strikes a superior efficiency-performance balance, with visualizations shown in Figure~\ref{fig:grounding_infrtime} where efficiency is indicate by both inference time and parameter count. Inference time is reported on 1 RTX2080Ti with batch\_size=1, averaged across the same 1K randomly selected samples from Refcocog\_val(umd) over 5 launches. Since different approaches require different post-processing for reducing an output Tensor to bbox coordinates, we time execution starting from feeding raw image and text tensors into the pipeline until the four-item coordinates tuple is obtained. Bespoke grounding models under comparison often involve components overspecialized for the end-goal of coordinate prediction, at the expense of directly optimizing for well-aligned V+L representations. Our results reveal that centralizing the expressive power in a single cross-modal encoder benefits V+L unification and efficiency without loss in performance.

\section{Implementation Details for Downstream Understanding \& Retrieval Tasks}

\label{sup:implementation1}
For downstream understanding and retrieval tasks, we fine-tune our model with a learning rate of $5e^{-4}$ for 10 epochs. We use $480\times 480$ as the input image resolution for the VQA task and $384\times 384$ for NLVR$^2$ and image-text retrieval tasks.
\paragraph{Visual Question Answering (VQA~\cite{goyal2017making}).} Given an input image and a question, the VQA task is to predict an answer from the visual content. We conduct experiments on VQAv2 dataset~\cite{goyal2017making} that is built on MSCOCO. It contains 83K images for training, 41K for validation, and 81K for testing. Following previous work~\cite{tan2019lxmert,chen2020uniter,li2021align}, we use the training, validation splits and additional question-answer pairs from Visual Genome while reserving $1,000$ validation image-question pairs for internal validation.  We report performance on the test-dev and test-std splits. We follow the standard practice~\cite{kim2021vilt} to convert the task to a multilabel classification task with $3,192$ answer classes. 
\paragraph{Natural Language for Visual Reasoning (NLVR$^2$~\cite{suhr2018corpus}).} Given a triplet of two images and a description, this task is to predict whether this description describes a pair of images. Following previous work~\cite{kim2021vilt,chen2020uniter}, we use the \emph{pair} method which treats one input sample as two image-text pairs by repeating the text twice. Each pair is passed through our model and we take the concatenation of two pooled representation \texttt{[CLS]} from our model as the representation of one input sample. 
Similar to the settings of the VQA task, we use a 2-layer MLP with a hidden size of $1,536$ to adapt \acronym{} to the NLVR$^2$ task.
\paragraph{Image-Text Retrieval.} Image-Text retrieval contains two subtasks: image-to-text retrieval (TR) and text-to-image retrieval (IR). We evaluate our pre-trained models on the Karpathy splits~\cite{karpathy2015deep} of MSCOCO~\cite{lin2014microsoft} and Flickr30K~\cite{plummer2015flickr30k} in fine-tuning settings. MSCOCO contains $123$K images, and each image has five corresponding human-written captions. We split the data into 82K/5K/5K training/validation/test images. To be consistent with previous work~\cite{kim2021vilt,chen2020uniter}, we use the additional 30K images from MSCOCO validation set to improve the performance. Flickr30K contains $31$K images with five captions for each image. We split the data into 30K/1K/1K as the training/validation/test set.

\section{Implementation Details for Downstream Grounding Tasks}
\label{sup:implementation2}

\newcommand{\cls}{\texttt{[CLS]}}

The main text has briefly argued that appropriate supervision enables generalizable and versatile representations which naturally translate to downstream grounding performance without tailored design. This section describes our approach of adapting the Transformer backbone to performing grounding tasks in detail. We conduct grounding experiments on Refcoco/+/g. Namely, Section~\ref{sup:Referring_Expression_Datasets} provides dataset details, Section~\ref{sup:Grounding_Finetuning} introduces the process of converting bounding box annotations to patch regions for supervised finetuning, and Section~\ref{sup:Grounding_Inference} describes how the resulting affinities can be re-interpreted as either a mask or a bounding box.

\subsection{Referring Expression Datasets}
\label{sup:Referring_Expression_Datasets}

The referring expression task feeds a model with an (image, region description) pair, then expects a localization output as either a bounding box (bbox) or a pixel-level mask corresponding to the description. It tests for visual grounding and reasoning as the model indicates which part of the image is referenced by the text. Note, a mask output is more challenging to produce since a mask can adopt various shapes and requires a dense prediction in the original image’s resolution. 

We use three Referring Expression datasets: Refcoco, Refcoco+, and Refcocog. All of which use MSCOCO \citep{lin2014microsoft} training images. Refcoco and Refcoco+ were collected from human annotators participating in a “ReferIt” Game, where one player selected regions in an image referenced by a text description provided by the other player. Where they differ is that Refcoco+ disallowed players from using location-based expressions (e.g. “person to the right”), thereby being purely appearance-based. Refcocog was collected in a non-game setting but contains more linguistically complex descriptions. 

Bounding boxes are evaluated by Intersection-over-Union (IoU), while masks are evaluated by pixel-level IoU. To compare with other relevant models, we follow the literature by reporting IoU\_acc, which is the proportion of testing samples on which the model achieves an IoU $>$ 0.5. 

\subsection{Finetuning}
\label{sup:Grounding_Finetuning}

Our approach builds on the notion of an affinity map.  Formally, we define the affinity map, $\mathcal{A}$, as the cosine distance between all $\mathcal{L}$anguage tokens and $\mathcal{V}$isual patches.
\begin{equation}
    \mathcal{\hat{A}}_{t,p} = \mathtt{cos(}\mathcal{L}_t, \mathcal{V}_p\mathtt{)}
    \label{eq:affinity}
\end{equation}
for $T$ tokens and $P$ patches.  This resulting matrix $[-1,1]^{T\times P}$ provides a normalized score for the relationship between every token and patch.

\paragraph{Supervision Signals} For supervision, we translate bounding box annotations to sets of patches and phrases simply correspond to their indices, with \cls{} being the initial index $0$. The ground-truth affinity scores can be obtained from labeled datasets where annotators selected bboxes corresponding to phrases in an (image, sentence) pair. Thus, for an (image, sentence) pair, we have a set of annotated (bbox, phrase) pairs, from which we would like to get a $T \times P$ matrix of affinity scores. Note that the full image has dimensions $H\times W$ and will be divided into a 
$\sqrt{P}\times\sqrt{P}$ grid. 

For a given  annotated (bbox, phrase) pair, if the lexical token $\mathcal{L}_t$ belongs to the phrase and patch $\mathcal{V}_p$ overlaps with the bbox, we set $\mathcal{A}_{t, p}$ in the ground-truth affinity map to be amount the patch overlaps with the bbox. 
\begin{equation}
    \mathcal{A}_{t, p} = \frac{|Patch\cap \mathrm{BBox}|}{|Patch|}
    \label{eq:overlap_ratio}
\end{equation}

If a bbox corresponds to the entire sentence, we set ground-truth affinity scores for the $0$-th token, \cls{}.

\paragraph{Hyperparameters} 

We opt for a higher resolution (384) and smaller patches (16) since the patch size will determine the granularity of our predictions. This yields a final 24$\times$24 grid. We finetune VLC on the combined Refcoco/+/g training sets for 50 epochs with AdamW \citep{loshchilov2017decoupled} optimizer and a $5e^{-4}$ learning rate. 

\paragraph{Finetuning Objective} 
We optimize our model with a binary-cross-entropy (BCE) loss between the ground-truth and predicted affinity scores. 
\begin{equation}
    Loss = \sum_{t\in T, p\in P} \mathrm{BCE}(\mathcal{A}_{t,p}, \mathcal{\hat{A}}_{t,p})
\end{equation}
We intentionally avoided the use of a softmax because normalization would force competition across patches and affect their individual judgments. Each token-patch pair relationship is independent during loss calculation. 

Note, that the domain expected by BCE is the full real line, but our affinities are computed via cosine similarities and are therefore bounded to [-1,1].  Empirically, we found that re-scaling the input by $k=2$ improved performance due to a better utilization of the [0,1] output range. However, performance decreases with a larger k because it tends to over-sharpen the sigmoid's decision boundary, leaving less ``wiggle room" for the raw logits. 

\subsection{Inference}
\label{sup:Grounding_Inference}

\begin{figure*}[t]
    \centering
    \includegraphics[width=0.8\linewidth]{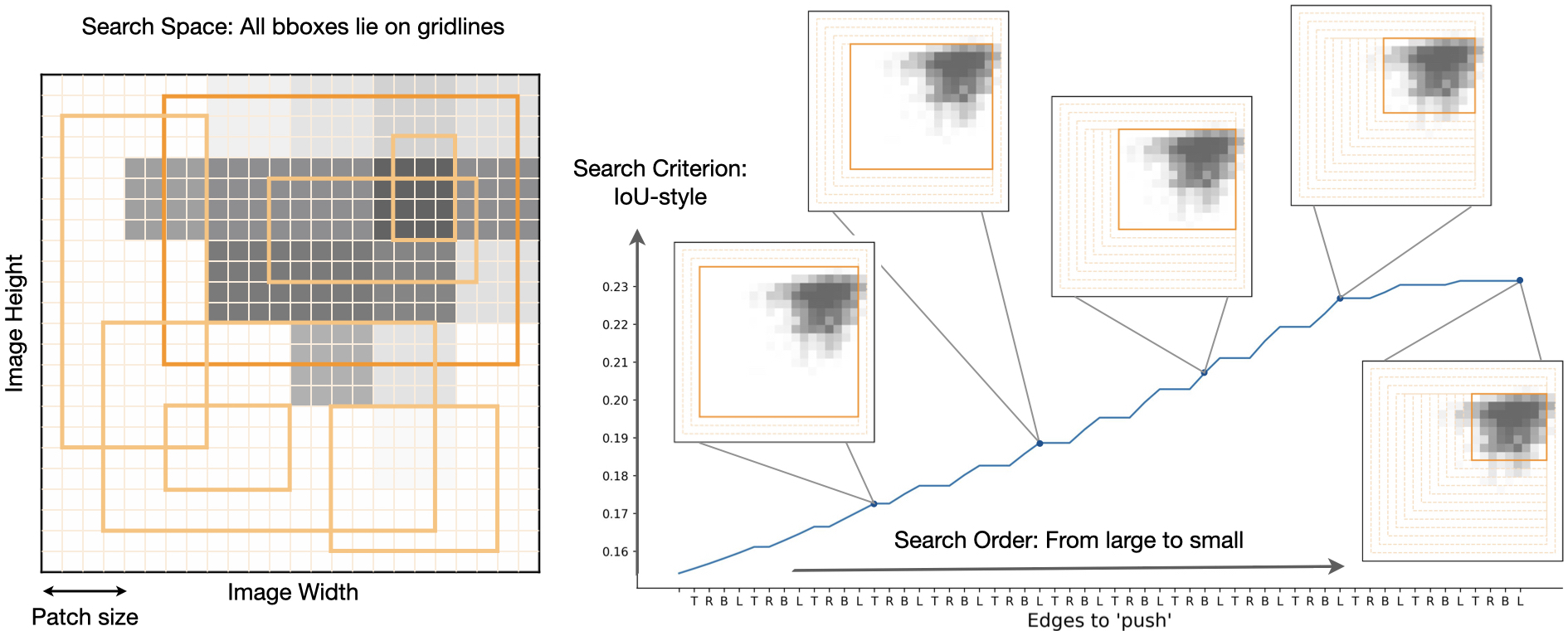}
    \caption{Inference procedure overview for ``Affinity Map $\Rightarrow$ BBox",  which is formulated as a search problem. Left: we define the search space to contain all candidate bboxes that lie on gridlines. Right: our algorithm greedily iterates over bboxes following the search order of large $\rightarrow$ small, and terminates once the search criterion no longer improves. }
    \vspace{-10pt}
    \label{fig:push_alg}
\end{figure*}

The forward pass through the Transformer outputs a single tensor of shape $L\!\times\!768$, or more accurately, the concatenation of two tensors of shape $T\!\times\!768$ and $P\!\times\!768$ for the encoded $\mathcal{L}$anguage and $\mathcal{V}$isual sequences, respectively. During inference, these matrices are used in Equation \ref{eq:affinity} to compute the predicted affinity matrix $\hat{\mathcal{A}}$. Next, we describe how a mask or a bounding box can be algorithmically derived from a predicted affinity matrix $\mathcal{\hat{A}}$. We emphasize that this procedure requires no additional learned parameters and is algorithmically efficient.

Since rows in $\mathcal{\hat{A}}_{actv}$ associated with a phrase can be pooled into a single row and reshaped into $\sqrt{P} \times \sqrt{P}$, we henceforth refer $\mathcal{\hat{A}}^r_{actv}$ to such a $[-1,1]^{\sqrt{P} \times \sqrt{P}}$ matrix with spatial correspondence to the image, albeit with a coarser resolution. Then, a binary mask can be simply derived by bilinearly interpolating $\mathcal{\hat{A}}^r_{actv}$ back to the original resolution, and binarizing the values with a threshold. 

The procedure for deriving a bbox from $\mathcal{\hat{A}}^r_{actv}$ is more complicated, which we detail below. \textbf{We formulate bbox prediction as a search problem}. A bbox is denoted by $\mathcal{B}=(x,\ y,\ w,\ h)$, satisfying $0\leq x\leq x+w \leq W$, $0\leq y\leq y+h \leq H$. The goal is to search for $\mathcal{B}$ given  $\mathcal{\hat{A}}^r_{actv}$, such that $\mathcal{B}$ best represents a rectangular region $\mathcal{\hat{A}}^r_{actv}$ tries to highlight (Figure \ref{fig:push_alg}, Left). To solve the search problem we need to decide on the following: the search space, the search order, and the search criterion.

\paragraph{Search space} Theoretically, we have an infinite number of boxes to search over which can be as large as the input image or as small as nearly a single point. We constrain our search to only those boxes whose edges lie on gridlines, with step  $t$ denoting the unit between two gridlines.

\paragraph{Search order} We start with a box encompassing the full image and greedily search for progressively smaller boxes.

\paragraph{Search criterion} 
We propose a satisfying criterion $\mathcal{M}$ that guarantees equivalence between $\mathcal{\hat{A}}^r_{actv}$ and the theoretically optimal bbox ($\mathcal{B}$). In the case where all values in $\mathcal{\hat{A}}^r_{actv}$ are binary, this reduces to calculating a standard IoU (\textit{Intersection-over-Union}). In practice, values will be real-valued. We overload the notation to handle the generalization from binary to real values in $\mathcal{\hat{A}}^r_{actv}$. 
Intersection is defined as the sum of the active values within $\mathcal{B}$.  Union is equal to the sum of three terms, namely the active values inside $\mathcal{B}$, the values that should have been active inside $\mathcal{B}$ and the active values outside $\mathcal{B}$. The first two terms in Union add up to $|\mathcal{B}|$. As a single equation this results in

\begin{equation}
    \mathcal{M} = \frac{|\mathcal{\hat{A}}^r_{actv} \cap \mathcal{B}|}{|B| + |\mathcal{\hat{A}}^r_{actv} \cap \neg \mathcal{B}|}
\end{equation}

In practice, this is an easy calculation, as the intersections and negations simply correspond to summing the values inside and outside of the candidate box.  Further, summing (in leiu of counting) trivially generalizes to the continuous valued activations.

We provide additional proof of the equivalence to optimizing F1 in the supplementary.  We will note, there is a potential hyperparameter which we will call $C$ that defines the desired ``tightness" of the box.
\footnote{For example, $C=0.5*|\mathcal{\hat{A}}^r_{actv}|$ would produce  bboxes systematically smaller than $C=|\mathcal{\hat{A}}^r_{actv}|$. The supplementary discusses about where $C$ is derived from. In this work, we leave the scaling to 1  and allow future work to analyze how $C$ would influence the returned value of Algorithm.\ref{alg:push}}

\begin{footnotesize}
\begin{algorithm}
\DontPrintSemicolon
\SetKwInOut{Input}{input}\SetKwInOut{Output}{output}
\caption{PUSH}\label{alg:push}
\Input{heatmap $\mathcal{\hat{A}}^r_{actv}$, image size $(W, H)$, step $t$, measure $\mathcal{M}: (\mathcal{\hat{A}}^r_{actv}, \mathcal{B})\rightarrow \mathcal{R}$, \\order $O \in Permutation([T, L, B, R])$}
\Output{$\mathcal{B}$: box coordinates $(x, y, w, h)\ s.t.$ $0\le x\le x+w \le W$ and $0\le y\le y+h \le H$}
\BlankLine
$(x, y, w, h) \gets (0, 0, W, H)$\\
$\mathcal{B} \gets (x, y, w, h)$\\
\While{$w>0$ and $h>0$ and move $\neq$ False}{
\For{$e\in O$}{
\lIf{$e == T$}{$\mathcal{B'} \gets (x, y+t, w, h-t)$}
\lElseIf{$e == B$}{$\mathcal{B'} \gets (x, y, w, h-t)$}
\lElseIf{$e == R$}{$\mathcal{B'} \gets (x, y, w-t, h)$}
\lElse{$\mathcal{B'} \gets (x+t, y, w-t, h)$}
\BlankLine
\If{$\mathcal{M}(\mathcal{\hat{A}}^r_{actv}, \mathcal{B'}) > \mathcal{M}(\mathcal{\hat{A}}^r_{actv}, \mathcal{B})$}{
    $\mathcal{B} \gets \mathcal{B'}$\\
    $(x, y, w, h) \gets \mathcal{B}$\\
    $move \gets True$}
}
}
\Return $\mathcal{B}$
\end{algorithm}
\end{footnotesize}

Next we leverage the the search space, order and criterion ($\mathcal{M}$) in Algorithm \ref{alg:push} to predict bounding boxes. Initializing $\mathcal{B}$ to surround the entire image, our \textit{PUSH} algorithm greedily iterates over progressively smaller $\mathcal{B}$ and terminates when $\mathcal{M}(\mathcal{\hat{A}}^r_{actv}, \mathcal{B})$ no longer improves. Within each iteration, there are ``push" attempts on the edges of $\mathcal{B}$, one at a time, by $t$-amount inwards according to an order  $\mathcal{O} \in Permutation([T, L, B, R])$. For example, $\mathcal{O} = [T, R,B, L]$ would translate to a clockwise progression of tightening edges from the $Top-Right-Bottom-Left$. A push operation is successful if and only if it increases $\mathcal{M}$. Otherwise, the edge is left unchanged. The algorithm terminates when it encounters unsuccessful push attempts on all four edges in a row. Figure \ref{fig:push_alg} visualizes a \textit{PUSH} execution. 

In our experiments, we use values $P = 24,\ W = H = 384$ and $t=\frac{1}{4}*patch\_size=4$.\footnote{Empirically, we find that the order of $[L, B, R, T]$ does not affect performance. We ran our bbox inference procedure on the Refcocog\_val(umd) 10 times, randomly permuting the order of $[L, B, R, T]$ at each \textit{PUSH} iteration. For 90\% samples the algorithm returned the same predicted bbox all 10 times. For the remaining samples, the 10 returned bboxes have a joint IoU=0.95$\pm$0.07.} 
A known limitation of our approach is highly discontiguous or concave affinity patterns.  We did not observe this condition in our setting. 
We provide empirical evidence that our ``Affinity Map $\Rightarrow$ BBox" formalization is effective but leave a mathematically rigorous investigation to future work.

\subsection{Additional Math Proof}

In the section above, we derive the search criterion as the Intersection-over-Union (IoU) between $\mathcal{\hat{A}}^r_{actv}$ and $\mathcal{B}$:

\begin{equation}
    \mathcal{M}^{IoU} = \frac{|\mathcal{\hat{A}}^r_{actv} \cap \mathcal{B}|}{|B| + |\mathcal{\hat{A}}^r_{actv} \cap \neg \mathcal{B}|}
\end{equation}

We will further argue that optimizing for IoU is equivalent to optimizing for F1\footnote{We ignore the scaling by 2 in the numerator of $\mathcal{M}^{F1}$ because a constant scalar will not affect the search procedure.} in the context of our \textit{PUSH} algorithm. Using the same notation, we write down F1 between $\mathcal{\hat{A}}^r_{actv}$ and $\mathcal{B}$ as follows:

\begin{equation}
\begin{aligned}
    \mathcal{M}^{F1} &= \frac{2|\mathcal{\hat{A}}^r_{actv} \cap \mathcal{B}|}{|B| + |\mathcal{\hat{A}}^r_{actv} \cap \mathcal{B}| + |\mathcal{\hat{A}}^r_{actv} \cap \neg \mathcal{B}|} \\
    &\equiv \frac{|\mathcal{\hat{A}}^r_{actv} \cap \mathcal{B}|}{|B| + |\mathcal{\hat{A}}^r_{actv}|} \\
\end{aligned}
\end{equation}

In order to prove the equivalence of the two criteria, $\mathcal{M}^{IoU}$ and $\mathcal{M}^{F1}$, we simply need to show that when a push operation causes one criterion to increase, the other criterion must also increase as well.

Assume a single push operation reduces $|\mathcal{\hat{A}}^r_{actv} \cap \mathcal{B}|$ from $a_1$ to $a_2$ and reduces $|B|$ from $b_1$ to $b_2$. Let $C=|\mathcal{\hat{A}}^r_{actv}|$, which holds constant throughout the search. Eq.\ref{Eq:equivalent_M} shows that $\mathcal{M}^{F1}$ increases iff $\mathcal{M}^{IoU}$ increases. 

\begin{equation}\label{Eq:equivalent_M}
    \begin{aligned}
        \quad&\frac{a_1}{b_1+C} < \frac{a_2}{b_2+C} \\
        \leftrightarrow \quad&a_1b_2+a_1C < a_2b_1+a_2C \\
        \leftrightarrow \quad&a_1b_2+a_1C-a_1a_2 < a_2b_1+a_2C-a_1a_2 \\
        \leftrightarrow \quad&\frac{a_1}{b_1+C-a_1} < \frac{a_2}{b_2+C-a_2} \\ 
    \end{aligned}
\end{equation}

\section{Different Initialization Methods \& Unimodal Pretraining}
\label{sec:init-pretrain}
\begin{figure}[!ht]
  \centering
  \includegraphics[width=0.6\linewidth]{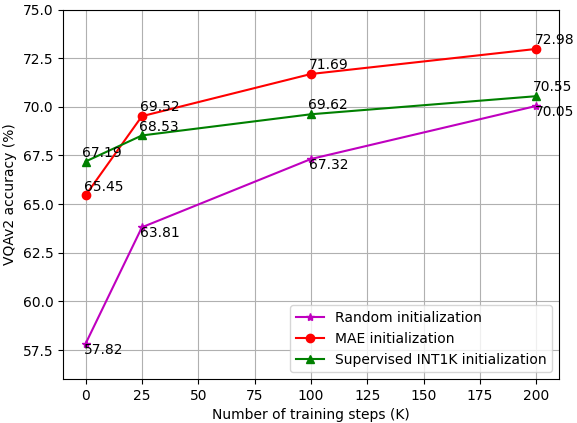}
  \caption{Comparisons of our \acronymbase{} with different initializations. We show that MAE initialization improves the training speed.}
  \label{fig:ini}
\end{figure}

The figure~\ref{fig:ini} compares \acronymbase{} with different initializations. We show that most of our performance gain comes from VL pretraining (VLP), while MAE initialization improves the training speed. \textbf{(1) Supervised INT1K.} It has a strong prior (67.19\% before VLP), but a flat learning curve (+3.36\% gain). We hypothesize that such initialization is more biased to predefined INT1K classes, and inconsistent with the visual embedding space learned by MIM. This makes the model difficult to learn multimodal representations by simply switching one objective to another. \textbf{(2) Random.} It has the most substantial improvement (+12.23\% gain), but still lags behind MAE initialization. We hypothesize that 200K steps (80 epochs) are not sufficient for images. As a comparison, MAE is trained with 1600 epochs on INT1K. More training may mitigate the gap. (3) Neither models, random nor MAE initialization are saturated.

Unimodal pretraining initializes the representation space, exposing or collapsing unique properties/clusters. We will expand this analysis, but also include three additional benefits. The table~\ref{tab:ini} compares unimodal pretraining approaches on VQA. (1) There is a general trend that better pretraining provides stronger downstream priors (METER with Swin/INT21K: 72.38\% before VLP), (2) Pretraining can improve the multimodal training speed. As the image/text modalities have different learning speeds. MAE and BERT require 1600 vs 40 epochs on INT1K and a text corpus, respectively. Better image training creates more meaningful representations which allows for ``fast mapping" in language acquisition (Fig. 5), and (3) Primary performance gains (+10.2\%) come from VLP which demonstrates the effectiveness of our model.
\begin{table}[!ht]
\vspace{-.3cm}
\centering
\renewcommand{\arraystretch}{0.95}
\resizebox{0.75\linewidth}{!}{
\begin{tabular}{c|c|c|c}
\toprule
Models    & \acronymlarge{} (Ours) & ViLT~\cite{kim2021vilt} & METER~\cite{dou2021meter} \\ 
\midrule 
Initialization & MAE  & ViT(21K)  & Swin(21K)/Roberta \\ 
\midrule
w/o VLP & 66.75  &67.19  &  72.38 \\
w/ VLP  & 76.95 (+10.2)  &  71.26 (+4.07) & 76.43 (+4.05) \\
\bottomrule
\end{tabular}}
\caption{Comparisons of the benefit of different unimodal pretraining on VQA. We show that most of our performance gain comes from VL pretraining.}
\label{tab:ini}
\end{table}

\section{Additional Visualizations}
\label{sup:examples}
\begin{figure*}[!htb]
    \centering
  \begin{tabular}{p{0.22\textwidth}ccc}
        Caption with \textcolor{red}{focus} &
        Original Image & 
        ViLT &
        \acronym{}
        \\
        \toprule
        A {\color{red}{hawk}} is perched on a metal bar
       &
        \raisebox{-.5\height}{\includegraphics[width=.2\linewidth]{supplementary/COCO_val2014_000000270705.jpg}}
        &
        \raisebox{-.5\height}{\fbox{\includegraphics[width=.2\linewidth]{supplementary/vilt_COCO_val2014_000000270705_hawk_2.png}}} 
                &
        \raisebox{-.5\height}{\fbox{\includegraphics[width=.2\linewidth]{supplementary/ours_COCO_val2014_000000270705_hawk_2.png}}}\\
        \midrule
        A gift wrapped with a {\color{red}{ribbon}} sits on a table with a knife
        &
        \raisebox{-.5\height}{\includegraphics[width=.2\linewidth]{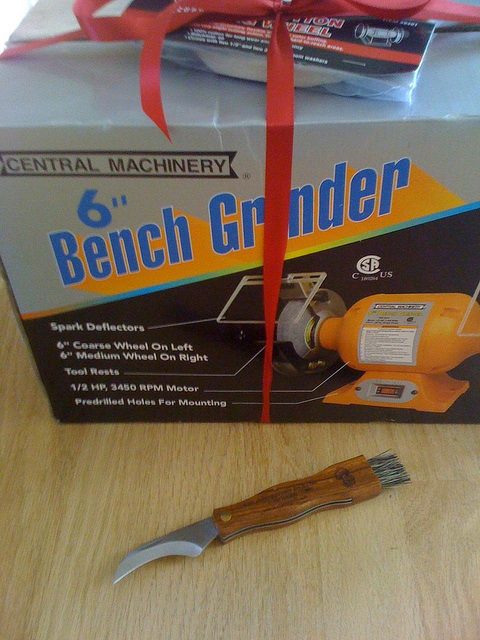}}
        &
        \raisebox{-.5\height}{\fbox{\includegraphics[width=.2\linewidth]{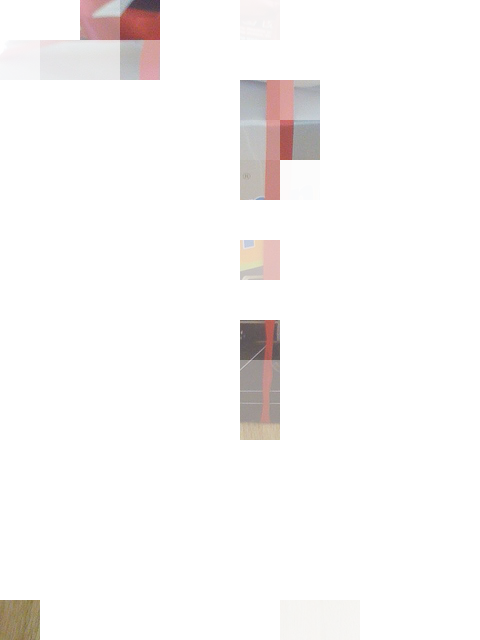}}}
                &
        \raisebox{-.5\height}{\fbox{\includegraphics[width=.2\linewidth]{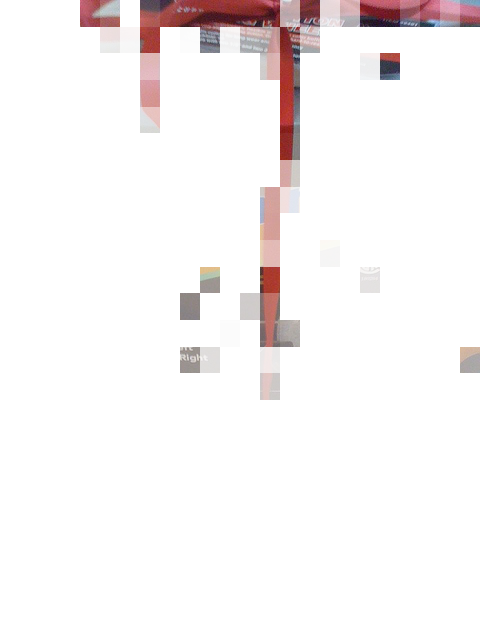}}}\\
        \midrule
        A plate with pancakes, {\color{red}{syrup}}, grits, and butter
        &
        \raisebox{-.5\height}{\includegraphics[width=.2\linewidth]{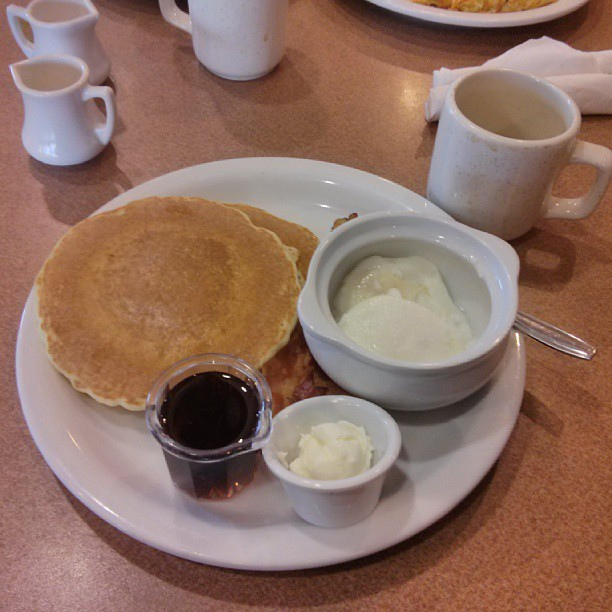}}
        &
        \raisebox{-.5\height}{\fbox{\includegraphics[width=.2\linewidth]{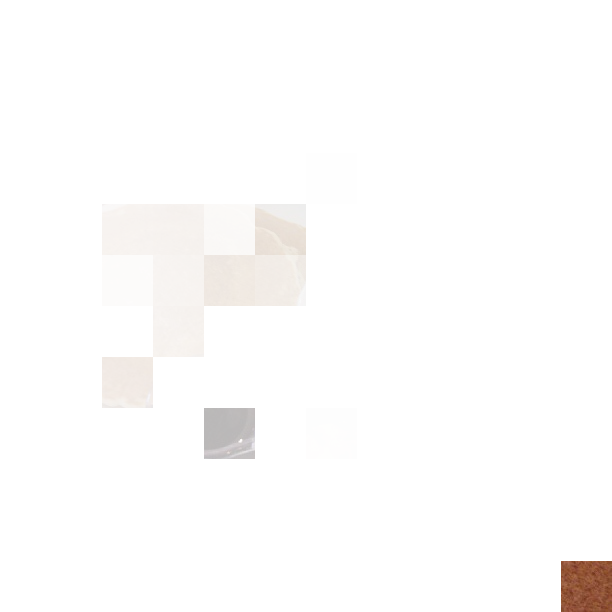}}}
                &
        \raisebox{-.5\height}{\fbox{\includegraphics[width=.2\linewidth]{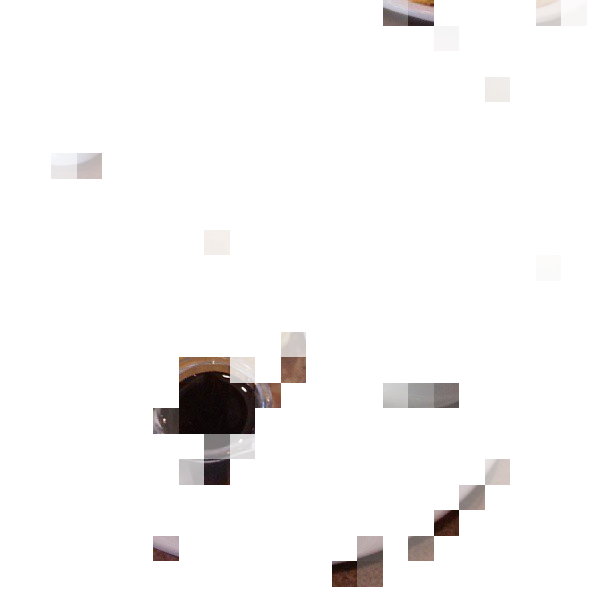}}}
        \\

        \midrule
        There is a colorful {\color{red}{parachute}} in the sky
        &
        \raisebox{-.5\height}{\includegraphics[width=.2\linewidth]{supplementary/COCO_val2014_000000007784.jpg}}
        &
        \raisebox{-.5\height}{\fbox{\includegraphics[width=.2\linewidth]{supplementary/vilt_COCO_val2014_000000007784_parachute_5.png}}}
                &
        \raisebox{-.5\height}{\fbox{\includegraphics[width=.2\linewidth]{supplementary/ours_COCO_val2014_000000007784_parachute_5.png}}}
        \\
\midrule

  \end{tabular}
\caption{We visualize patch similarities with nouns on OOD images.  Note that ViLT is often picking up on relevant features but has the strongest correlation with a single, presumably predictive patch.}
\label{fig:sup-noun-examples}
\end{figure*}

\begin{figure*}[!htb]
    \centering
  \begin{tabular}{p{0.22\textwidth}ccc}
        Caption with \textcolor{red}{focus} &
        Original Image & 
        ViLT & 
        \acronym{}
        \\
        \toprule
        A {\color{red}{red}} fire hydrant in front of a skyscraper
        &
        \raisebox{-.5\height}{\includegraphics[width=.2\linewidth]{supplementary/COCO_val2014_000000266082.jpg}}
        &
        \raisebox{-.5\height}{\fbox{\includegraphics[width=.2\linewidth]{supplementary/vilt_COCO_val2014_000000266082_red_2.png}}}  
        &
        \raisebox{-.5\height}{\fbox{\includegraphics[width=.2\linewidth]{supplementary/ours_COCO_val2014_000000266082_red_2.png}}}
        \\
        \midrule

        A monarch butterfly lands on a {\color{red}{pink}} flower.
        &
        \raisebox{-.5\height}{\includegraphics[width=.2\linewidth]{supplementary/0451aa5cf163feab.jpg}}
        &
        \raisebox{-.5\height}{\fbox{\includegraphics[width=.2\linewidth]{supplementary/vilt_0451aa5cf163feab_69_697_pink_7.png}}}
        &
        \raisebox{-.5\height}{\fbox{\includegraphics[width=.2\linewidth]{supplementary/ours_0451aa5cf163feab_69_697_pink_7.png}}}
        
        \\
        \midrule
        A {\color{red}{small}} orange and blue ladybug sitting on long green leaves
        &
        \raisebox{-.5\height}{\includegraphics[width=.2\linewidth]{supplementary/12f6ce6440f84315.jpg}}
        &
        \raisebox{-.5\height}{\fbox{\includegraphics[width=.2\linewidth]{supplementary/vilt_12f6ce6440f84315_359_3596_small_2.png}}}
        &
        \raisebox{-.5\height}{\fbox{\includegraphics[width=.2\linewidth]{supplementary/ours_12f6ce6440f84315_359_3596_small_2.png}}}
        
        \\
        \midrule
        A brown and white dog is holding a {\color{red}{yellow}} Frisbee
       &
        \raisebox{-.5\height}{\includegraphics[width=.2\linewidth]{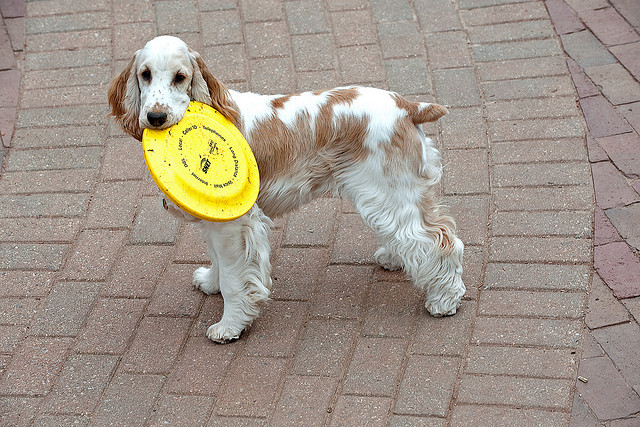}}
        &
        \raisebox{-.5\height}{\fbox{\includegraphics[width=.2\linewidth]{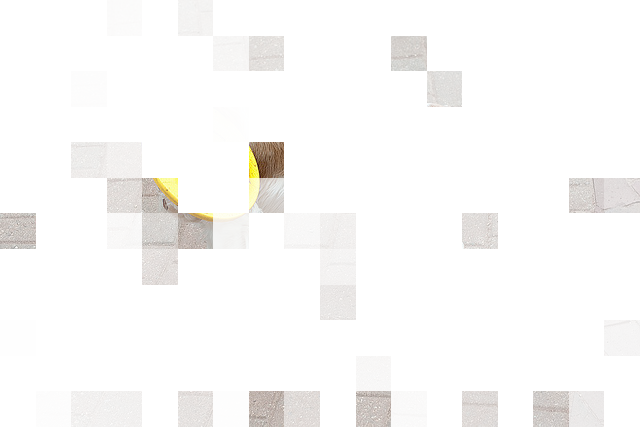}}} 
        &
        \raisebox{-.5\height}{\fbox{\includegraphics[width=.2\linewidth]{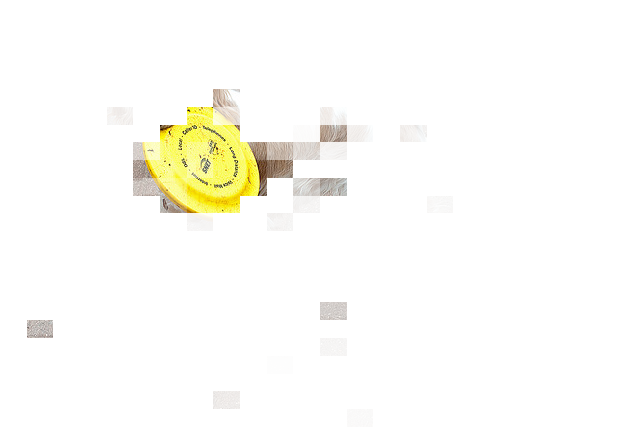}}}\\
        \midrule

  \end{tabular}
\caption{We visualize patch similarities with adjectives on OOD images. \acronym{} produces more accurate and comprehensive alignment. Note that the lady bug is correctly associated with the relative size \textit{small}. Future work would ideally investigate how models align visual entities to more abstract and comparative concepts in the language domain.}
\label{fig:sup-adj-examples}
\end{figure*}

\begin{figure*}[!htb]
  \begin{tabular}{p{0.22\textwidth}ccc}
        Caption with \textcolor{red}{focus} &
        Original Image & 
        ViLT &
         \acronym{}
        \\
        \toprule
        A person who is {\color{red}{hitting}} a ball with a bat.
        &
        \raisebox{-.5\height}{\includegraphics[width=.2\linewidth]{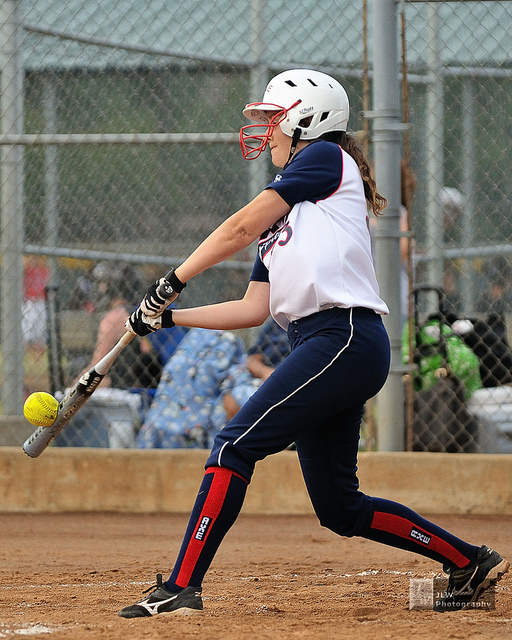}}
        &
        \raisebox{-.5\height}{\fbox{\includegraphics[width=.2\linewidth]{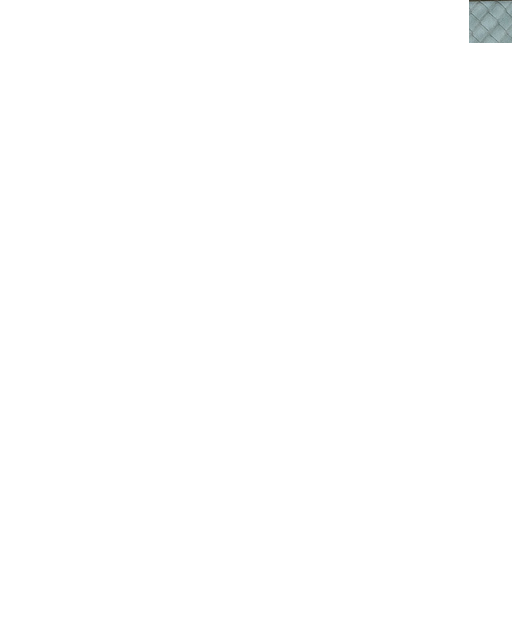}}}
                &
        \raisebox{-.5\height}{\fbox{\includegraphics[width=.2\linewidth]{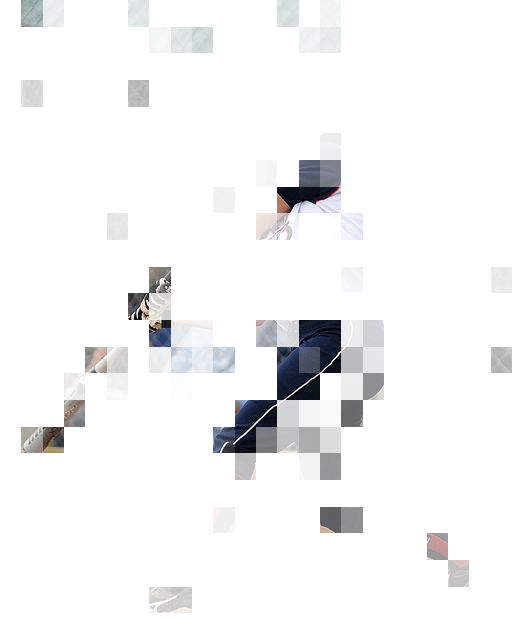}}}\\
        \midrule
        A person {\color{red}{holding}} a cell phone in their hand
        &
        \raisebox{-.5\height}{\includegraphics[width=.2\linewidth]{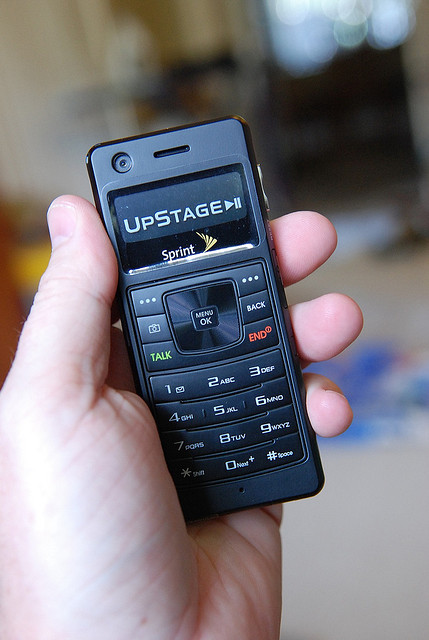}}
        &
        \raisebox{-.5\height}{\fbox{\includegraphics[width=.2\linewidth]{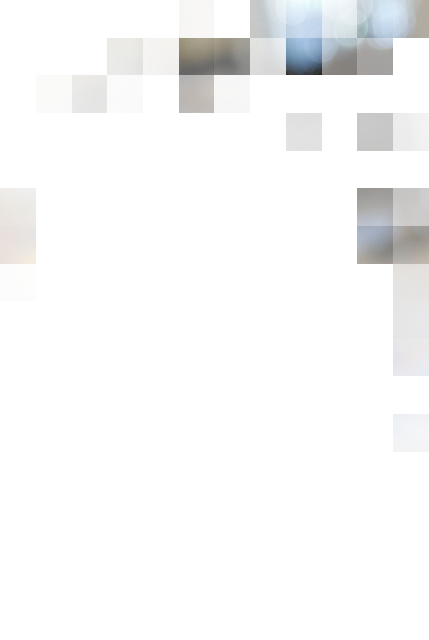}}}
                &
        \raisebox{-.5\height}{\fbox{\includegraphics[width=.2\linewidth]{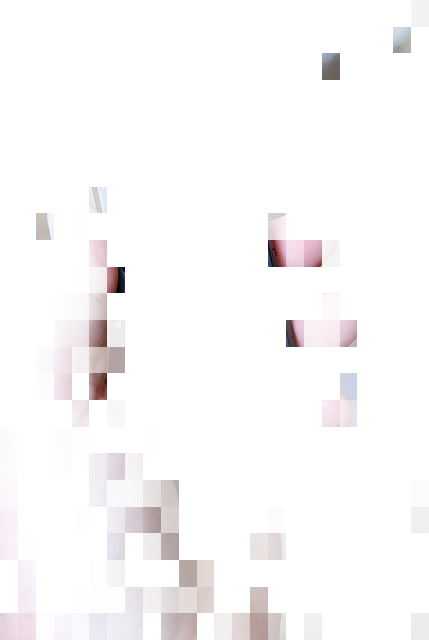}}}
        \\
        \midrule
        A green boat {\color{red}{floating}} on top of a body of water
       &
        \raisebox{-.5\height}{\includegraphics[width=.2\linewidth]{supplementary/COCO_val2014_000000311518.jpg}}
        &
        \raisebox{-.5\height}{\fbox{\includegraphics[width=.2\linewidth]{supplementary/vilt_COCO_val2014_000000311518_floating_4.png}}} 
                &
        \raisebox{-.5\height}{\fbox{\includegraphics[width=.2\linewidth]{supplementary/ours_COCO_val2014_000000311518_floating_4.png}}}\\
        \midrule
        an orange and white cat sitting on a bed {\color{red}{staring}} at the viewer
        &
        \raisebox{-.5\height}{\includegraphics[width=.2\linewidth]{supplementary/COCO_val2014_000000190690.jpg}}
        &
        \raisebox{-.5\height}{\fbox{\includegraphics[width=.2\linewidth]{supplementary/vilt_COCO_val2014_000000190690_staring_10.png}}}
                &
        \raisebox{-.5\height}{\fbox{\includegraphics[width=.2\linewidth]{supplementary/ours_COCO_val2014_000000190690_staring_10.png}}}
        \\
\midrule
  \end{tabular}
\caption{We visualize patch similarities with verbs on OOD images. Note that aligning verbs to static images is a slightly strange task, but our model still manages to associate key perceptual indicators to a verb's semantics. For example, \textit{holding} is aligned to a person's hand and \textit{staring} picks up on the cat's eyes.}
\label{fig:sup-verb-examples}
\end{figure*}

We provide additional visualizations for patch-token alignment resulted from pre-training. Specifically, we choose tokens to be nouns in Figure~\ref{fig:sup-noun-examples}, adjectives in Figure \ref{fig:sup-adj-examples}, and verbs in Figure~\ref{fig:sup-verb-examples}.

The inductive bias of organizing patch-token affinity patterns into bounding boxes can be correctly learned through finetuning, seamlessly adapting the pretrained model to localization tasks. However, in downstream applications when only a single object is expected to be aligned to a phrase (i.e. a span of tokens), model predictions are occasionally distracted by attributes or relations. Table \ref{tb:viz_vary_the_expression} uses the same input image to illustrate the model's sensitivity to varying attributes and relations (after finetuning). The original referring expression  was ``man walking with purple shirt woman", to which the model gave an incorrect answer. We identify two potential sources of error. First, the model might not be able to detect the woman or the man because they are in the background. Second, the model was able to detect background entities, but the reasoning step might fail to distinguish between entities. We linguistically vary the input expression in order to single out the error source.

Table \ref{tb:viz_vary_the_expression}, Middle, indicates that detection error was not the cause of an incorrect prediction, because both the woman and the man can be successfully detected via more accurate and distinct attributes. Table \ref{tb:viz_vary_the_expression}, Right suggests that the reasoning step could go wrong for subtle reasons. For example, viewing the scene as a 2D image, the blue shirt man is indeed walking \textit{next to} the purple shirt woman. But it requires nuanced reasoning about the depth or the head direction to infer that the blue shirt man could not be \textit{with} the purple shirt woman. Moreover, both ``black shirt" and ``purple shirt" occurred twice in the image. Therefore, they could possibly confuse the model. On the other hand, ``white cap" and ``yellow shorts" are distinct attributes, which easily help the model with making the correct choice.

\section{Additional Results and Error Analysis for Downstream Grounding Tasks}
\label{sup:grounding_analysis}

Table \ref{tb:REC_extended} compares the Referring Expression Comprehension performance on both the validation and test splits. Results on the test splits exhibit a similar trend as validation splits. Our performance is superior to the previous specialized models, on par with moderate-size unified models, but is lower than those very largest and slowest unified models. 
\definecolor{Gray}{gray}{0.9}

\begin{table*}[th]
\centering
\small
\resizebox{1.0\linewidth}{!}{
\begin{tabular}{@{\hspace{2pt}}p{0.02\textwidth}@{\hspace{4pt}}p{0.32\textwidth}@{\hspace{2pt}}p{0.06\textwidth}@{\hspace{4pt}}p{0.07\textwidth}@{\hspace{8pt}}p{0.06\textwidth}@{\hspace{4pt}}p{0.06\textwidth}@{\hspace{4pt}}p{0.06\textwidth}@{\hspace{12pt}}p{0.06\textwidth}@{\hspace{4pt}}p{0.06\textwidth}@{\hspace{4pt}}p{0.06\textwidth}@{\hspace{12pt}}p{0.06\textwidth}@{\hspace{4pt}}p{0.06\textwidth}@{}}

\toprule
 & Model & Time & Params & \multicolumn{3}{@{}c@{}}{Refcoco} & \multicolumn{3}{@{}c@{}}{Refcoco+} & \multicolumn{2}{@{}c@{}}{Refcocog(umd)} \\
 &  &  &  & val & testA & testB & val & testA & testB & val & test \\
\midrule
 \multirow{8}{*}{\rotatebox{90}{\emph{Modular}}} & VL-T5~\citep{VL-T5} & \phantom{0}5.9x & 290M & --- & --- & --- & --- & --- & --- & 71.2 & 71.3 \\
 & TransVG~\citep{TransVG} & \phantom{0}2.8x & 169M & 80.83 & 83.38 & 76.94 & 68.00 & 72.46 & 59.24 & 68.71 & 67.98  \\
 & TransVG++~\citep{TransVG++} & \phantom{0}--- & --- & 86.28 & 88.37 & 80.97 & 75.39 & 80.45 & 66.28 & 76.18 & 76.30 \\
 & QRNet~\citep{QRNet} & \phantom{0}5.5x & 273M &  84.01 & 85.85 & 82.34 & 72.94 & 76.17 & 63.81 &  73.03 & 72.52 \\
 & RefTrans~\citep{ReferringTrans} & \phantom{0}--- & --- & 85.65 & 88.73 & 81.16 & 77.55 & 82.26 & 68.99 & 79.25 & 80.01 \\
 & SeqTR~\citep{SeqTR} & \phantom{0}4.4x & 108M & 83.72 & 86.51 & 81.24 & 71.45 & 76.26 & 64.88 & 74.86 & 74.21 \\
 & UNITER$\mathrm{_{Large}}$~\citep{chen2020uniter} & \phantom{0}--- & 371M & 81.41 & 87.04 & 74.17 & 75.90 & 81.45 & 66.70 & 74.86 & 75.77 \\
 & VILLA$\mathrm{_{Large}}$~\citep{VILLA} & \phantom{0}--- & 371M & 82.39 & 87.48 & 74.84 & 76.17 & 81.54 & 66.84 & 76.18 & 76.71 \\
 \midrule
 & \acronymbase{} (ours) & \phantom{0}1.0x & 110M & 86.67 & 89.15 & 81.99 & 77.44 & 82.57 & 69.71 & 80.43 & 80.22 \\
 \midrule
 \midrule
 \multirow{6}{*}{\rotatebox{90}{\emph{Unified}}} 
 & OFA$\mathrm{_{Medium}}$~\citep{OFA} & \phantom{0}--- & 93M & 85.34 & 87.68 & 77.92 & 76.09 & 83.04 & 66.25 & 78.76 & 78.58 \\
 & MDETR-R101~\citep{Mdetr} & \phantom{0}3.0x & 185M & 86.75 & 89.58 & 81.41 & 79.52 & 84.09 & 70.62 & 81.64 & 80.89 \\
 & MDETR-ENB3~\citep{Mdetr} & \phantom{0}2.6x & 153M & 87.51 & 90.40 & 82.67 & 81.13 & 85.52 & 72.96 &  83.35 & 83.31 \\
 & {\color{gray}UniTAB~\citep{UniTAB}} & {\color{gray}19.4x} & {\color{gray}198M} & {\color{gray}88.59} & {\color{gray}91.06} & {\color{gray}83.75} & {\color{gray}80.97} & {\color{gray}85.36} & {\color{gray}71.55} & {\color{gray}84.58} & {\color{gray}84.70} \\
 & {\color{gray}OFA$\mathrm{_{Base}}$~\citep{OFA}} & {\color{gray}\phantom{0}6.0x} & {\color{gray}180M} & {\color{gray}88.48} & {\color{gray}90.67} & {\color{gray}83.30} & {\color{gray}81.39} & {\color{gray}87.15} & {\color{gray}74.29} & {\color{gray}82.29} & {\color{gray}82.31} \\
 & {\color{gray}OFA$\mathrm{_{Large}}$~\citep{OFA}} & {\color{gray}11.5x} & {\color{gray}470M} & {\color{gray}90.05} & {\color{gray}92.93} & {\color{gray}85.26} & {\color{gray}85.80} & {\color{gray}89.87} & {\color{gray}79.22} & {\color{gray}85.89} & {\color{gray}86.55} \\
 & {\color{gray}OFA~\citep{OFA}} & {\color{gray}\phantom{0}---} & {\color{gray}930M} & {\color{gray}92.04} & {\color{gray}94.03} & {\color{gray}88.44} & {\color{gray}87.86} & {\color{gray}91.70} & {\color{gray}80.71} & {\color{gray}88.07} & {\color{gray}88.78} \\
\bottomrule
\end{tabular}}
\caption{Results on Referring Expression Comprehension. Models are compared along two axes: efficiency-performance and efficiency-versatility. Our model beats or competes with other models, while being much lighter and faster. The Params column counts \textit{all} parameters required for inference, including the upstream RoI extractor (if needed), the main architecture, and specialized output heads.}
\label{tb:REC_extended}
\end{table*}

\begin{table*}[t]
    \centering
    \resizebox{1.0\linewidth}{!}{
    \begin{tabular}{@{}p{0.25\textwidth}p{0.36\textwidth}p{0.36\textwidth}@{}}
    
    Input image & Predicted (incorrect) & Predicted (correct) \\
    \toprule

    \raisebox{.28\totalheight}{\includegraphics[width=\linewidth]{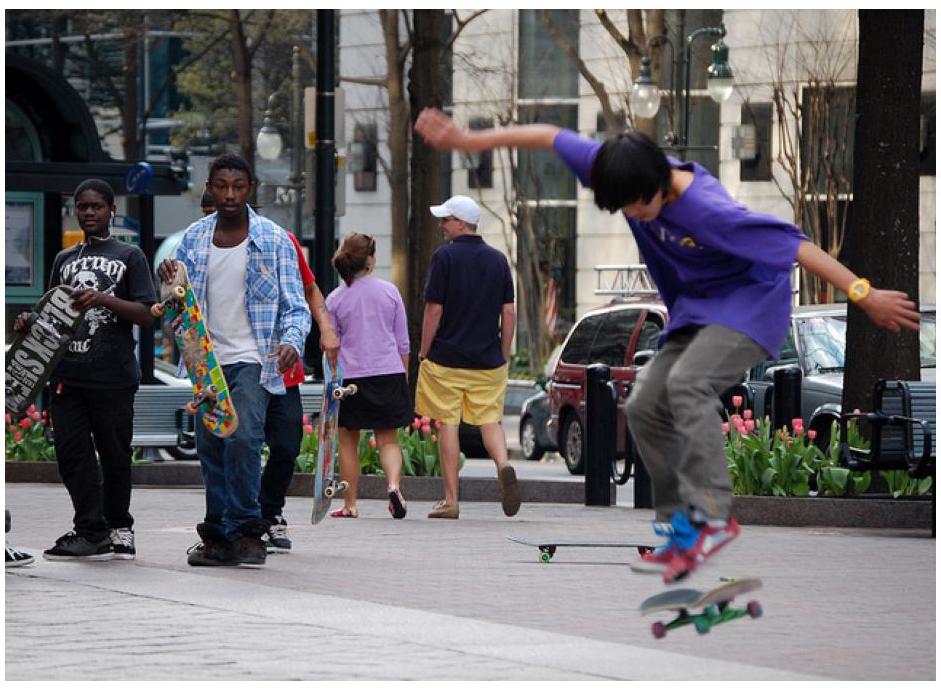}} & \includegraphics[width=0.9\linewidth]{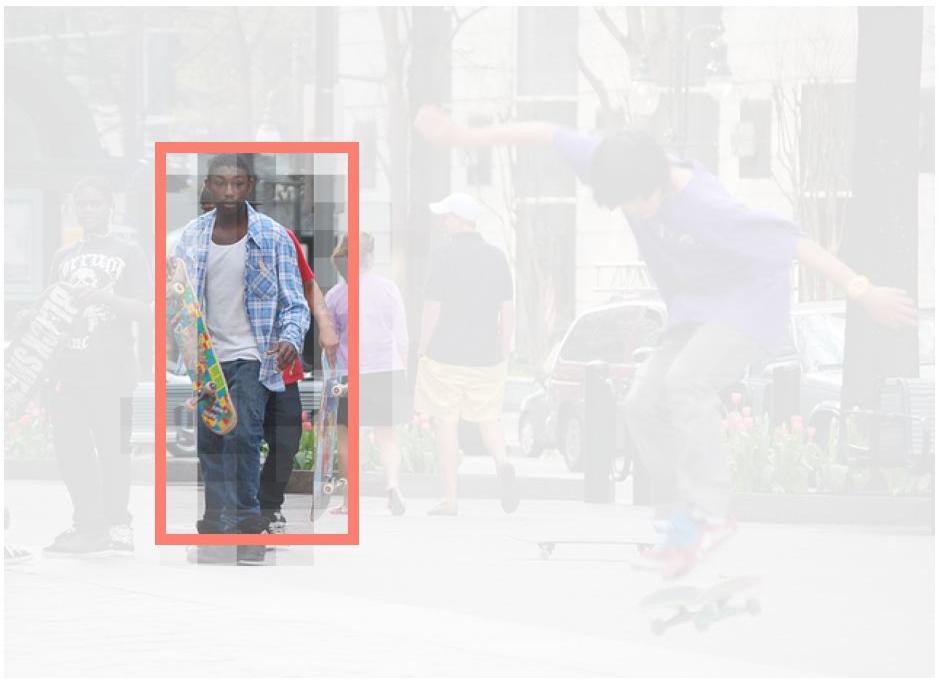} & \includegraphics[width=0.9\linewidth]{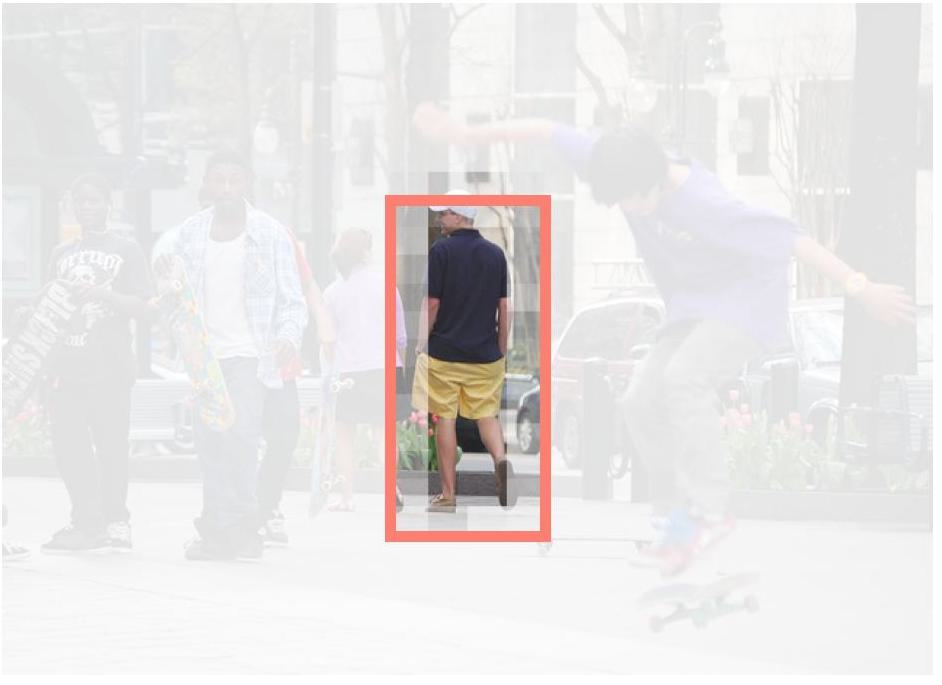} 
     \\
    
    & man walking \textbf{with purple shirt woman} & man \textbf{in white cap} \\
    \cmidrule{2-3}

    & \includegraphics[width=0.9\linewidth]{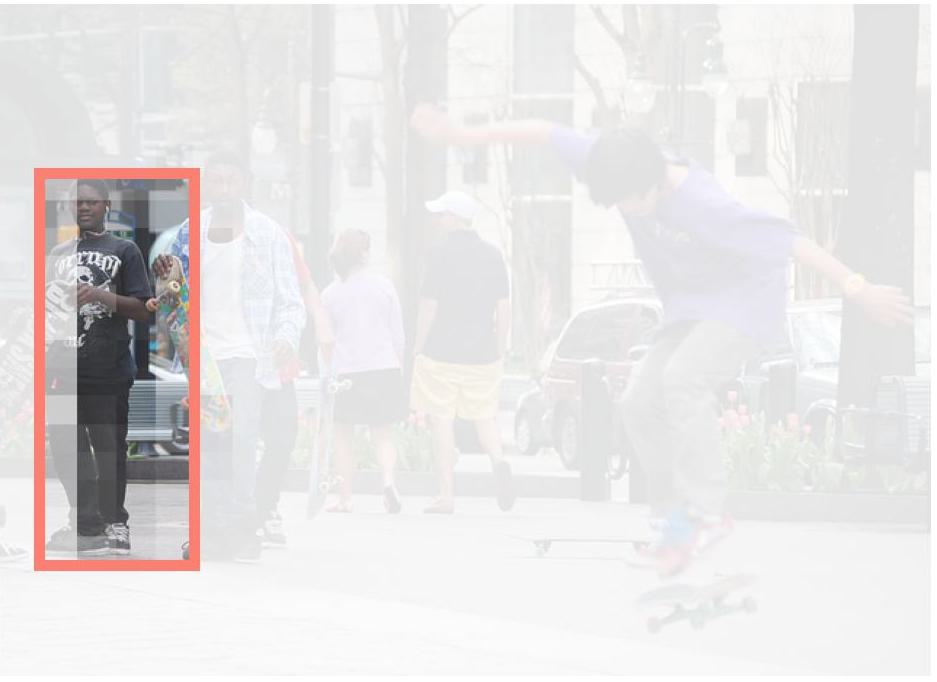} & \includegraphics[width=0.9\linewidth]{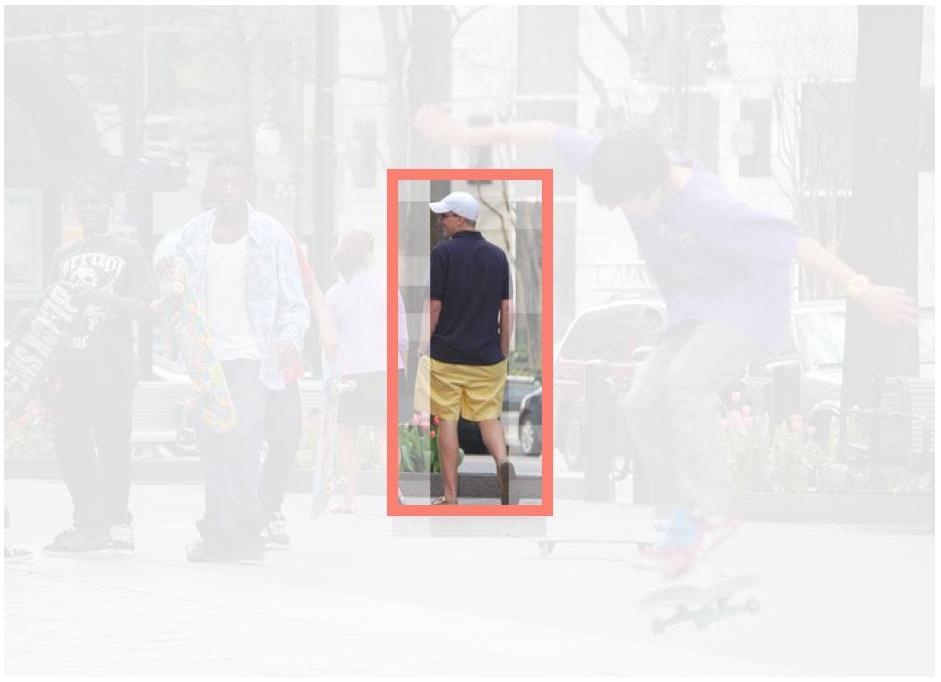} \\
    
    & man in \textbf{black shirt} walking  & man in \textbf{yellow shorts} walking \\
    & with purple shirt woman & with purple shirt woman\\
    \cmidrule{2-3}
    
    & \includegraphics[width=0.9\linewidth]{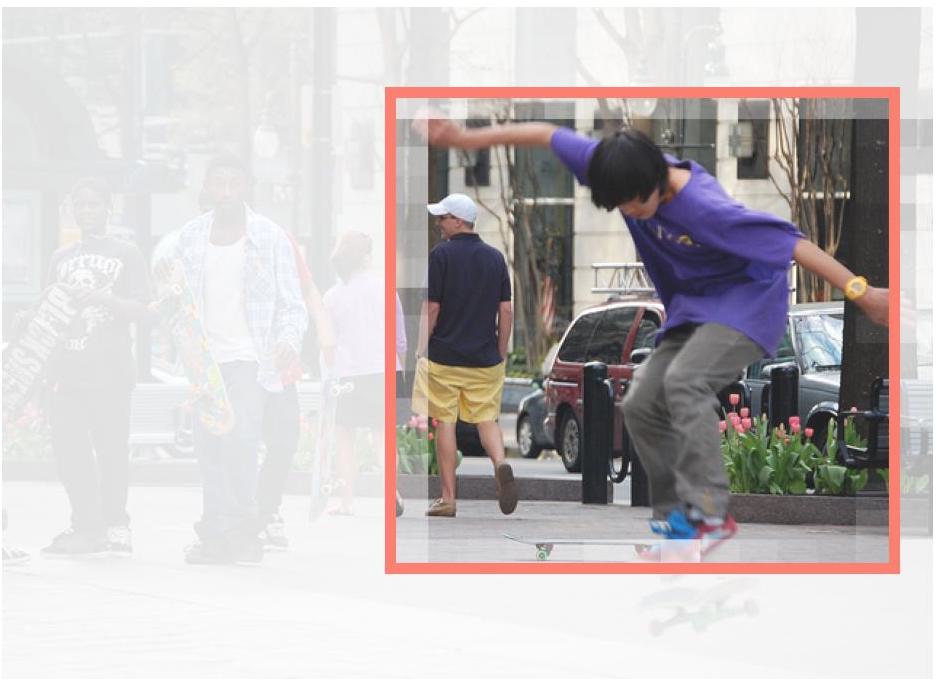} & \includegraphics[width=0.9\linewidth]{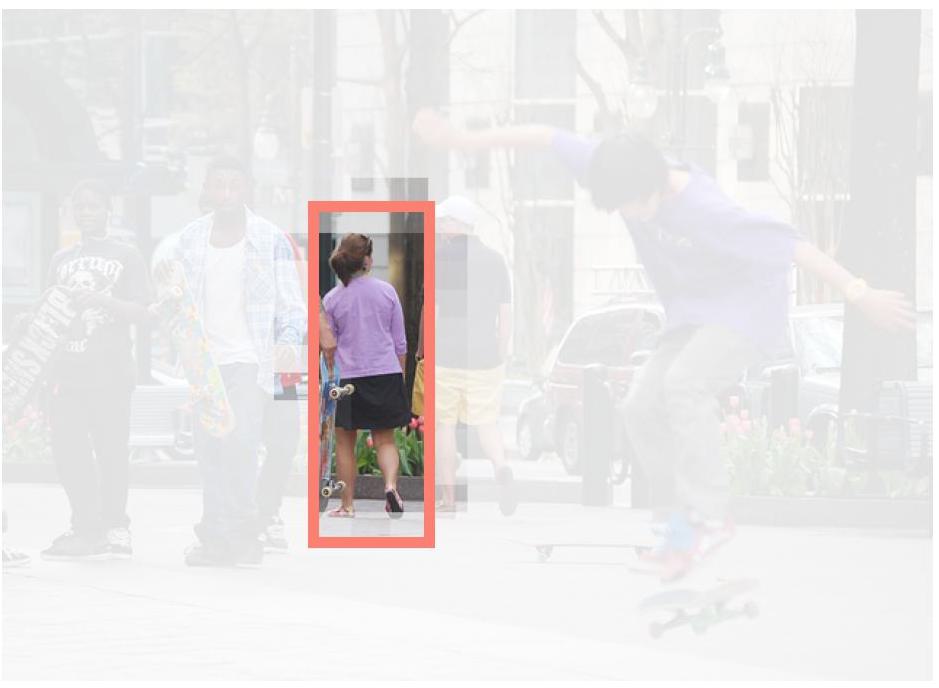} \\
    
    & woman in purple shirt & woman in purple shirt \textbf{and black shorts} \\
    \cmidrule{2-3}
    
    & \includegraphics[width=0.9\linewidth]{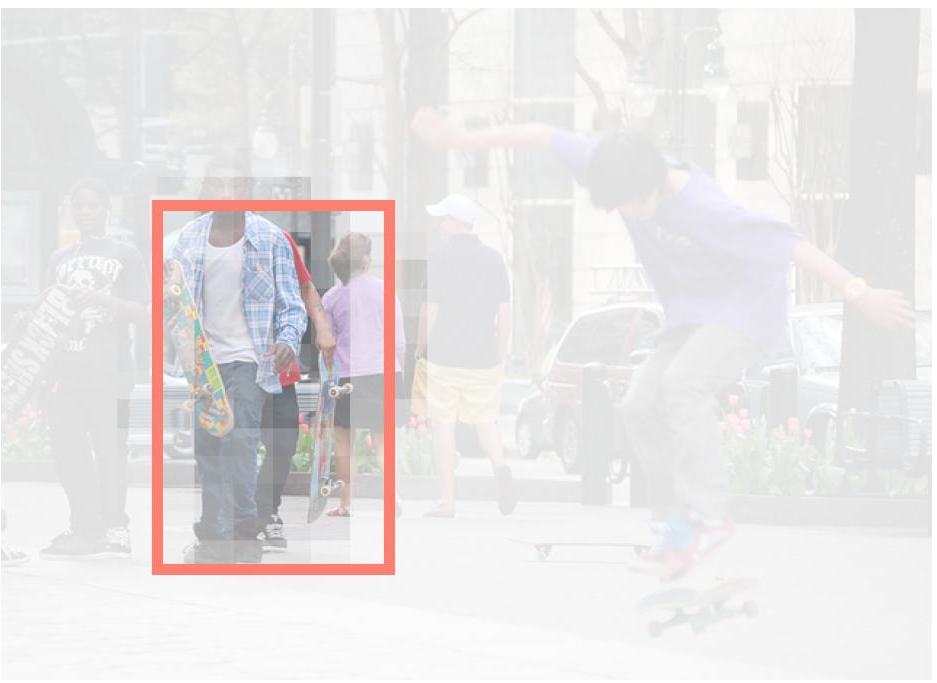} & \includegraphics[width=0.9\linewidth]{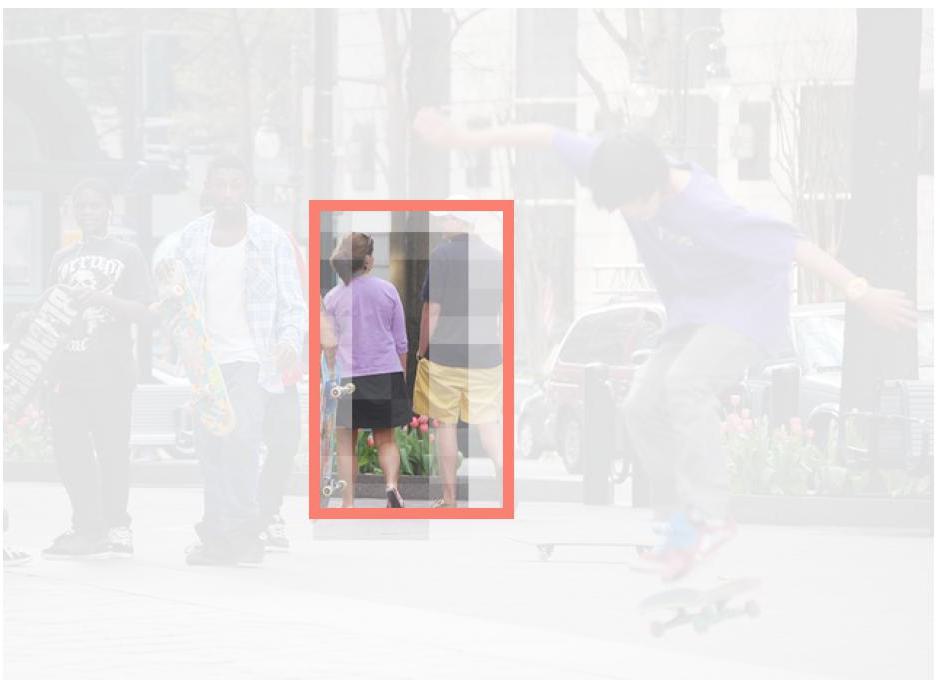} \\
    
    & woman walking with a man & women walking with a man \textbf{in yellow shorts} \\

    \bottomrule
    \end{tabular}
    }
    \caption{Qualitative demonstration: The model is able to detect small entities in the background. But the model's success at distinguishing between entities relies on accurate and distinct attributes or relations. Here we juxtapose changes to phrasing or addition of a clause.}
    \label{tb:viz_vary_the_expression}
    \vspace{20pt}
\end{table*}

Next, we analyze common prediction errors which could inform future research on the unsolved difficulties of Referring Expression tasks. Table \ref{tb:viz_refexp_errors} visualizes the error-prone cases on Referring Expression Comprehension. The model blunders when the target entity is specified through complex attributes or relations. The model also occasionally confuses a head noun with its modifiers, or gets distracted by an irrelevant entity sharing a mentioned attribute. Mistakes can also stem from inadequate text-reading ability from visual stimuli.

Table \ref{tb:breakdown} shows a performance breakdown according to four potential sources of difficulty. Table \ref{tb:breakdown}a suggests that the model particularly struggles when a referring expression contains multiple noun chunks. We believe that the number of noun chunks is a reasonable indicator of difficulty because it also correlates well with the sentence length and linguistic complexity. Table \ref{tb:breakdown}b suggests that the model is less performant when it encounters an unfamiliar head noun. We parse the expressions and extract noun chunks via SpaCy. A token is identified as the head noun if its part-of-speech tag is one of ['NOUN', 'PROPN'] and its role in the dependency tree is one of ['nsubj', 'root']\footnote{These heuristics occasionally fail to identify a head noun or identify more than one head noun in $<5\%$ of the cases.}. The head noun frequencies are counted across Refcoco/+/g\_val.

Table \ref{tb:breakdown}c plots the performance against the area of the ground-truth bounding box. We observe that the performance drops drastically on extremely small regions. Table \ref{tb:breakdown}d plots the (zero-out) segmentation performance against the convexity of the ground-truth polygon mask. We measure convexity as the ratio between the area of the polygon and the area of its convex hull. Note, the presence of holes inside a mask or non-contiguity could also lead to a small convexity score. Thus, our findings reflect ample room for  improvement on localizing small and less convex regions. 

\begin{table}[t]
    \centering
    \resizebox{1.0\linewidth}{!}{
    \begin{tabular}{@{}p{0.23\textwidth}p{0.23\textwidth}p{0.23\textwidth}p{0.23\textwidth}@{}}
    
    Input image & Predicted affinity map & Predicted bbox reduced & \textcolor{Salmon}{Predicted} vs. \textcolor{Green}{Ground-truth} \\
    \toprule
    \multicolumn{4}{@{}c@{}}{
        \includegraphics[width=\linewidth]{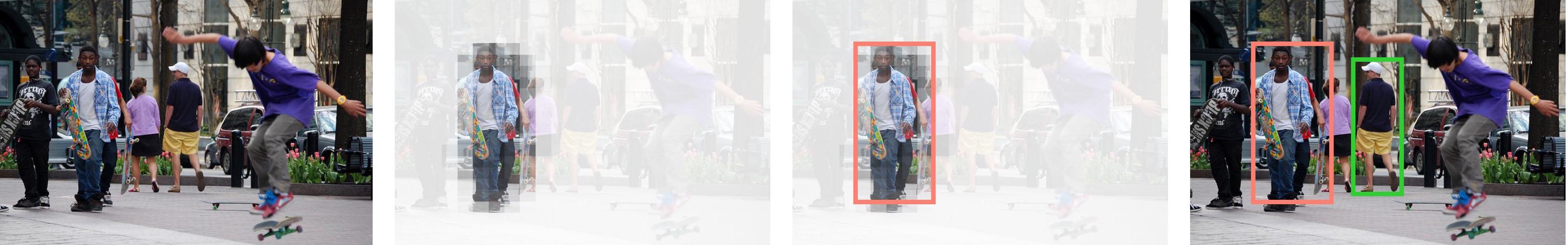}
    } \\
    \multicolumn{2}{@{}l@{}}{man walking with purple shirt woman} & \multicolumn{2}{@{}r@{}}{The model failed to disambiguate via relationship-based attributes.} \\
    \multicolumn{4}{@{}c@{}}{
        \includegraphics[width=\linewidth]{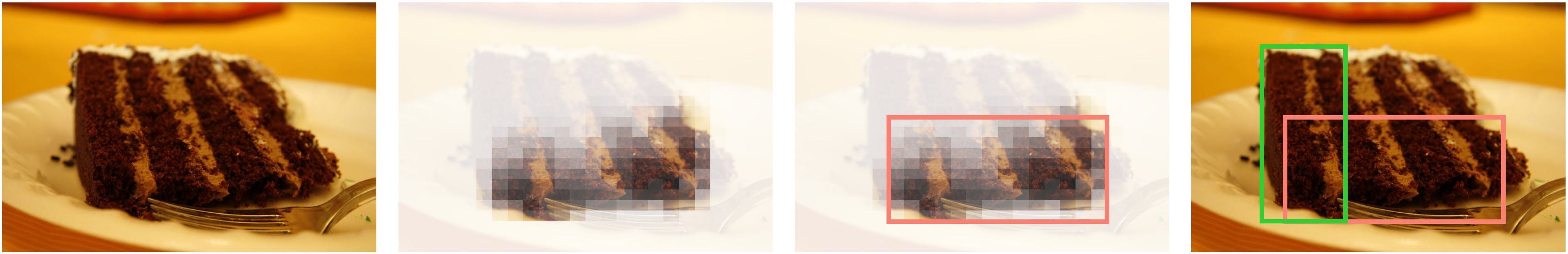}
    } \\
    cake layer by fork end & \multicolumn{3}{@{}r@{}}{The model tried to highlight the fork end while the head phrase is ``cake layer".} \\ 
    
    \multicolumn{4}{@{}c@{}}{
        \includegraphics[width=\linewidth]{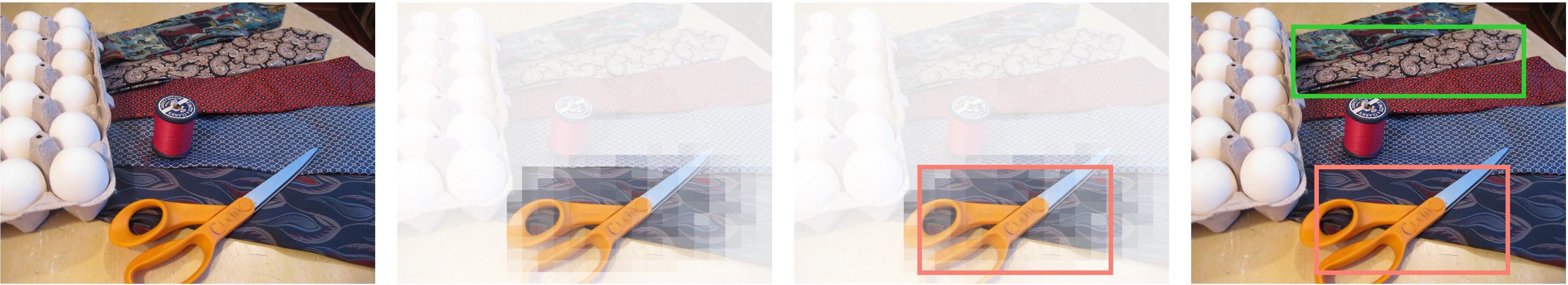}
    } \\
    black and gold tie & \multicolumn{3}{@{}r@{}}{The model was distracted by a gold scissors on a black tie.} \\
    \multicolumn{4}{@{}c@{}}{
        \includegraphics[width=\linewidth]{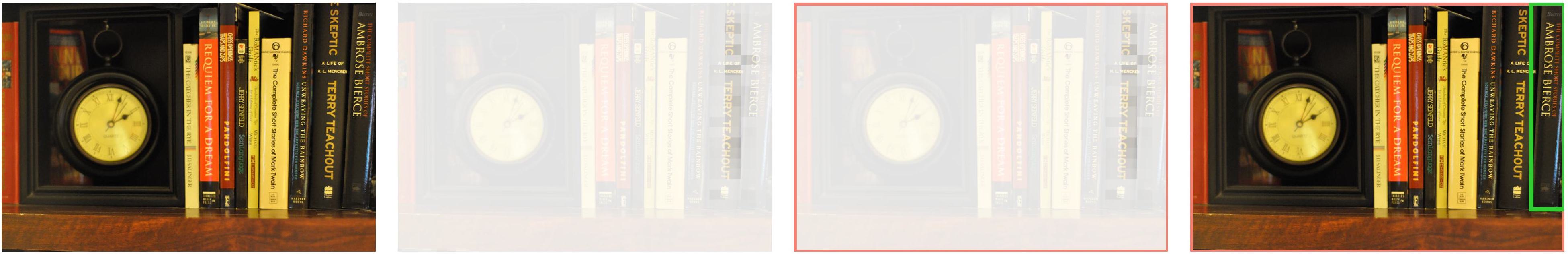}
    } \\
    anbrise bierce book & \multicolumn{3}{@{}r@{}}{The model struggled at text-reading from image, thus predicting at an extremely low confidence.} \\

    \bottomrule
    \end{tabular}
    }
    \caption{Qualitative demonstration. Common error types include 1) failure to resolve relationship-based attributes, 2) incorrect head-noun identification,  3) distraction by an irrelevant object sharing a mentioned attribute, and 4) inadequate text-reading from visual stimuli.}
    \vspace{-10pt}
    \label{tb:viz_refexp_errors}
\end{table}

\begin{table*}[ht]
    \centering
    \vspace{20pt}
    \begin{tabular}{@{}p{0.49\textwidth}p{0.49\textwidth}@{}}
    
    \includegraphics[width=\linewidth]{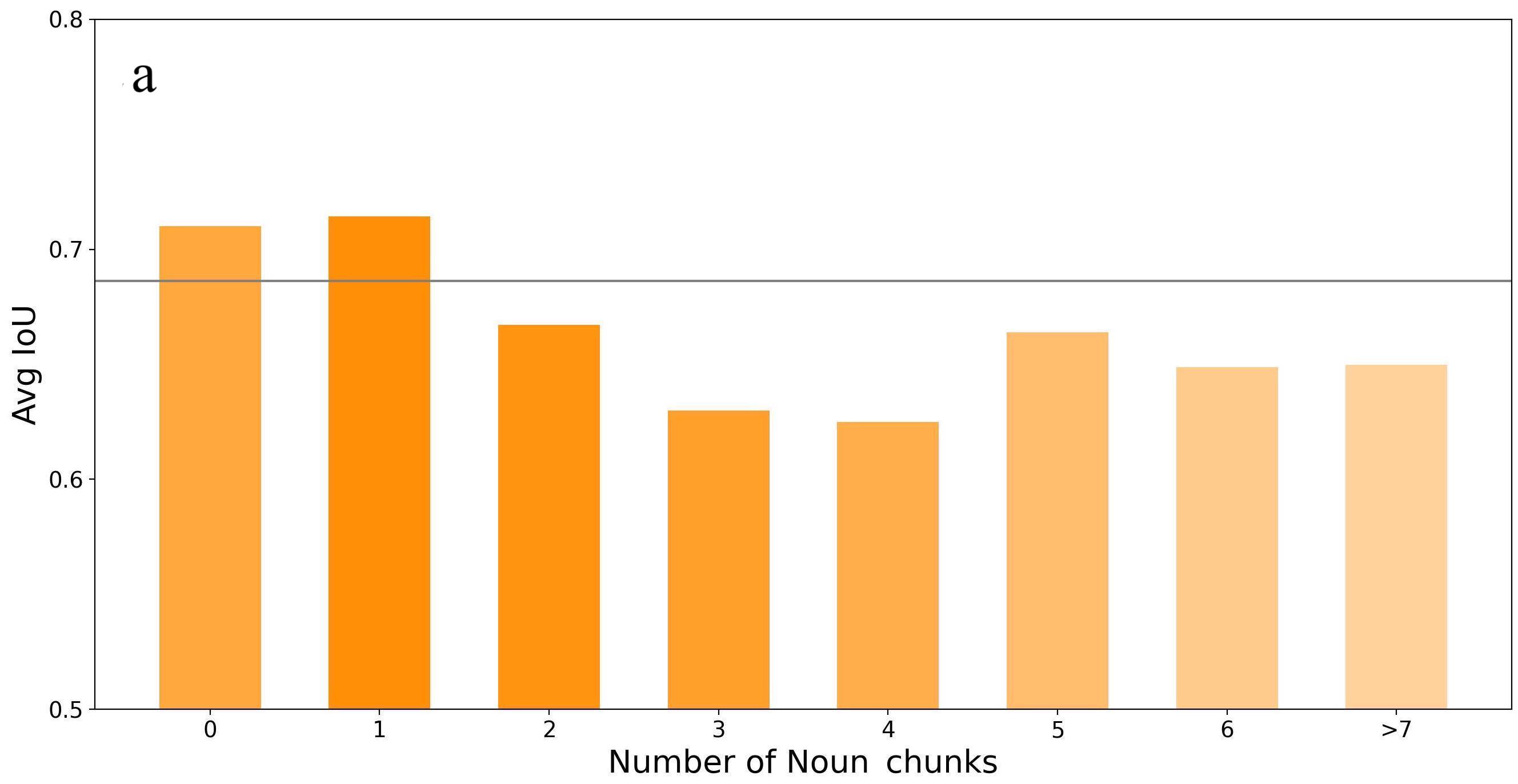} &  \includegraphics[width=\linewidth]{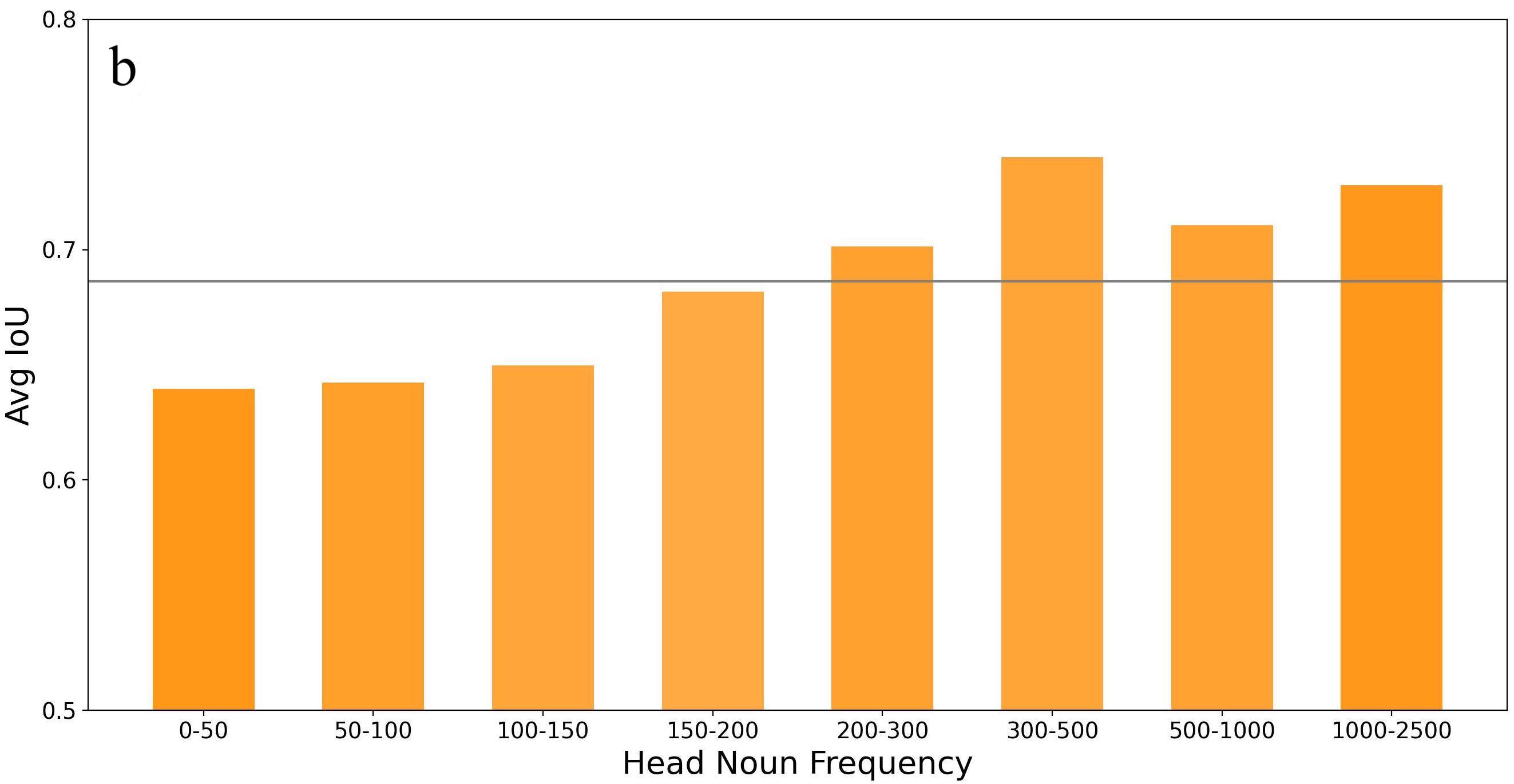} \\
    \includegraphics[width=\linewidth]{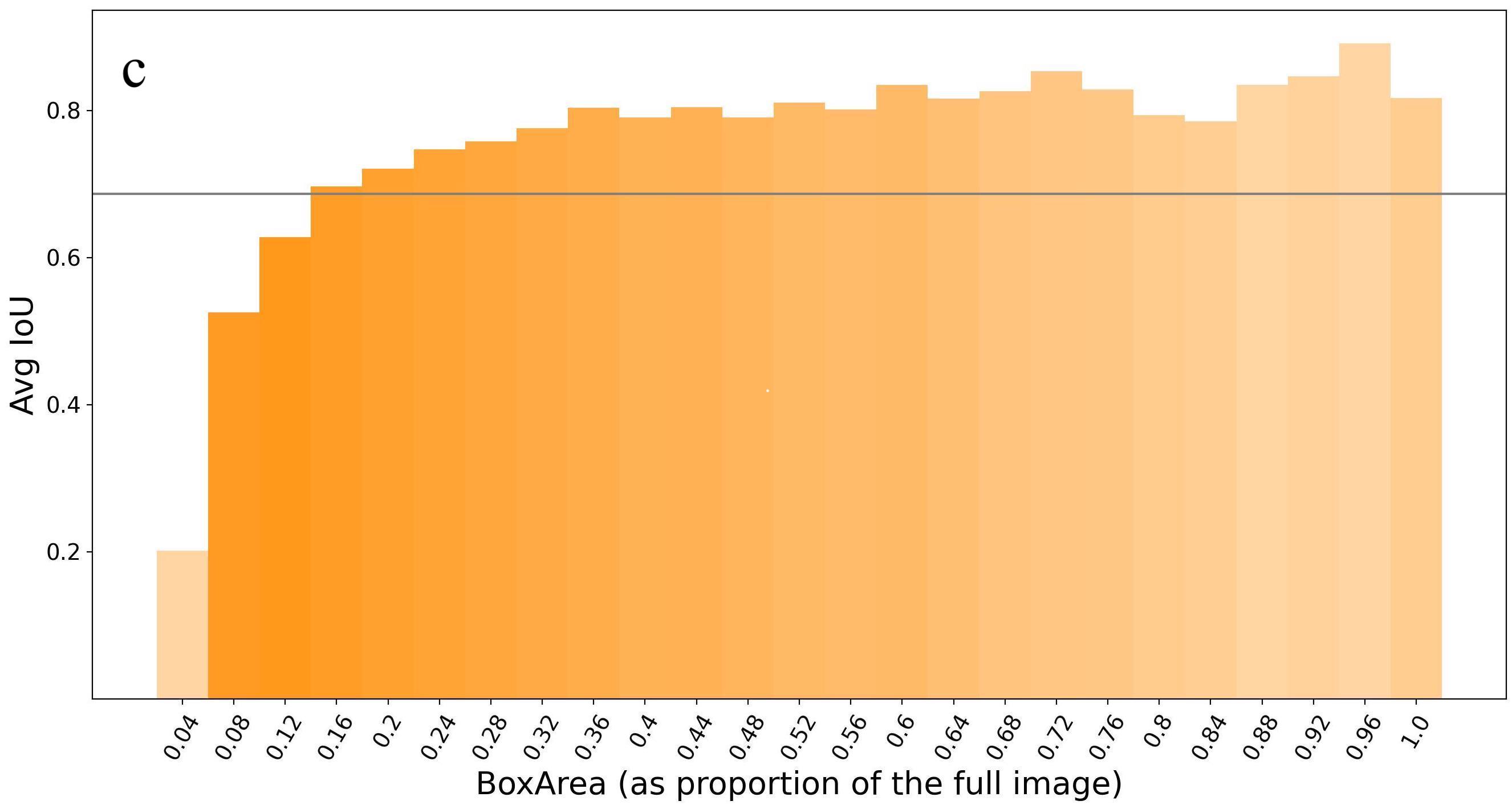} & \includegraphics[width=\linewidth]{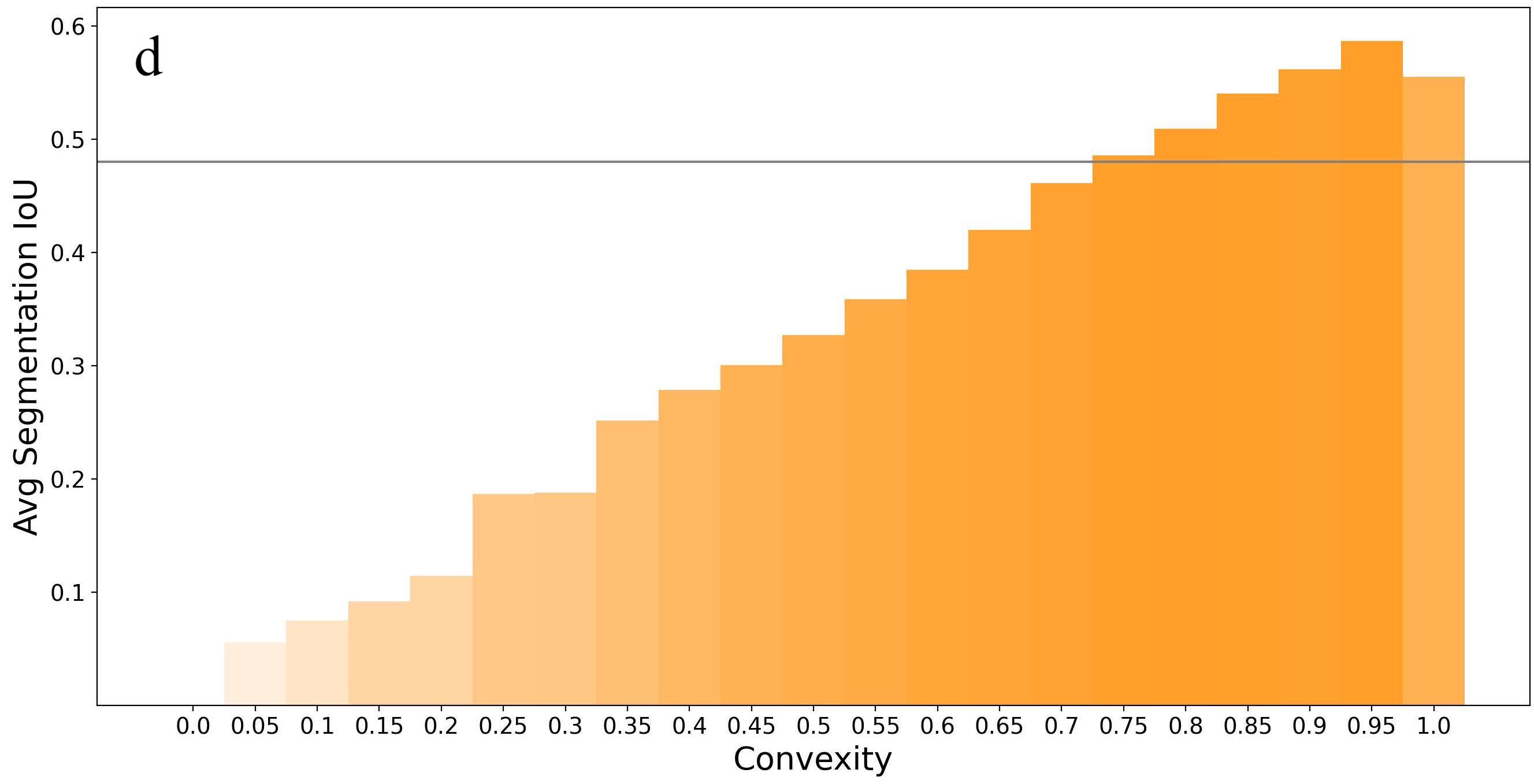} \\

    \end{tabular}
    \caption{Performance breakdown according to different indicators of the difficulty level. We observe that the performance drops when a) the referring expression contains multiple noun chunks, b) the referring expression has a rare head noun, c) the ground-truth bounding box has a small area, d) the ground-truth polygon mask is less convex. Statistics are computed across Refcoco/+/g\_val. Shades indicate the number of  examples in each bin. The horizontal line marks the global performance averaged over all bins.}
    \label{tb:breakdown}
\end{table*}

\section{Limitations and Societal Impact}
\label{sup:limitation}
In our work, we train our model with image-text pairs of 4M/5.6M unqiue images. Limited to our computation resource, it is unclear how the model performance will be with further scaling. In addition, we only focus on images with English descriptions. The generalization ability to other languages is left for the futher work.

While our paper shows promising results on vision-language pretraining and the scaling of training data and model size, additional analysis should be undertaken to assess the quality of pretraining web data before further deployment. Because web data may contain gender/racial bias, hate speech, private information and other unsuitable images. Only optimizing the performance may have unwanted model behaviors.

\end{document}